\documentclass[manuscript,screen]{acmart} 
\AtBeginDocument{%
  }

\copyrightyear{2025}
\acmYear{2025}
\acmMonth{12}
\acmISBN{XXX-X-XXXX-XXXX-X/202x/xx}

\usepackage{pifont}

\usepackage{makecell}

\usepackage{subcaption}

\let\svthefootnote\thefootnote

\begin{document}

\title[No Evaluation without fair representation]{No evaluation without fair representation : Impact of label and selection bias on the evaluation, performance and mitigation of classification models}

\author{Magali Legast}
\email{magali.legast@uclouvain.be}
\orcid{0000-0003-4246-1158}
\affiliation{%
  \institution{Universit\'{e} catholique de Louvain}
  \department{Institute of Information and Communication Technologies, Electronics and Applied Mathematics}
  \city{Louvain-la-Neuve}
  \country{Belgium}
}
\author{Toon Calders}
\email{toon.calders@uantwerpen.be}
\orcid{0000-0002-4943-6978}
\affiliation{%
  \institution{Universiteit Antwerpen}
  \department{Adrem Data Lab}
  \city{Antwerpen}
  \country{Belgium}
}

\author{Fran\c{c}ois Fouss}
\email{francois.fouss@uclouvain.be}
\orcid{0000-0001-6383-9514}
\affiliation{%
  \institution{Universit\'{e} catholique de Louvain}
  \department{Louvain Research Institute in Management and Organizations}
  \department{Institute of Information and Communication Technologies, Electronics and Applied Mathematics}
  \city{Mons}
  \country{Belgium}
}

\renewcommand{\shortauthors}{Legast et al.}

\begin{abstract}
   
    Bias can be introduced in diverse ways in machine learning datasets, for example via selection or label bias. Although these bias types in themselves have an influence on important aspects of fair machine learning, their different impact has been understudied. 
    In this work, we empirically analyze the effect of label bias and several subtypes of selection bias on the evaluation of classification models, on their performance, and on the effectiveness of bias mitigation methods. We also introduce a biasing and evaluation framework that allows to model fair worlds and their biased counterparts through the introduction of controlled bias in real-life datasets with low discrimination. Using our framework, we empirically analyze the impact of each bias type independently, while obtaining a more representative evaluation of models and mitigation methods than with the traditional use of a subset of biased data as test set.
    Our results highlight different factors that influence how impactful bias is on model performance.
    They also show an absence of trade-off between fairness and accuracy, and between individual and group fairness, when models are evaluated on a test set that does not exhibit unwanted bias. They furthermore indicate that the performance of bias mitigation methods is influenced by the type of bias present in the data. 
    Our findings call for future work to develop more accurate evaluations of prediction models and fairness interventions, but also to better understand other types of bias, more complex scenarios involving the combination of different bias types, and other factors that impact the efficiency of the mitigation methods, such as dataset characteristics. 
    
\end{abstract}



\maketitle

\section{Introduction}\label{sec:intro}

    \let\thefootnote\relax\footnotetext{The implementation of all experiments and additional results are available at \url{https://github.com/Magalii/ControlledBias/tree/BiasComp}
    .}
    \let\thefootnote\svthefootnote
    
    Machine learning (ML) classification models are more and more present into people's lives with the many ethical challenges that come along. These include concerns about the fair treatment of individuals in the predictions made about them by such models. It is indeed well documented that ML models are subject to biases that can lead to discriminatory outputs, which especially affect groups that are already socially disadvantaged \cite{mehrabi_survey_2021, ntoutsi_biasSurvey_2020}.
    To counter this situation, a large number of fairness metrics to uncover and report model bias, as well as many bias mitigation methods, have been developed in the field of algorithmic fairness  \cite{hort_biasClassSurvey_2024}.
    It has been noted that selecting the right mitigation method for a given situation is an important challenge for researchers and practitioners, especially with each method having its own specificity and (dis)advantages \cite{caton_fairnessSurvey_2024, chen_empiricalMitigMethod_2023}.
    Notably, applying the same bias mitigation method in different contexts, like different datasets (or even different splits of the same dataset \cite{friedler_comparative_2019}), can lead to different results, both regarding model accuracy and fairness improvement \cite{berk_convex_2017, baumann_biasOnDemand_2023, ganesh_differentHorses_2025}. 
    Several researchers call for practitioners to take the context into account when choosing a specific fairness intervention \cite{ganesh_differentHorses_2025, koumeri_metricsEUlaw_2023}, with however insufficient guidance on how to do that.
    It is thus needed to better understand how the context affects the performance of fairness interventions to better determine which mitigation technique should be applied in a given situation.

    Several comparative studies have been made in that direction, such as \cite{biswas_compKaggle_2020, cardoso_benchmarkingframework_2019, chen_empiricalMitigMethod_2023, defrance_abcfair_2024, friedler_comparative_2019, ganesh_differentHorses_2025, han_ffb_2024, hort_fairea_2021, jones_comparison_2020}.
    They each analyze and compare several bias mitigation methods for classification tasks involving tabular data. Most of them share some limitations that are common in the field of fairness, yet hinder a comprehensive understanding of these methods' behavior. We discuss these shortcomings in the next two paragraphs.

    A first limitation is that bias mitigation methods are typically evaluated on datasets that are coincidentally acknowledged as biased, hence the need for bias mitigation, and used to evaluate the fairness and accuracy achieved by the mitigated and supposedly fairer model \cite{cooper_emergentUnfairness_2021, wick_unlocking_2019}. This evaluation setting induces the emergence of the so-called fairness-accuracy trade-off, which considers that fairness and accuracy\footnote{As in most work in fairness research, we consider accuracy as label alignment, i.e. the proportion of predicted labels that match known labels \cite{cooper_emergentUnfairness_2021}} are in tension with one another, and that the increase of one inevitably leads to a decrease of the other \cite{cooper_emergentUnfairness_2021}.
    The existence of such a trade-off is largely accepted in the fairness research community, even though instances of fairness and accuracy growing together have been observed as early as 2010 \cite{kamiran_tree_2010}. Moreover, the inevitability of the fairness-accuracy trade-off has been disproved in more recent work \cite{sharma_dataAugmentation_2020, lenders_studentExtended_2023, wick_unlocking_2019}.
    \citet{favier_biasTheoretical_2023} and \citet{wick_unlocking_2019} show that the use of biased data as ground truth leads to biased results and prevents the correct evaluation of how far the model predictions are from a fair outcome. 
    Beyond criticism of the fairness-accuracy trade-off itself, these results also call into question the most common evaluation practice regarding bias mitigation methods, and in turn the conclusions drawn from it. This includes most of the results in the comparative studies mentioned above. 
    
    A second limitation, and very prominent in relation to this work, is the use of datasets in which the source and extent of bias are unclear.
    Most work in the field of fair classification use readily available datasets \cite{hort_biasClassSurvey_2024}, which most often present limitations \cite{fabris_storySoFar_2022}. This includes the three most popular ones, namely Adult \cite{adult_1996}, COMPAS \cite{dressel_compas_2018} and German credit \cite{german_credit}, that have attracted serious criticism suggesting that these dataset flaws have the potential to hinder the significance of results drawn from experiments based on their use \cite{bao_COMPASlicated_2021, Barenstein_ProPublicasCD_2019, ding_retiringAdult_2021, gromping_GermanCorrection_2018}. 
    Beyond the problem of dataset quality, only having access to biased datasets with no controlled or precise understanding of these biases makes it difficult to draw conclusions about the performance and behavior of bias mitigation methods.
    Indeed, if we do not know what bias is at play in the data, determining the effect a method has on them becomes difficult, as is predicting if the performance observed will be replicated in another setting.

    Several results highlight the process of bias injection as an important element of context that needs to be taken into account throughout the different stages of the ML pipeline. 
    \citet{ceccon_dataBiasProfile_2025} point out that the respective impact of label and selection bias on models has been misevaluated. \citet{favier_biasTheoretical_2023} theoretically show that these two bias types can perturb the evaluation of models by affecting the measurement of fairness metrics, which also influences the performance of the bias mitigation methods relying upon them. The theoretical result that different types of biases are not as efficiently reduced by the same mitigation methods is confirmed by the small-scale experiments in \cite{baumann_biasOnDemand_2023, wick_unlocking_2019}.
    
    Building on these findings, we empirically study the effect of label bias and different subtypes of selection bias on the performance of classification models, on their evaluation, and on the effectiveness of bias mitigation methods.
    
    \paragraph{\textbf{Main contributions}}

        We propose a biasing and evaluation framework that allows to train models on data containing known bias and to evaluate them on an unbiased test set, leading to a fair and context-sensitive evaluation of prediction models and bias mitigation methods. Our findings empirically confirm that using biased data to evaluate models can lead to misleading results, highlighting the need for such a framework. 
        With experiments on unmitigated models, we identify factors that lead the effect of training data bias to be more or less impactful on model performance. 
        Moreover, our results show that the way bias is introduced in the dataset has a significant influence on the performance of mitigation methods, with clear differences between injection types, confirming the theoretical findings in \citet{favier_biasTheoretical_2023} and preliminary results of \citet{baumann_biasOnDemand_2023} and \citet{wick_unlocking_2019}.
        We provide a discussion of the methods performance in relation with each bias type.
        The efficiency of the methods also varies according to other contextual factors, such as dataset distribution and bias intensity.

        Our work opens up several avenues for future research, including :
        \begin{itemize}
            \item The creation of more dual datasets that can be used to fairly evaluate bias mitigation methods, or development of other approaches that can lead to a fair evaluation,
            \item Research on the impact of bias types beyond label and selection bias, as well as combination of different bias types, on evaluation metrics, model behavior, and bias mitigation methods,
            \item The study of the influence of datasets characteristics, such as proportion of unprivileged group and bias intensity, 
            \item The development of metrics and mitigation methods that are more robust to bias in the training data.
        \end{itemize}
        Advancement in these directions would provide further guidance to researchers and practitioners in choosing the best interventions to obtain sufficiently fair prediction models.

    \paragraph{\textbf{Structure}}
    We discuss the related work in Section \ref{sec:rel_work}. We present the way we model and evaluate biases in Section \ref{sec:method}. We then motivate and describe our biasing and evaluation framework in Section \ref{sec:framework_sec}, before exposing the experiment setup in Section \ref{sec:expe}. 
    The results are presented in Section \ref{sec:results}. We first relate measurements obtained from fair and biased evaluations in Section \ref{sec:res_biased_eval}, then we analyze the impact of bias on model performances in Section \ref{sec:res_model} and on the efficiency of the different bias mitigation methods in Section \ref{sec:res_mitig}. We conclude this paper with Section \ref{sec:ccl}.
    
\section{Related work}\label{sec:rel_work}

    \subsection{Analysis of different bias types}
        
        Research related to bias types may cover different topics, ranging from the identification of such types to their respective influence on different aspects of the ML pipeline.
        Several authors have theorized the various origins of data bias and categorized them into distinct bias types. This includes \citet{olteanu_socialData_2019} who provide a comprehensive list of bias types related to social data, \citet{suresh_biasFramework_2021} who organize bias categories according to different stages of ML life cycle, and \citet{guerdan_groundlessTruth_2023} who identify five sources of bias that affect the reliability of proxy labels.
        
        Work focusing on the detection and identification of bias present in a specific dataset is limited. \citet{clavell_biasSignals_2024} focus on such bias diagnosis through the use of operational and demographic reference data in addition to the training data. \citet{ceccon_dataBiasProfile_2025} propose mechanisms to detect label, selection and proxy bias, and experimentally study their effect on model fairness.
        
        Beyond their impact on the prediction models themselves, data bias may also distort evaluation results, as exposed by several mostly theoretical articles. \citet{fogliato_noisyLabels_2020} propose a sensitivity analysis framework to identify when label bias can lead to misleading fairness measurements, while \citet{brzezinski_imbalance_2024} use probability mass functions to study such evaluation skew in case of class imbalance and variation in protected group ratio. In \citet{favier_biasTheoretical_2023}, the influence of label and selection bias on fairness metrics is explained mathematically to assess the effectiveness of mitigation methods relying upon these metrics. 
        \citet{wick_unlocking_2019}, which is more focused on the relationship between fairness and accuracy, show the different effects of label and selection bias, and call for future work to better understand how label bias impacts fairness and mitigation methods.
        To allow for such research on the varying performances of mitigation methods for different bias type, \citet{baumann_biasOnDemand_2023} propose a bias modeling framework. Their small-scale experiment shows divergent mitigation results depending on bias types and they call for further research on the topic. 

        Our work extends the existing results with an empirical study based on real-life data that covers label bias and three different manifestations of selection bias. We use a variety of measures to analyze the impact of these bias types on the performance of bias mitigation methods, but also on the models themselves and their evaluation.

    \subsection{Refutation of fairness-accuracy trade-off}
    
        Even though the fairness-accuracy trade-off is usually taken as a given in fairness research, several papers have refuted its ubiquity. 
        Among these, \citet{wick_unlocking_2019} perform experiments with simulated data and show that fairness and accuracy can be concordant when measured on unbiased data. \citet{favier_biasTheoretical_2023} provide a theoretical analysis that explains the former results. In parallel, \citet{dutta_mismatchedHypothesis_2020} use information-theory tools to theoretically demonstrate that there exists ideal distributions in which fairness and accuracy are in agreement if they are considered with regard to the ideal distribution. They thus further argue against the evaluation of fairness and accuracy metrics on biased datasets.
        
        To permit training with biased label and evaluation with fair ones in a realistic scenario, \citet{lenders_studentExtended_2023} propose an extension of a real-life dataset with a biased version of the existing labels. The preliminary experiments they perform show that some bias mitigation methods that perform well when evaluated on biased labels have lower performance when evaluated on fair ones, which is also observed in \cite{defrance_abcfair_2024}. \citet{sharma_dataAugmentation_2020}, who propose the creation of fairer datasets through data augmentation, report similar results when accuracy is evaluated against the fairer dataset rather than the original biased one.
        Using a socio-technical lens, \citet{cooper_emergentUnfairness_2021} analyze how implicit normative assumptions, such as considering that accuracy and fairness can be measured independently from historical context, lead to the formulation of a trade-off between fairness and accuracy, and how unfairness emerges from it.
    
        We add to the existing literature with our empirical comparison between the evaluation of bias mitigation methods with a biased test set on one hand, and an unbiased test set representing a fair world on the other. We thus empirically confirm the absence of the fairness-accuracy trade-off under the assumption that the goal of fairness interventions is to bring model predictions closer to the fairer version of the test labels.         
            
    \subsection{Comparison of bias mitigation methods}
        
        Several papers have proposed comparison of different bias mitigation methods, each with their own approach and goal. \citet{friedler_comparative_2019} are among the first to perform such a comparative study. They focus on group-fairness metrics and use different train-test splits to simulate variations in dataset composition. \citet{biswas_compKaggle_2020} aim for a more realistic setting by applying the mitigation methods studied on models from Kaggle. 
        Some articles have a specific focus on the supposed fairness-performance trade-off. \citet{jones_comparison_2020} introduce the \textit{fair efficiency} metric to quantify it and study several methods in different modeling pipelines. The benchmark Fairea \cite{hort_fairea_2021} provides baselines for the fairness-accuracy trade-off. On the other hand, \citet{ganesh_differentHorses_2025} highlight the influence of the learning pipeline and show how hyper-parameter optimization can reduce this trade-off. \citet{chen_empiricalMitigMethod_2023}, which offer the most comprehensive study, include 17 methods and 20 different fairness-performance trade-offs.
        
        All the above mentioned studies use biased datasets to both train and test their models in order to evaluate the mitigation methods and the supposed trade-offs. The benchmark FFB \cite{han_ffb_2024}, which focuses on studying in-processing methods with consistent pre-processing, also uses biased test sets, while acknowledging the limitation of that practice and calling for research on methods to measure unbiased accuracy. Going further, ABCFair \cite{defrance_abcfair_2024}, a benchmark that can adapt to different problem settings, includes the dataset proposed in \cite{lenders_studentExtended_2023}, which contains both fair and biased labels. Most of their experiments are however performed on traditional datasets due to the scarcity of datasets with dual labels. The earlier framework proposed by \citet{cardoso_benchmarkingframework_2019} also allows to use fairer datasets as test sets. They generate synthetic datasets by sampling data from Bayesian networks learned on real-world datasets and change the conditional probabilities of the sensitive attribute on the outcome to introduce bias. They perform their experiment using the original datasets as test set, which includes already biased datasets.

        In parallel to the use of biased test sets, these studies are based on datasets exhibiting bias from unclear source, with the sole exception of \cite{jones_comparison_2020} that also uses synthetic data with a predefined bias type. In most of these works, the datasets employed include the most popular and criticized datasets that are Adult \cite{adult_1996}, COMPAS \cite{dressel_compas_2018} and German credit \cite{german_credit}.

        \begin{table}[h]
            \begin{tabular}{|l|c|c|c|c|}
            \hline
               Article & Dual dataset & Main findings based on fair test set & Known bias & Avoid criticized datasets \\\hline
               \citet{friedler_comparative_2019}  & \ding{55} & \ding{55} & \ding{55}  &  \ding{55} \\%
               \citet{biswas_compKaggle_2020}  & \ding{55} & \ding{55} & \ding{55}  &  \ding{55} \\
               \citet{jones_comparison_2020}  & \ding{55} & \ding{55} & \ding{51}  &  \ding{81} \\%
               Fairea \cite{hort_fairea_2021}  & \ding{55} & \ding{55} & \ding{55}  &  \ding{55} \\%
               \citet{ganesh_differentHorses_2025}  & \ding{55} & \ding{55} & \ding{55}  &  \ding{81} \\
               \citet{chen_empiricalMitigMethod_2023}  & \ding{55} & \ding{55} & \ding{55}  &  \ding{55} \\%
               FFB \cite{han_ffb_2024}  & \ding{55} & \ding{55} & \ding{55}  &  \ding{51} \\
               ABCFair \cite{defrance_abcfair_2024}  & \ding{51} & \ding{55} & \ding{55}  &  \ding{51} \\%
               \citet{cardoso_benchmarkingframework_2019}  & \ding{51} & \ding{81} & \ding{55}  &  \ding{55} \\\hline%
               This work  & \ding{51} & \ding{51} & \ding{51}  &  \ding{51} \\
            \hline
            \end{tabular}
        \caption{Overview of the evaluation settings in related work compared to ours. A \ding{81} indicates a more nuanced response, like the use of both criticized datasets and more than two other alternatives.}
        \label{tab:related_work}
        \end{table}
        
        With regard to the evaluation of bias mitigation methods, the novelty of our work is twofold. On the one hand, we evaluate the bias mitigation methods using test sets that can be considered as unbiased. On the other hand, we use train sets with known bias, as we create them by introducing controlled perturbation in real-world datasets considered to be fair. An overview of our contribution with regard to the evaluation setting in existing work is provided in Table \ref{tab:related_work}.

\section{Bias modeling and measurement}\label{sec:method}

    In this section, we present the conceptual framework considered in this work (Section \ref{sec:fwf}), the description and modeling of the bias types we study (Section \ref{sec:bias_model}), and the evaluation metrics we use to evaluate their presence (Section \ref{sec:eval}).

    \subsection{Conceptual framework}\label{sec:fwf}
    
\begin{figure}[h]
            \includegraphics[alt={The observation of the fair world is distorted and leads to data with unknown bias. This data is used to train prediction models, potentially with bias mitigation.},width=0.75\textwidth]{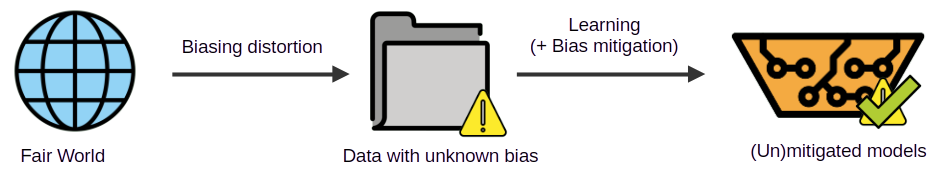}
            \caption{Fair World Framework. The fair world represents an ideal situation in which the desired fairness criteria hold. The available observable data is a distorted representation of that world, typically obtained through an unknown biasing process. Such data is traditionally used to both train prediction models and evaluate their performance.\protect\footnotemark}
            \label{fig:fair_world_framework}
        \end{figure}

        Biases and their introduction process have been represented and modeled using different existing frameworks. For this article, we mainly draw upon the \textit{Fair World Framework} \cite{favier_biasTheoretical_2023}. We thus suppose the existence of a fair underlying world in which fairness is satisfied. This world is distorted by the introduction of biases, leading to the biased and potentially unfair representations in the observed datasets. This biasing process is represented in Figure \ref{fig:fair_world_framework}.
        In this framework, a fair model not only respects certain fairness criteria, but also produces accurate predictions regarding to the fair world. 
       
    \subsection{Bias modeling}\label{sec:bias_model}

    \footnotetext{All symbols composing the diagrams in Figure\ref{fig:fair_world_framework} and \ref{fig:biasing_framework} have been designed by \href{https://openmoji.org/}{OpenMoji}, the open-source emoji and icon project, and were adapted for this document. License: \href{https://creativecommons.org/licenses/by-sa/4.0/}{CC BY-SA 4.0}}

        The biasing process leading from the fair world to its distorted observable version(s) can have many different origins. We model in this work two common potential bias, label bias and selection bias. We point toward \cite{baumann_biasOnDemand_2023} for other fundamental bias types in data generation and ways to model them. 
        
        We assume a binary classification setting in which the training set contains binary labels obtained by applying a threshold to a numerical score $S$ that is also available, such as an exam grade. We consider the sensitive attribute $A$ to be binary, with $A=1$ indicating membership to the unprivileged group and $A=0$ to the privileged group.
    
        \subsubsection{Label bias} This distortion occurs when the label encoded in the training data does not correspond to that of the fair world \cite{favier_biasTheoretical_2023}. As in \cite{baumann_biasOnDemand_2023}, which calls it \textit{measurement bias on label}, we model it as a penalty on the unprivileged group.
        
        We consider a biased score $S_b$ obtained through a distortion of the original and fair score $S$. The biased binary label is obtained by applying a threshold on $S_b$. Based on \cite{baumann_biasOnDemand_2023}, we compute the biased score as follow :
        \begin{displaymath}
            S_b = S - \beta_l*A*scale + N,\quad N \sim \mathcal{N}(0,\beta_n*scale)
        \end{displaymath}
        where the parameter $\beta_l$ is a value in $[0,1]$ controlling the bias intensity and $scale$ corresponds to half of the difference between the maximal and minimal values observed for $S$. 
        $N$ is a normally distributed random noise, with the parameter $\beta_n$ controlling the noise intensity.
        
        \subsubsection{Selection bias} Called \textit{representation bias} in \cite{baumann_biasOnDemand_2023}, it occurs when the sample of individuals in the training data does not accurately represent the distribution of the fair world \cite{favier_biasTheoretical_2023}.
        Selection bias encompasses different possible ways the original distribution can be misrepresented. 
        We focus in this work on three distinct manifestations of biased selection :
        \begin{itemize}
            \item \textit{Random selection} : We model this bias subtype by randomly undersampling a proportion $p_u$ of the underprivileged group, as in \cite{ceccon_dataBiasProfile_2025}. This can occur when data about a certain group is harder to access, but still accurately represents the original within-group distribution.
    
            \item \textit{Self-selection} : We undersample individuals from the unprivileged group with a higher removal probability for individuals with lower score $S$. The parameter $p_u$ controls the proportion of the unprivileged group that is removed from the dataset. This represents a scenario in which the (self-)selection process is stricter for unprivileged individuals, such as job application in male-typed domains where women with lower qualifications are less likely to apply than similarly qualified men \cite{coffman_apply_2024}.
            
            \item \textit{Malicious selection} : To represent a highly discriminative scenario, we perform random undersampling among both the unprivileged group with positive label and the privileged group with negative label. The parameter $p_u$ controls the proportion of individuals to be removed for both groups. This scenario corresponds for example to a situation in which someone would maliciously select privileged individuals with positive labels and unprivileged individuals with negative labels to show a difference in outcome between the two groups.
        \end{itemize}

    \subsection{Evaluation metrics}\label{sec:eval}

        To evaluate the fairness and accuracy of models' predictions, we select five evaluation metrics, including accuracy and four fairness metrics assessing different fairness goals.
        
        Recall $A$ denotes the sensitive attribute, with $A=1$ indicating membership to the unprivileged group. Let $Y$ be the binary classification label, with $D_f$ the distribution of fair labels and $D_b$ the distribution of biased labels. 
        Let $\hat{Y}$ be the predicted outcome.

        \subsubsection{Accuracy}
        As in most ML fairness research, we consider accuracy as the percentage of predicted labels $\hat{Y}$ that match known labels $Y$ considered as ground truth ($Acc = P[\hat{Y}=y|Y=y]$).
        If the unbiased labels are used, accuracy can also be viewed as a measure of fairness since it evaluates the alignment between the model predictions and the fair world labels. As explained in Section \ref{sec:intro}, using the biased labels leads to a biased accuracy value and a trade-off between fairness metrics and that biased accuracy measure \cite{cooper_emergentUnfairness_2021, wick_unlocking_2019}.
        
        \subsubsection{Statistical Parity Difference}
        Statistical Parity is achieved when the label and the sensitive attribute are statistically independent, each group thus having identical probability to receive the same outcome \cite{calders_naiveBayes_2010}. We use the Statistical Parity Difference metric (SPD) to evaluate how far a dataset is from that fairness criterion.
        \begin{displaymath}
            SPD = P(\hat{Y}=+ | A=1) - P(\hat{Y}=+ | A = 0)
        \end{displaymath}
        SPD is a group fairness metric based on predicted outcome \cite{verma_definitions_2018}.
        As such, \cite{wachter2020bias} categorizes it as a bias transforming metric\footnote{\citet{wachter2020bias} defines biased transforming metrics as "not necessarily satisfied by a perfect classifier". They are defined in opposition to bias preserving metrics that are "always satisfied by a perfect classifier that exactly predicts its target labels with zero error, replicating bias present in the data."}
        and argues that it has the potential to promote equality in practice (\textit{substantive equality}) since it does not require to replicate the labels of the biased observed data. \citet{favier_biasTheoretical_2023} however theoretically shows that its result is skewed when it is applied on data with selection bias, e.g. a dataset exhibiting selection bias and a SPD value of zero does not imply that SPD would be zero in the fair world as well.

        \subsubsection{Equalized Odds Difference} A model satisfies the Equalized Odds fairness criterion if its predictions $\hat{Y}$ and the sensitive attribute $A$ are independent conditional on known labels $Y$ \cite{hardt_equalityOpp_2016}. We use the Equalized Odds Difference metric (EqOd), which corresponds to the greater absolute value of between group differences for True Positive Rate (TPR) and False Positive Rate (FPR).
        \begin{displaymath}
            EqOd = max(|TPR_{A=1} - TPR_{A=0}|, |FPR_{A=1} - FPR_{A=0}|)
        \end{displaymath}

        EqOd is based on group-conditioned accuracy \cite{friedler_comparative_2019}. It is thus correlated with accuracy itself and can suffer the same bias. In the traditional evaluation scheme where $D_b$ is used as ground truth, it favors the replication of past biases and is categorized as bias preserving \cite{wachter2020bias}. However, if it is evaluated with $D_f$, it would then be bias transforming.

        We will also use two related and relaxed metrics in the bias mitigation methods. The Average Odds Difference (AvOd) \cite{hardt_equalityOpp_2016} is the average of the differences in TPR and FPR between the unprivileged and privileged groups.
        \begin{displaymath}
            AvOd = \frac{(TPR_{A=1} - TPR_{A=0}) + (FPR_{A=1} - FPR_{A=0})}{2}
        \end{displaymath}
         The Equality of Opportunity Difference (EqOp) \cite{hardt_equalityOpp_2016} is the difference between the TPR of each group.
        \begin{displaymath}
            EqOp = TPR_{A=1} - TPR_{A=0}
        \end{displaymath}

        \subsubsection{Balanced Conditioned Consistency} Based on Consistency \cite{zemel_learning_2013}, Balanced Conditioned Consistency (BCC) \cite{waller_beyondConsistency_2025} compares the prediction $\hat{Y}$ of each individual $\mathbf{x}$ with that of its $k$-nearest neighbors, $\mathit{kNN}(\mathbf{x})$, and gives the sum of the individual consistency scores $c^{k}(x)$ above or equal to a certain threshold $\delta$ divided by the total number $N$ of individuals in the dataset.
        \begin{displaymath}
         BCC = \frac{1}{N} \sum_{i=1}^N v(\mathbf{x}_i) \quad \text{where}\quad v(\mathbf{x}_i) =
         \Biggl\{
              \begin{array}{l@{}l}
                c^{k}(\mathbf{x}_i) = 1 - \big|\hat{y}_i -\frac{1}{k} \sum_{j\in kNN(\mathbf{x}_i)} \hat{y}_j \big|, \quad &\text{if}\ c^{k}(\mathbf{x}_i) \ge \delta\\
                0, \quad &\text{otherwise}
              \end{array} 
        \end{displaymath}
       
        We set $k=5$ and $\delta=0.8$ in our experiments and exclude the sensitive attribute from the features considered for the measure of distance between neighbors.
        
        BCC is thus an individual fairness metric based on similarity \cite{hort_biasClassSurvey_2024} and is bias transforming.
        
        \subsubsection{Generalized Entropy Index} The Generalized Entropy Index (GEI), in the version we use, quantifies unfairness by evaluating how unequally benefits are distributed between individuals \cite{speicher_unifiedMetrics_2018}. 
        The benefit $b_i$ of individual $i$ considers the desirability of receiving an accurate outcome and of receiving a positive one, with $b_i = \hat{y}_i - y_i + 1$.
        \begin{displaymath}
        GEI^\alpha = \frac{1}{N \alpha (\alpha-1)}\sum_{i=1}^N\left[\left(\frac{b_i}{\mu}\right)^\alpha - 1\right]
        \end{displaymath}
        where $\mu$ is the mean benefit and $N$ the number of individuals. $\alpha$ is a constant we set to $2$ in our experiments.
        
        GEI evaluates the overall individual-level unfairness. Like EqOd, it is related to accuracy and is bias preserving as long as the biased label $Y_b$ is considered as the accurate outcome.

\section{Biasing and evaluation framework}\label{sec:framework_sec}
    
    From the fair world framework, it follows that evaluating models on biased data does not lead to an accurate assessment of the model behavior, which is theoretically demonstrated in \cite{favier_biasTheoretical_2023}. In Section \ref{sec:biasVSunbiased}, we empirically confirm that evaluating models through metrics applied on a biased dataset can lead to biased measurements.  
    Once we have established the need for a more accurate evaluation process, we present our biasing and evaluation framework in Section \ref{sec:framework}.

    \subsection{Consequence of biased model evaluation}\label{sec:biasVSunbiased}

        We refer to the measurement of a metric on a biased dataset as \textit{biased evaluation} and also denote that result as \textit{biased}, for example with \textit{biased EqOd} referring to the measurement of EqOd on biased data. We refer to metric measurements on a baseline unbiased datasets as \textit{fair evaluation} and their results as \textit{fair}.
        
\begin{figure}[h]
        \centering
            \begin{minipage}[c]{.33\textwidth}
                \centering
                \includegraphics[width=\textwidth]{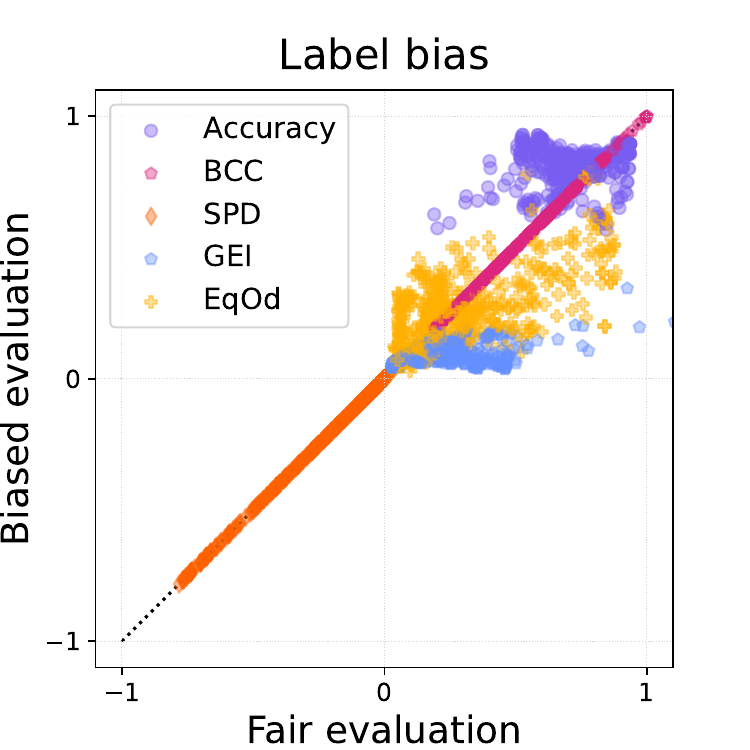}
            \end{minipage}
            \begin{minipage}[c]{.33\textwidth}
                \centering
                \includegraphics[width=\textwidth]{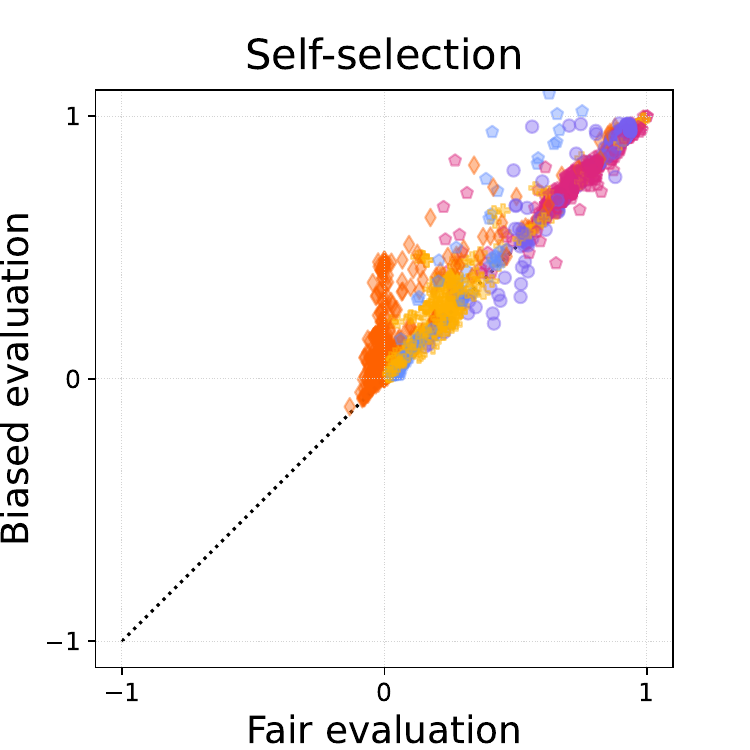}
            \end{minipage}
            \begin{minipage}[c]{.33\textwidth}
                \centering
                \includegraphics[width=\textwidth]{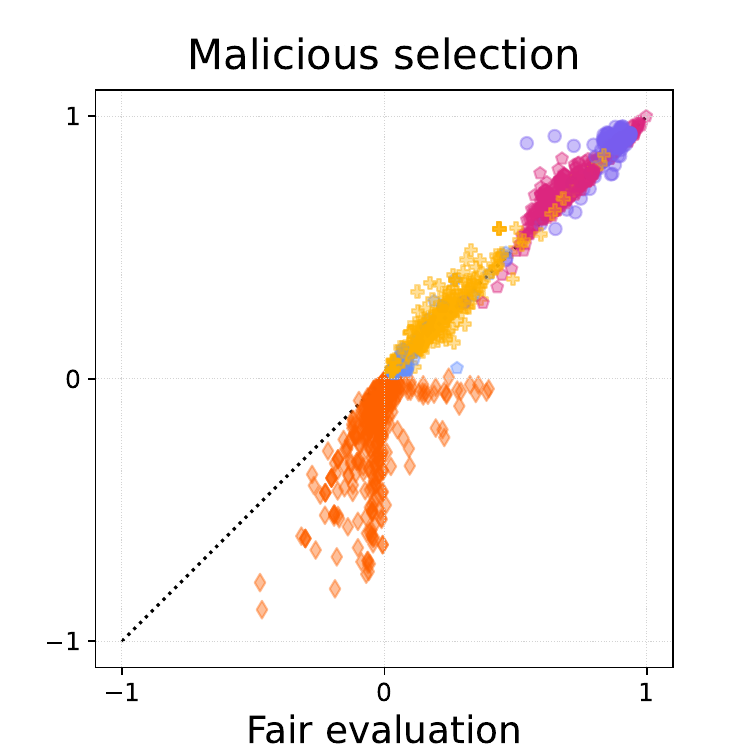}
            \end{minipage}
         \caption{Comparison of metric results evaluated on a biased test set (Biased evaluation) or a fair test set (Fair evaluation) for values in $(-1,1)$. Points that fall outside of the x=y diagonal indicate that the metric measurement is skewed by the bias present in the unfair test set. Are included the results for 4 datasets with bias intensities ranging from 0 to 0.9, 3 fairness-agnostic training algorithms, 8 bias mitigation procedures, and unmitigated models.}
        \label{fig:BvsF}
        \end{figure}

        To analyze the distortion of metrics measurement, we compare the values given by the \textit{fair} and \textit{biased} evaluation of the different metrics and report them in Figure \ref{fig:BvsF}. Each data point corresponds to a specific metric applied to a given model. The x-axis gives the value obtained by computing that metric on the unbiased version of the test set, while the y-axis gives the value as evaluated on the biased version of the test set. 
        A point above the x=y diagonal indicates that the biased evaluation of the metric gives a higher value than its fair evaluation, and reciprocally for the points below the diagonal.
        We report the results for four datasets with bias intensities ranging from 0 to 0.9, three fairness-agnostic training algorithms, eight bias mitigation procedures, and unmitigated models (as presented in Section \ref{sec:expe}). Graphs that include values greater than 1 can be found in Appendix \ref{app:biasVSunbiased}, along with the results for random selection.

        From these graphs, we can make the following observations :
    
        \textbf{Label bias skews the measurement of metrics relying on the ground truth, but does not affect the measurement of metrics based on predicted outcome only.}
        On the first graph, all data points for SPD and BCC are aligned on the x=y diagonal, which indicates that neither of them is affected by label bias. 
        Accuracy, GEI and EqOd all give different results when evaluated on a biased or fair test set, with fair GEI always indicating more discrimination than biased GEI.
        
        \textbf{Selection bias skews the measurement of all five metrics.}
        All subtypes of selection bias affect the measurement of the five metrics, with BCC and EqOd showing smaller variation than the others. More generally, the discrepancy between the fair and biased results caused by selection bias does not become as large as with label bias.
        With self-selection, biased SPD overestimates fairness because the biased test set exhibits an overrepresentation of "deserving" unprivileged individuals. On the other hand, malicious selection leads to underestimated SPD values since the biased test set contains a lower proportion of individuals "deserving" a positive outcome in the unprivileged group than in the privileged one.

        \paragraph{Discussion} As these observations demonstrate, evaluating models on biased data can produce misleading results. Another approach is thus necessary in order to obtain a reliable assessment of model behavior. 
        To analyze the effect of bias and the performance of mitigation methods, we encourage researchers and practitioners to use biased data for training and fair data for evaluation.
        Dual label datasets, which contain both biased and fair labels \cite{defrance_abcfair_2024}, are particularly useful for this. \citet{lenders_studentExtended_2023} introduce Student, which is based on real life data, but such datasets remain very scarce. Fair evaluation can also be done with synthetic datasets for which the bias level can be controlled, like in \cite{blanzeisky_creditscoredata_2021} or \cite{cardoso_benchmarkingframework_2019}. 
        Other options may include the use of census data from periods for which the bias level is different like in \cite{kamiran_tree_2010} or using data augmentation to create fairer test sets as suggested in \cite{sharma_dataAugmentation_2020}, although these alternatives are more likely to suffer from other unwanted biases.
        
        Given the limitations of the existing solutions, we introduce a new approach that allows for the injection of artificial and controlled bias in real-life datasets that we consider to represent a fair baseline. With this novel approach, we can obtain dual-label datasets presenting different levels of a chosen bias type.

    \subsection{Framework description}\label{sec:framework}
    
\begin{figure}[h]
            \includegraphics[alt={The image is fully described in the caption},width=0.75\textwidth]{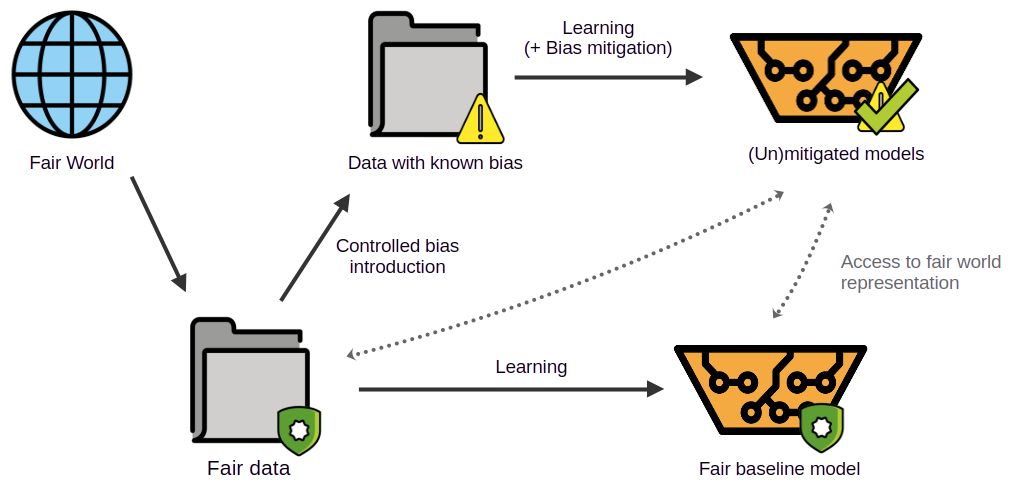}
            \caption{Biasing and evaluation framework. We assume that a given dataset with high fairness level, the \textit{fair data}, accurately represents the fair world. We introduce a specific bias type in this fair data to obtain datasets with known bias. This biased data can be used to train prediction models with or without bias mitigation. The performance and fairness of these models can then be evaluated using the fair data. They can also be compared to baseline models trained on that fair data.}
         \label{fig:biasing_framework}
        \end{figure}

        As seen in the previous section and according to the Fair World Framework, we need access to a ground truth that perfectly captures the fair world in order to accurately evaluate prediction models, which is not obtainable in practice. Furthermore, to study the differentiated effect of distinct bias types, we need to know what biasing process has occurred, i.e., which type of bias has been introduced. 
        Considering these constraints, we propose a biasing and evaluation framework for a fair and context-sensitive evaluation of models and bias mitigation methods, as represented in Figure \ref{fig:biasing_framework}. 
        
        We start with baseline datasets that exhibit a very low level of discrimination with regard to a chosen sensitive attribute, with the assumption that they represent the fair world accurately. More specifically, we assume that the selection of data points is representative of the fair distribution and that every individual has received the label they deserve.
        We then artificially introduce bias in these datasets to represent the distorting procedure leading to the misrepresentative observed data. The bias modeling we use for our experiments is the one described in Section \ref{sec:bias_model} and the framework supports other bias types as well. 
        With this approach, we obtain datasets for which we have both the biased versions, representing different biasing scenarios, and a fair version. 
        We can thus study the impact of each bias type by training models on the biased data, with or without mitigation, and evaluate them on the fair data. We can also create fair baselines models by using the fair dataset for training.
    
        The practical decision of using real-life data that are thus not guaranteed to be completely exempt of bias does not negate the validity of our results. Indeed, it still allows to analyze the difference between a given fair world, represented by the original dataset, and distorted unfair data, represented by the biased versions of the dataset.
        For this article, we have chosen two datasets with a high level of fairness according to Statistical Parity \cite{dwork_fairnessAwareness_2012}, namely Student Performance \cite{cortez_studentDataset_2008} and OULAD \cite{kuzilek_oulad_2017}. If statistical parity holds, we can consider that all groups are similar in the fair world, which corresponds to the We are All Equal (WAE) assumption \cite{friedler_impossibility_2021}.
    
        Finally, let us note that, in this work, we consider the fair world as an ideal situation in which all desired fairness criteria hold and there is no unwanted bias. We do not address the complex and deeply rooted social discussions of how biases arise and what should be considered to be \textit{fair} or not. Instead, our work focuses on understanding the technical tools used in fair ML so they can be integrated in an optimal way to broader fairness interventions strategies.

\section{Experimental setup}\label{sec:expe}

    Our experimental setup\footnote{The implementation of all experiments and additional results are available at \url{https://github.com/Magalii/ControlledBias/tree/BiasComp}
    .} 
    relies on the biasing and evaluation framework introduced in Section \ref{sec:framework}. We describe the methodology in Section \ref{sec:expe_method}, the datasets used in Section \ref{sec:expe_dataset} and the bias mitigation methods analyzed in Section \ref{sec:mitig_methods}.
        
    \subsection{Methodology}\label{sec:expe_method}
    
         We first introduce a specific type of bias in a baseline dataset, considered to be an accurate representation of the fair world. 
        For each biasing procedure presented in Section \ref{sec:bias_model} (\textit{label}, \textit{random selection}, \textit{self-selection} and \textit{malicious selection}), we create 10 datasets with bias intensity going from 0 to 1, which corresponds to the values of parameter $\beta_l$ or $p_u$. Label bias is introduced independently for each bias level, while selection bias is introduced incrementally, meaning all individuals undersampled at a certain bias level are also absent from all higher bias levels. 

\begin{figure}[h]
        \includegraphics[alt={Fair data is transformed into biased data trough a biasing procedure. The biased data is either used directly in the learning algorithm or preprocessed so the debiased data may be used. The models are either used directly or can undergo post-processing, which makes use of biased data. The (de)biased models are then given a subset of the fair data to make predictions. The same fair data is used for the evaluation metrics that give performance values.},width=\textwidth]{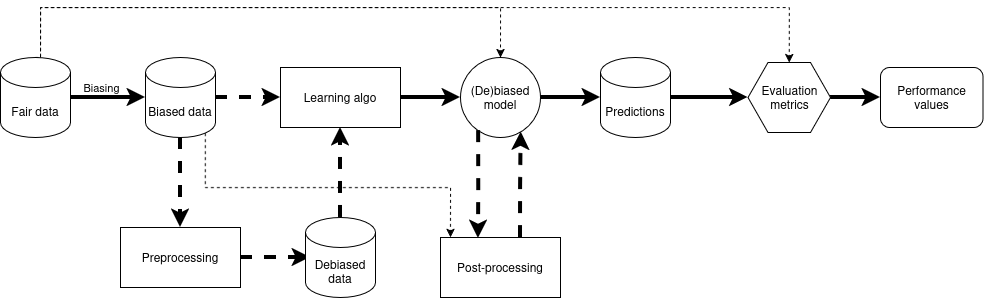}
        \caption{Experiment pipeline representing data biasing, model training (with or without bias mitigation), and evaluation of models on fair data. Thin dashed arrows indicate where each type of data is used. Bold dashed arrows indicate an optional path. We do not cumulate the use of pre- and pot-processing methods.}
     \label{fig:expe_pipeline}
    \end{figure}

        We then train different models on each of those datasets, using either fairness-agnostic training or applying one of the bias mitigation method studied (see Section \ref{sec:mitig_methods}). The fairness-agnostic models trained on the original unbiased data, with $\beta_l=0$ and $p_u=0$, are considered as the baseline for fair models. Those trained on the different biased datasets are used as reference for biased models.
        The mitigated models are obtained by applying either a pre-processing method on the biased dataset before fairness-agnostic training or a post-processing method on fairness-agnostic models. The experiment pipeline is represented in Figure \ref{fig:expe_pipeline} for the evaluation of models on fair data. We also perform model evaluation with the biased data to analyze the impact of biases on performance values.

        We consider three fairness agnostic models : a multi-layer perceptron neural network (maximum 1500 iterations), a classification tree (maximum depth of 6), and random forest (RF) (maximum depth of 6, minimum 10 samples required for splits and for leaf nodes). 
        We use the \texttt{scikit-learn} \cite{sklearn} implementation of those algorithms with default parameters except for the ones specified above. For bias mitigation methods, we use the AIF360 \cite{bellamy2018aif360} implementation, apart for massaging that we implemented directly.
    
        We split our datasets into 10 folds for scenarios with label bias, and 5 folds for selection bias to unsure a sufficient number of entries in the test set despite the undersampling. We reserved one fold for validation (used in post-processing only\footnote{The validation sets are required to modify the trained models according to the post-processing techniques used. They are however not needed for the pre-processing methods.}) and one for testing. The same partitioning is used across the different bias intensities, meaning that each model can be evaluated using either a biased or unbiased version of its test set. 
        We report the mean value and standard deviation of the metrics presented in Section \ref{sec:eval} over the 10 or 5 iterations.
        
    \subsection{Datasets}\label{sec:expe_dataset}
    
        The three baseline datasets we use to represent the fair world are described below. The technical characteristics of each dataset are presented in Table \ref{tab:datasets}.
        
        \subsubsection{Student}
        Also known as Student Performance or Student Alcohol Consumption, \textbf{Student} \cite{cortez_studentDataset_2008} contains data about students following a Portuguese course, with demographic, social, and educational information. The prediction task is to determine if the final grade of the student is a passing grade (\textit{G3} $\geq 10$). The positive label is thus \textit{pass} and the negative one \textit{fail}. The original numerical grades are used as the scores for label biasing and the noise intensity $\beta_n$ is set to $0.1$. In addition to the original dataset, we create a \textbf{StudentBalanced} version by randomly removing girls receiving positive outcome in order to obtain a dataset that is perfectly balanced group-wise, i.e, same size and number of positive labels for each group.
        
        \subsubsection{Open University Learning Analytics Dataset (OULAD)} 
        The OULAD \cite{kuzilek_oulad_2017} dataset contains records of students following online classes. The Bayesian network analysis in \cite{le_quy_surveyDatasets_2022} found that \textit{gender} has a very low effect on the \textit{final\_result} outcome. We extract two subsets of the larger OULAD dataset to create our baselines. We construct \textbf{OULADstem} with the records of students following a specific STEM course (\textit{code\_module}=`FFF') and \textbf{OULADsocial} with students registered to a specific course in social science (\textit{code\_module}=`BBB'). For both OULAD subsets, we have removed instances with missing values and the first occurrence of students who were duplicated (students who repeated the course over different semesters). We also use the same feature selection and prediction task in both cases. 
        
        We use features of the studentInfo file and compute additional ones as described in \cite{riazy_fairness_2020}. This gives us the following list of 17 features, with the 9 features from studentInfo listed first :
        \textit{code\_presentation}, \textit{gender}, \textit{region}, \textit{highest\_education}, \textit{imd\_band}, \textit{age\_band}, \textit{num\_of\_prev\_attempts}, \textit{studied\_credits}, \textit{disability}, \textit{num\_CMA}, \textit{num\_TMA}, \textit{login\_day}, \textit{num\_logins}, \textit{forumng}, \textit{glossary}, \textit{homepage}, and \textit{resource}.
        
        Our task is to predict the \textit{final\_result} target. The positive binary label corresponds to the values \textit{Pass} or \textit{Distinction} and the negative one to \textit{Fail} or \textit{Withdrawn}. We assigned numerical values to each of those nominal ones to obtain the scores for label biasing. Since these scores are less granular than for Student, we set the noise intensity $\beta_n$ to $0.2$.

    \begin{table}[h]
        \begin{tabular}{|c|c|c|c|c|c|c|c|c|}
        \hline
        Dataset  & Size & Target & Sens. attr. & Unpriv. group & Unpriv. prop. & SPD & BCC & $P(y=1)$ \\\hline
        Student & 649 & \textit{G3} (final grade) & \textit{sex} & Boys & $40.9\%$ & -0.057 & 0.721 & 0.846  \\ 
        StudentBalanced & 532 & \textit{G3} (final grade) & \textit{sex} & Boys & $50\%$ & 0.0 & 0.680 & 0.812  \\ 
        OULADstem & 7040 & \textit{final\_result} & \textit{gender} & Men & $81.7\%$ & -0.025 & 0.662 & 0.482 \\ 
        OULADsocial  & 7632 & \textit{final\_result} & \textit{gender} & Men & $11.6\%$ & -0.026 & 0.596 & 0.487 \\ 
        \hline
        \end{tabular}
        \caption{Overview of datasets characteristics}
        \label{tab:datasets}
    \end{table}

    \subsection{Bias mitigation methods}\label{sec:mitig_methods}
    
        We perform our experiments with eight pre- and post- processing methods with different characteristics. We do not include in-processing techniques, as coupling the optimization of model reliability and a fairness metric can force models to exhibit unwanted and unfair behavior such as the selection of worst individuals from the minority group when better ones from the same minority should be selected \cite{favier_cherry_2025}. We thus chose to focus our work on mitigation techniques that do not suffer from this problem.

        \subsubsection{Reweighing} Introduced in \cite{kamiran_preprocessing_2012}, reweighing is a pre-processing method that assigns weights to instances of the dataset based on the label and the sensitive attribute such that they satisfy statistical parity in the weighted dataset. As shown in \cite{favier_biasTheoretical_2023}, the reweighing process can be seen as opposing that of selection bias. We thus expect this method to perform well in such scenarios, with the limitation that enforcing statistical parity on biased data does not allow to retrieve the original fair distribution \cite{favier_biasTheoretical_2023}.

        \subsubsection{Massaging} Another pre-processing method in \cite{kamiran_preprocessing_2012}, massaging changes the labels of instances close to the decision boundary in order to satisfy statistical parity. An additional ranker is used to identify the members of the unprivileged (resp. advantaged) group whose label should be changed to a positive (resp. negative) one. We expect this method to have a positive influence on label bias, but to be unable to retrieve the fair distribution in case of selection bias because of the change in labels and the reliance on statistical parity that is sensitive to that bias type.
        
        \subsubsection{Fairness Through Unawareness (FTU)} Also known as \textit{blinding}, FTU \cite{gajane_formalizingFairness_2018} consists in the removal of the sensitive attribute from the training set. Early works have shown FTU to be insufficient when there are correlations between the sensitive attribute and other features \cite{pedreshi_discrimination-aware_2008}, but \cite{favier_biasTheoretical_2023} presents it as a viable procedure when statistical parity holds in the fair world.

        \subsubsection{Equalized Odds Postprocessing (EOP)} The post-processing technique EOP \cite{hardt_equalityOpp_2016} changes predicted labels with a probability determined by the resolution of a linear program that aims to satisfy equalized odds and optimize accuracy. As those two criteria depend on data labels, we expect EOP to be more vulnerable to label bias. It is worth noting that the AIF360 implementation used in our experiments relies on binary predicted labels rather than on prediction probabilities.
        
        \subsubsection{Calibrated Equalized Odds Postprocessing (CEO)} 
        Building on \cite{hardt_equalityOpp_2016}, the post-processing method CEO \cite{pleiss_fairnessCalib_2017} aims to both preserve calibration and achieve a relaxed notion of equalized odds, similar in idea to average odds, that is based on probabilistic scores and compatible with the calibration constraint\footnote{As per the impossibility result in \cite{kleinberg_inherentTradeoff_2017}, two classifiers $h_{priv}$ and $h_{unpriv}$ for groups $G_{priv}$ and $G_{unpriv}$ can only satisfy both equalized odds and calibration if and only if $h_{priv}$ and $h_{unpriv}$ are perfect predictors or $G_{priv}$ and $G_{unpriv}$ have the same base rate ($P(\hat{y}=1)$).}. To do so, it withholds predictive information for randomly chosen individuals who then receive the mean probability of the group. We also expect CEO to be more vulnerable to label bias.
        
        \subsubsection{Reject Option Classifier (ROC)} This post-processing technique changes the labels of individuals with low prediction confidence to improve a given fairness criterion \cite{kamiran_decision-theory_2012}. The label are only changed from positive to negative for privileged individuals and from negative to positive for unprivileged individuals. ROC can be used with different fairness criteria. We perform experiments with SPD, EqOp, and AvOd, noting the different versions respectively ROC-SPD, ROC-EqOp and ROC-AvOd. We expect ROC-SPD to perform similarly as massaging, and ROC-EqOp and -AvOd to retain some bias in the model since they rely on bias preserving metrics.

\section{Results}\label{sec:results}
    \newcounter{resultSubsec}
    \setcounter{resultSubsec}{0}
    \newcounter{observ}[resultSubsec]
    
    We present here the results of our different experiments. We first relate the fair measurement of evaluation metrics with the biased one in Section \ref{sec:res_biased_eval}. We then present the impact of the different bias types on fairness-agnostic models in Section \ref{sec:res_model} and their influence on the effectiveness of mitigation methods in Section \ref{sec:res_mitig}.

    The results displayed were obtained with RF as the fairness-agnostic algorithm because they were the most accurate and stable. Results for the decision trees and the neural network can be found in the experiment repository\footnote{The implementation of all experiments and additional results are available at \url{https://github.com/Magalii/ControlledBias/tree/BiasComp}.}.

    \subsection{Relating fair results to biased measurements}\label{sec:res_biased_eval}
    \stepcounter{resultSubsec}

        Since our biasing and evaluation framework allows to compare the results of a fair evaluation with a biased one, we can analyze both types of measurement to highlight potential relationships between them.

        \begin{figure}[h]
        \centering
            \begin{minipage}[c]{.36\textwidth}
                \centering
                \includegraphics[width=\textwidth]{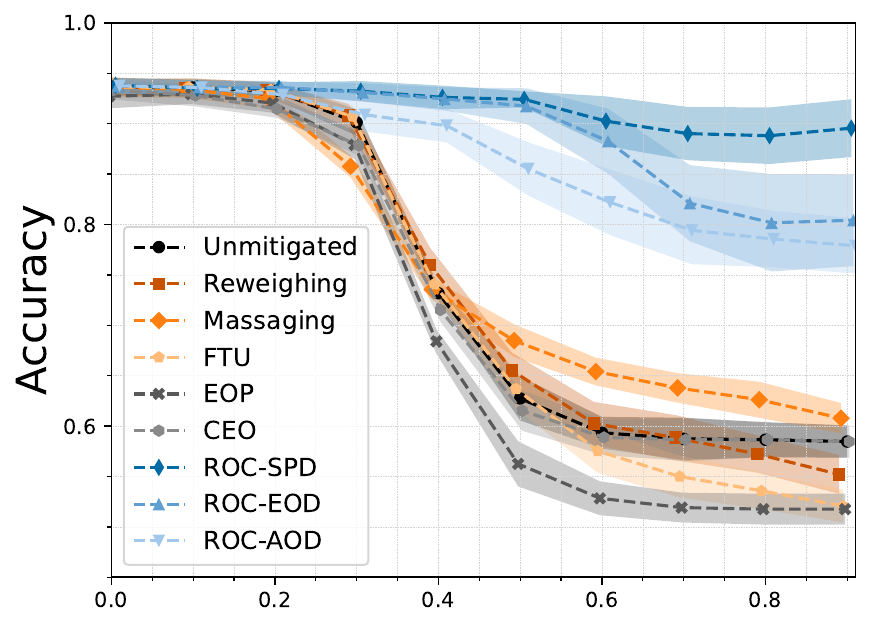}
            \end{minipage}
            \begin{minipage}[c]{.36\textwidth}
                \centering
                \includegraphics[width=\textwidth]{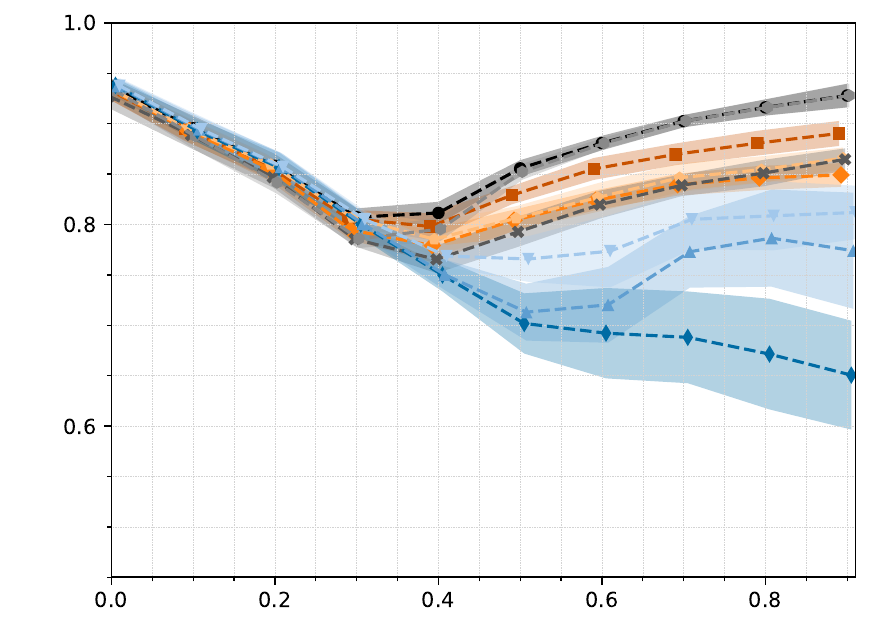}
            \end{minipage}
            \\
            \begin{minipage}[c]{.36\textwidth}
                \centering
                \includegraphics[width=\textwidth]{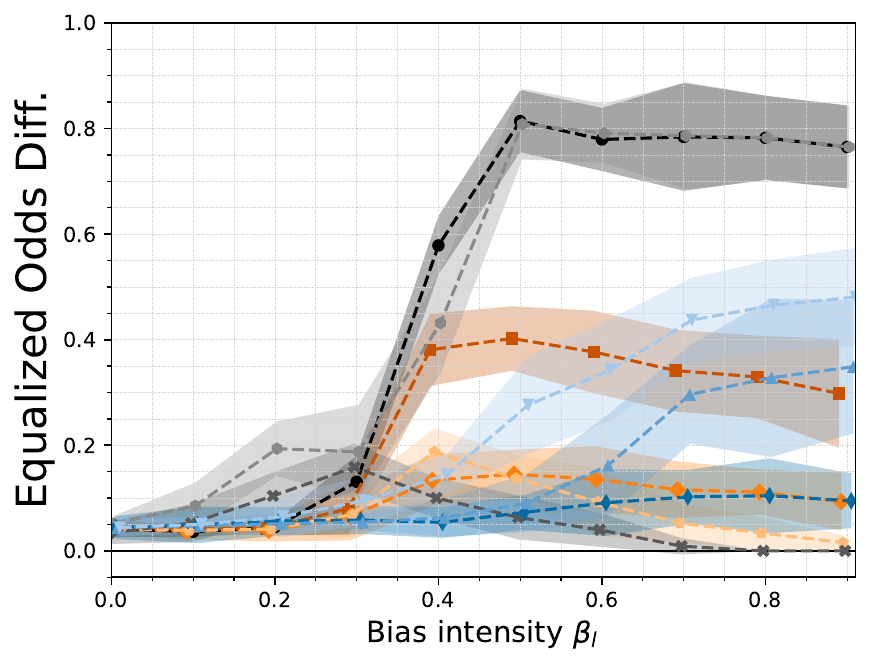}
                \subcaption{Fair evaluation}
            \end{minipage}
            \begin{minipage}[c]{.36\textwidth}
                \centering
                \includegraphics[width=\textwidth]{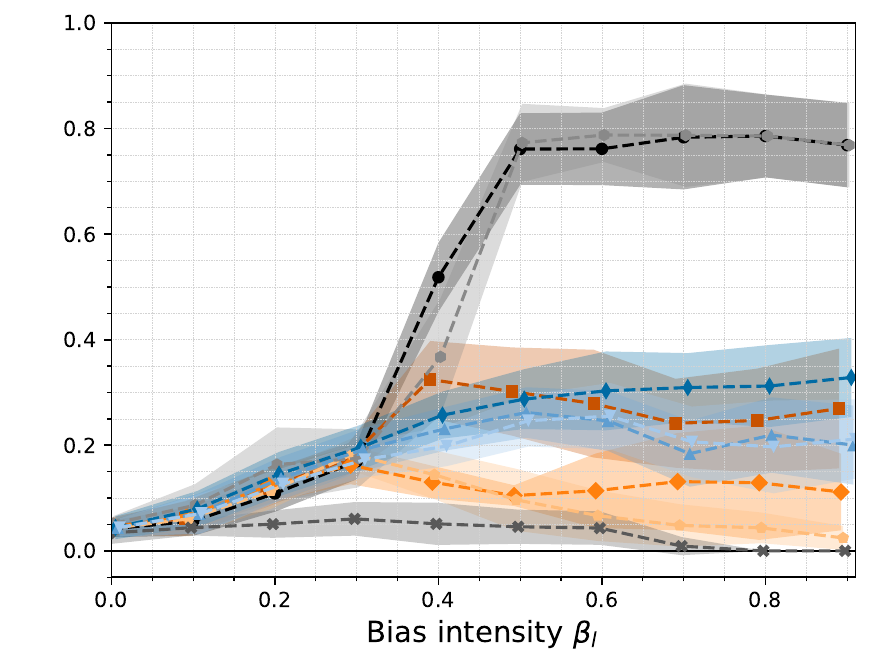}
                \subcaption{Biased evaluation}
            \end{minipage}
        \caption{Fair and biased evaluation of Accuracy and EqOd for increasing levels of \textbf{label bias} injected in OULADstem training sets}
        \label{fig:trad_label_stem}
        \end{figure} 
        
\begin{figure}[h]
        \centering
            \begin{minipage}[c]{.36\textwidth}
                \centering
                \includegraphics[width=\textwidth]{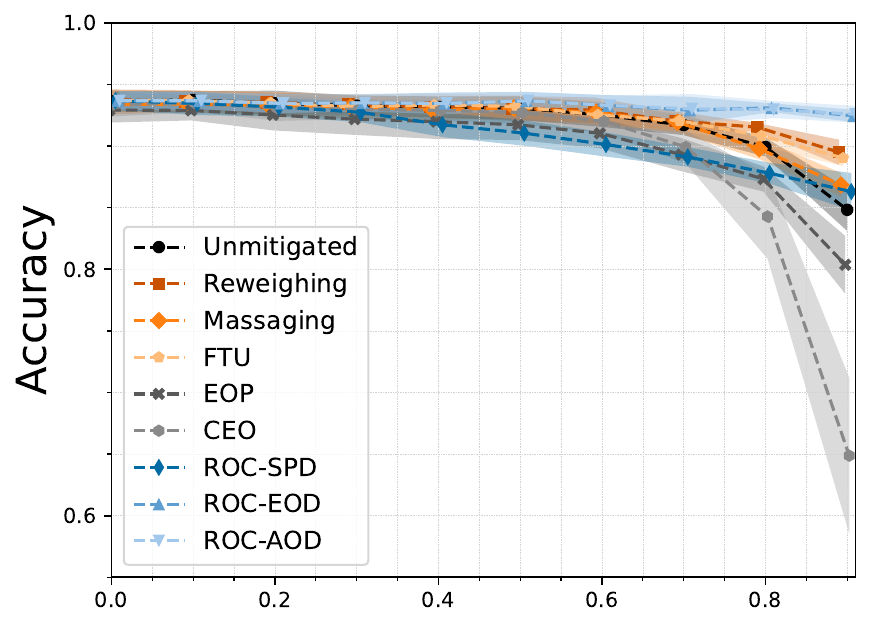}
            \end{minipage}
            \begin{minipage}[c]{.36\textwidth}
                \centering
                \includegraphics[width=\textwidth]{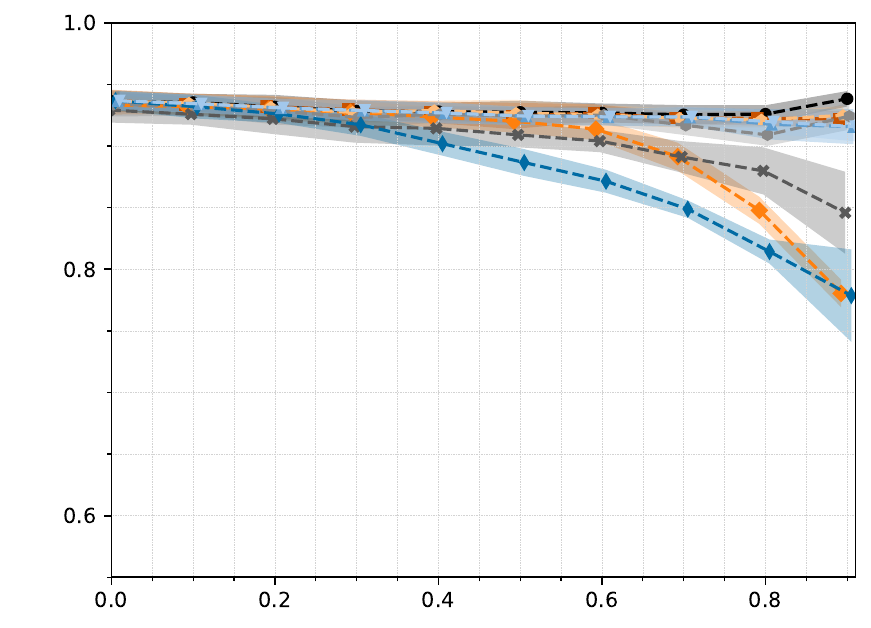}
            \end{minipage}
            \\
            \begin{minipage}[c]{.36\textwidth}
                \centering
                \includegraphics[width=\textwidth]{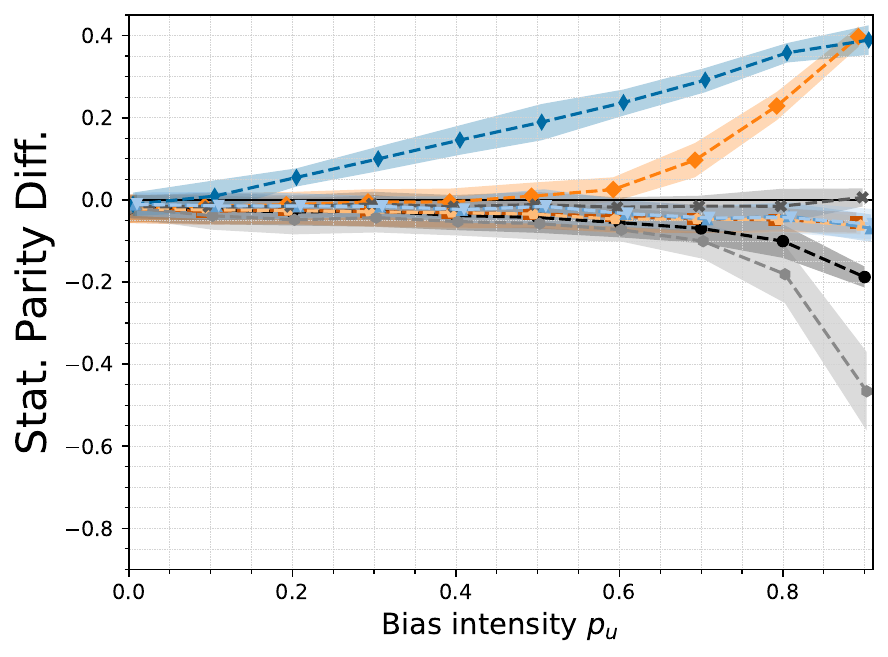}
                \subcaption{Fair evaluation}
            \end{minipage}
            \begin{minipage}[c]{.36\textwidth}
                \centering
                \includegraphics[width=\textwidth]{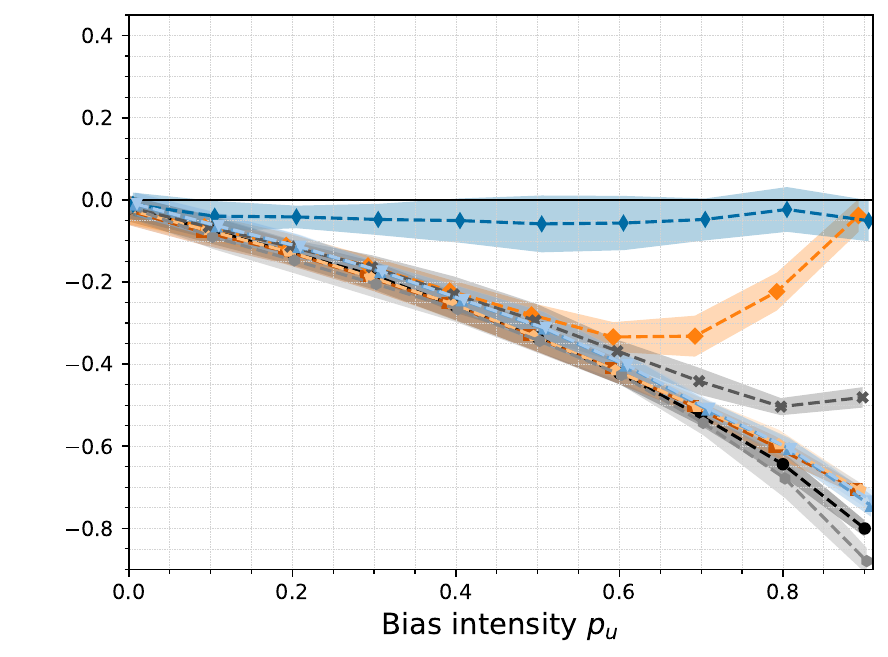}
                \subcaption{Biased evaluation}                
            \end{minipage}
        \caption{Fair and biased evaluation of Accuracy and SPD for increasing levels of \textbf{malicious selection} injected in OULADstem training sets.}
        \label{fig:trad_selectD_stem}
        \end{figure}

        Figures \ref{fig:trad_label_stem} and \ref{fig:trad_selectD_stem} display the values of metrics evaluated on fair data in the left graphs (Fair evaluation) and on biased data in the right graphs (Biased evaluation), and so for different (un)mitigated models trained on OULADstem presenting increasing levels of label bias or malicious selections.
        The top graphs report the accuracy, while the bottom graphs display the fairness measures. For label bias, we report EqOd, as SPD and BCC are not affected by that bias type (see Section \ref{sec:biasVSunbiased}) and GEI is more closely related to accuracy. For selection bias, we report SPD, which is the metric most vulnerable to it. 
        The observations obtained with malicious selection generalize to self-selection and the conclusions drawn hold for all datasets. These results can be seen in Appendix \ref{app:biasVSunbiased}. We omit random selection from this analysis as its impact is negligible (see Section \ref{sec:res_model}).
                
        \paragraph{Label bias} Figure \ref{fig:trad_label_stem} displays the results for the introduction of label bias in OULADstem, which leads to an increasing number of unprivileged individuals to be wrongfully assigned a negative label.

        \stepcounter{observ}
        \underline{Observation \arabic{resultSubsec}.\arabic{observ}} : \textbf{To perform well on fair data according to metrics relying on ground truth, models must have poor results for these metrics when evaluated on test data affected by label bias.}
        This is what we see for models mitigated with ROC-SPD and ROC-EqOp. They are the two best performing methods according to the fair evaluation, although the biased evaluation suggests that they perform poorly with regard to both correcting unfairness and maintaining accuracy. On the other hand, massaging, FTU and EOP give good results according to the biased evaluation, but the fair evaluation reveals that all these methods lead to a great loss of accuracy. They thus fail to obtain predictions close to that of the fair world.

        \paragraph{Selection  bias}
        The influence of malicious selection can be seen in Figure \ref{fig:trad_selectD_stem}. The results for self-selection are visible in Appendix \ref{app:biasVSunbiased}. 
        
        \stepcounter{observ}
        \underline{Observation \arabic{resultSubsec}.\arabic{observ}} : \textbf{In case of selection bias, the methods that perform best on fair data are those for which the biased accuracy and SPD are closest to the unmitigated biased values.} The unfairness measured on the biased test set is thus caused by fair predictions on a test set affected by selection bias. On the other hand, a biased SPD value close to zero corresponds to unfair models. 
        These results indicate that selection bias has a more problematic effect on model evaluation than for their training.

        In Figure \ref{fig:trad_selectD_stem}, the fair evaluation shows that reweighting, FTU, ROC-EOD and ROC-AOD all improve both fairness and accuracy, while their biased evaluation indicate a reduction in accuracy and no improvement of SPD. Conversely, massaging and ROC-SPD introduce bias against the privileged group in order to obtain an improvement of the biased SPD value.

        \paragraph{Discussion}
            These results highlight some relationships between the fair and biased measurements of evaluation metrics, but do not provide sufficient pointers to infer the behavior of models or fairness interventions from biased metric measurement. It thus remains necessary to use fair test data to assess the impact of bias on model performance and to evaluate the efficiency of bias mitigation methods.

    \subsection{Bias impact on models}\label{sec:res_model}
    \stepcounter{resultSubsec}
    
        We analyze in this section the impact of different bias types on classification models in the absence of any bias mitigation. The models are trained on datasets presenting increasing levels of bias and evaluated on the unbiased test set representing the fair world. Our results can be related to \cite{ceccon_dataBiasProfile_2025}, which includes a study on the effect of label bias (modeled by flipping labels of randomly chosen unprivileged individuals from positive to negative) and random selection on the group fairness of models. Their results confirm the negative impact of label bias on model fairness, but the authors argue that the influence of undersampling has been overemphasized. Furthermore, since including the unprivileged group in the training data is often unlikely to meaningfully improve fairness according to their results and since high-quality data from that group is often scarce, the authors suggest to prioritize its use for model evaluation instead of training. 
        We extend and discuss their interesting findings by analyzing different manifestations of label and selection bias using diverse evaluation metrics.
   
        The graphs in this section show how RF models are impacted by the injection of different bias type in their training data. In addition to traditional evaluation metrics, we also report the reliance of models on the sensitive attribute ("Sens. attr. usage"), characterized by the proportion of trees in the random forest that use the sensitive attribute as a classification feature. To allow for more interpretability, we display the between group difference in false negative and false positive rates (FNR and FPR diff.) rather than a compound metric like EqOd. Regarding model correctness, balanced accuracy is reported and gives results consistent with ROC-AUC.

        \subsubsection{Label bias} 
\begin{figure}[h]
            \centering
                \begin{minipage}[c]{.335\textwidth}
                    \centering
                    \includegraphics[width=\textwidth]{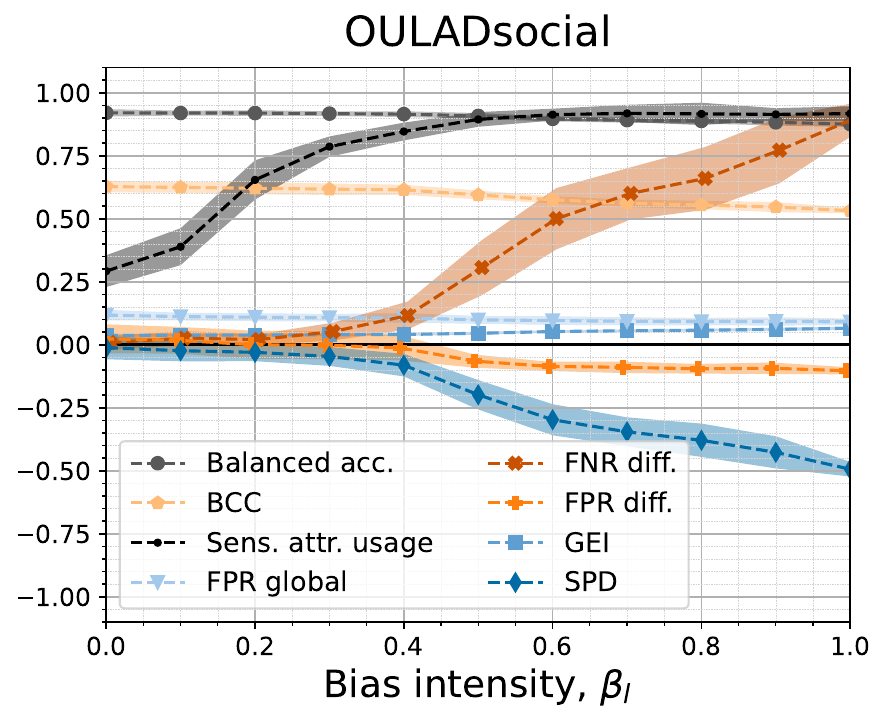}
                \end{minipage}
                \begin{minipage}[c]{.32\textwidth}
                    \centering
                    \includegraphics[width=\textwidth]{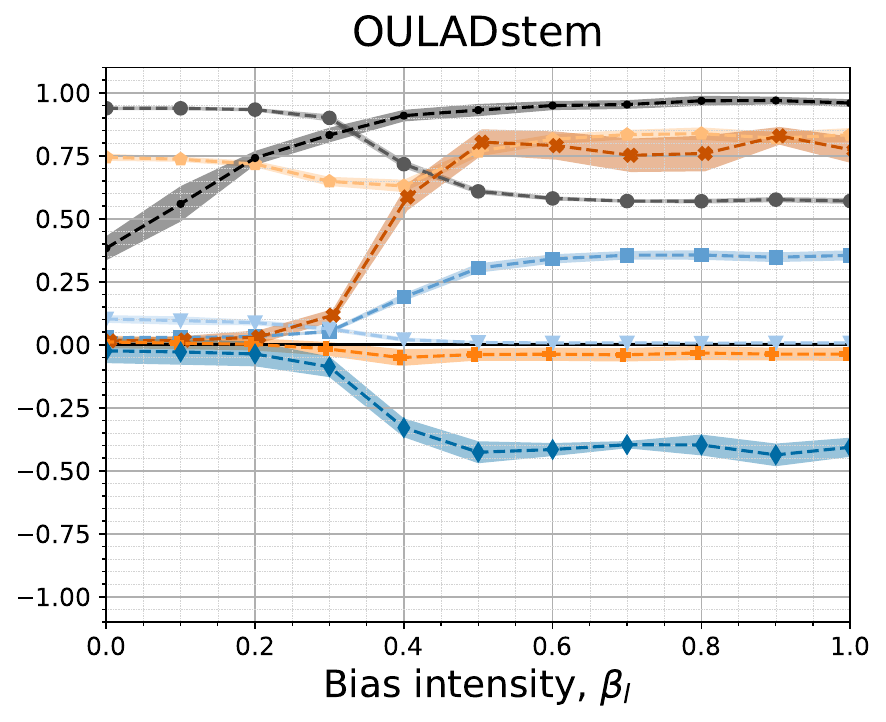}
                \end{minipage}
                \begin{minipage}[c]{.32\textwidth}
                    \centering
                    \includegraphics[width=\textwidth]{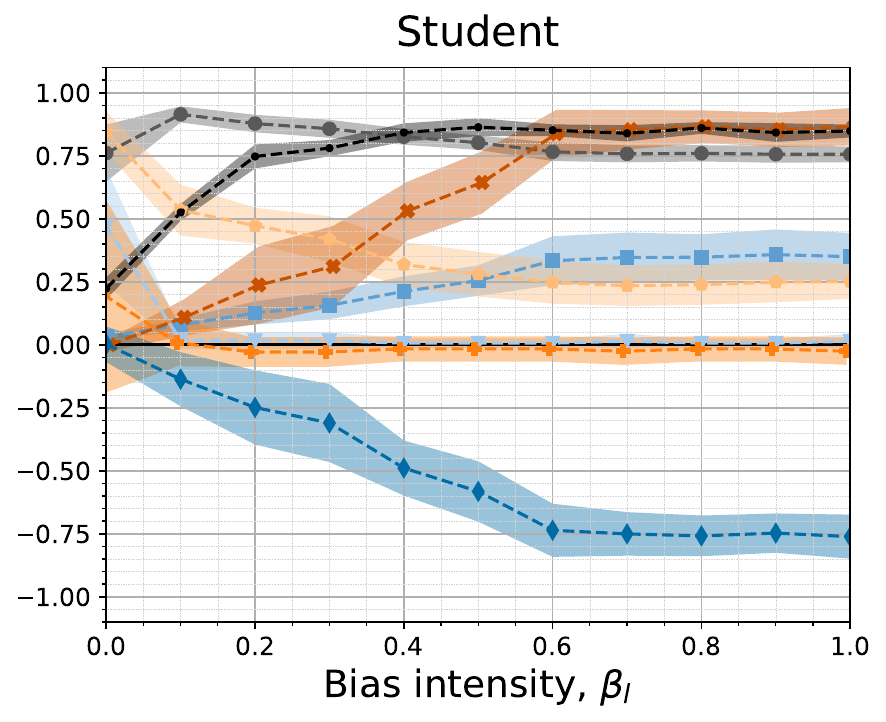}
                \end{minipage}
            \caption{Evolution of RF models performance when trained on datasets presenting increasing levels of \textbf{label bias}, as evaluated on unbiased data.}
            \label{fig:impact_label}
            \end{figure}

            The result regarding the introduction of label bias are visible in Figure \ref{fig:impact_label}.

            \stepcounter{observ}
            \underline{Observation \arabic{resultSubsec}.\arabic{observ}} : \textbf{Label bias has a very detrimental effect on accuracy, group fairness, and individual fairness.} 
            This can be expected since that distortion directly impacts the relationship between the features and the correct label. 
            
            In OULADsocial, the effect on globally aggregated metrics is minimal because the biasing penalty is only applied on the unprivileged group, which makes out less than $12\%$ of the dataset. Its negative effect is however visible in group-dependent metrics. We expect the impact of label bias on this type of metrics to be less dependent on group ratios if it is applied to all groups, for example with a bonus for privileged individuals in addition to the penalty received by unprivileged ones.

        \subsubsection{Selection bias}

            As we will see in this section, the effect of selection bias is much more limited than that of label bias in many situations. We explore the factors leading to the small effect of undersampling by performing our experiments under different circumstances.
            \begin{itemize}
                \item As our base case, we train models with increasingly biased versions of OULADsocial, OULADstem, and Student.
                \item To analyze the effect of base rate imbalance, we inject selection bias in StudentBalanced, which has an SPD of exactly $0$ unlike the regular Student dataset. 
                \item To consider the complexity of the learning task, we train models on datasets from which the most relevant features have been removed. These datasets, \textit{OULADsocial-Complex} and \textit{OULADstem-Complex}, contain the same instances as the regular OULAD datasets, but only the 9 features from the studentInfo file (see Section \ref{sec:expe_dataset}).
            \end{itemize}

\begin{figure}[h]
            \centering
                \begin{minipage}[c]{.335\textwidth}
                    \centering
                    \includegraphics[width=\textwidth]{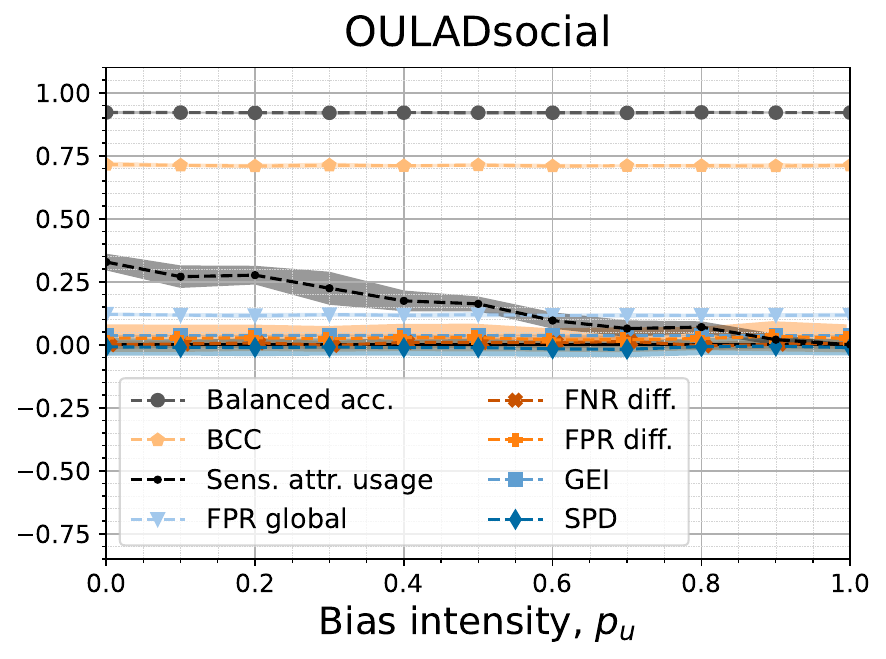}
                \end{minipage}
                \begin{minipage}[c]{.32\textwidth}
                    \centering
                    \includegraphics[width=\textwidth]{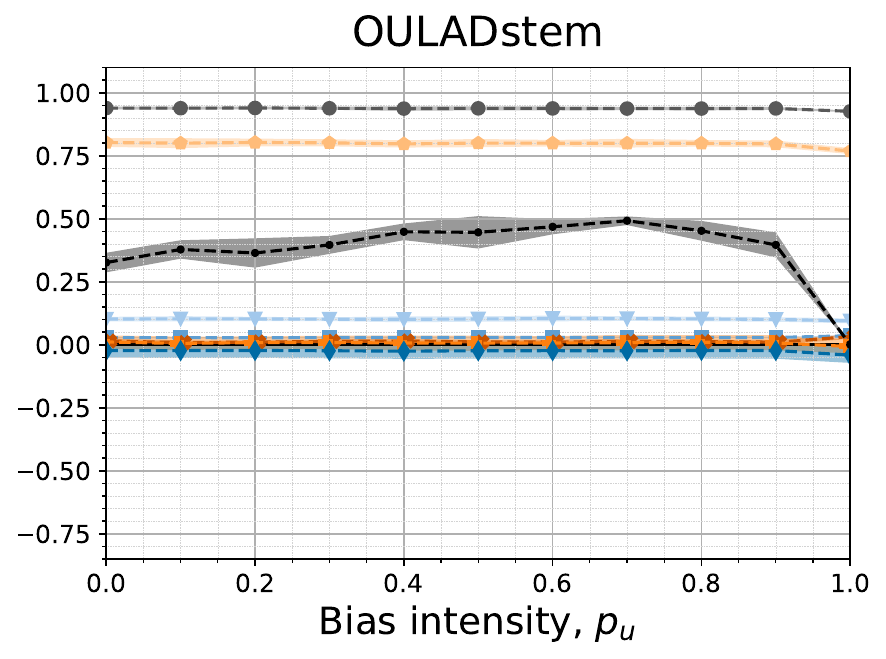}
                \end{minipage}
                \begin{minipage}[c]{.32\textwidth}
                    \centering
                    \includegraphics[width=\textwidth]{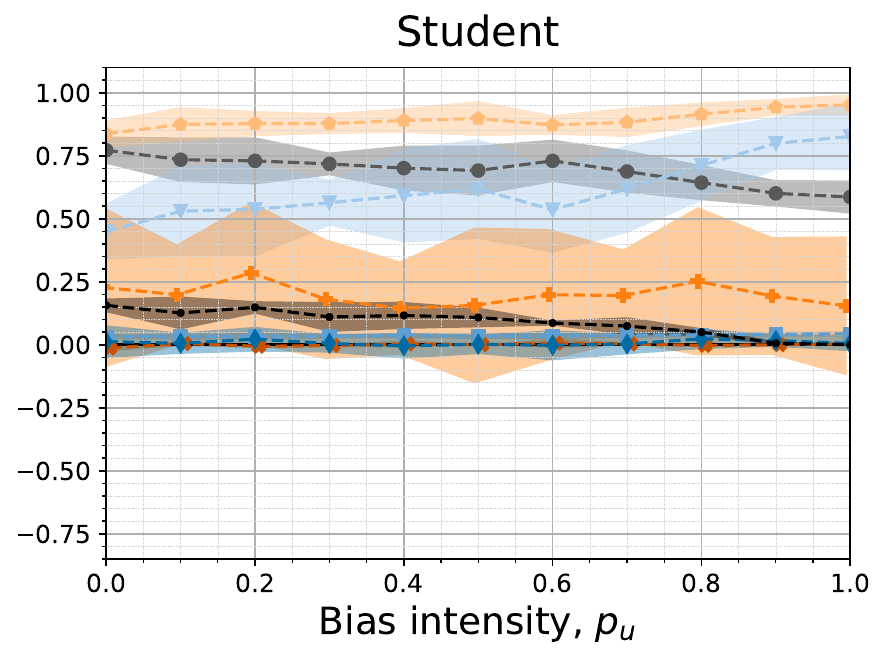}
                \end{minipage}
                \\
                \begin{minipage}[c]{.335\textwidth}
                    \centering
                    \includegraphics[width=\textwidth]{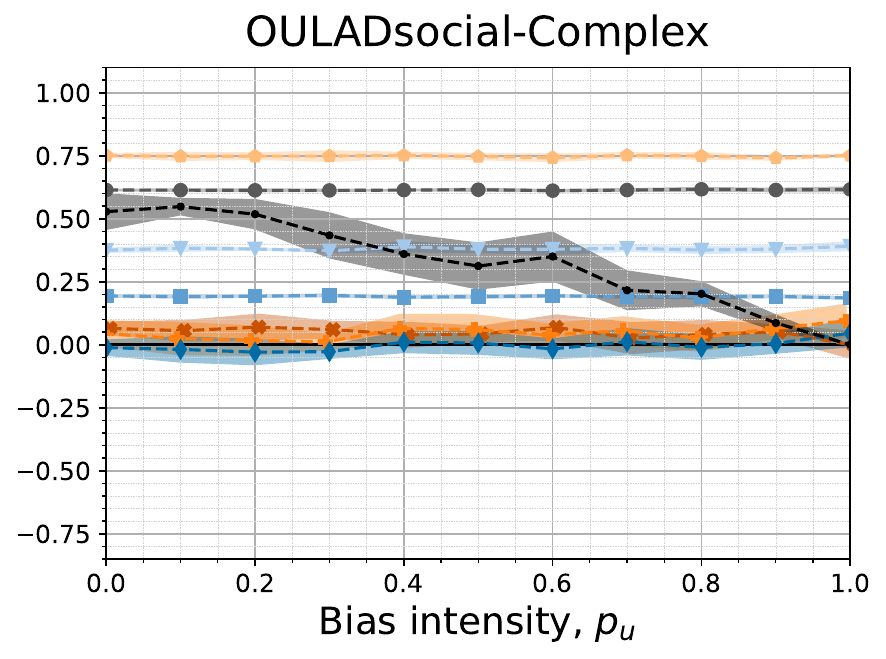}
                \end{minipage}
                \begin{minipage}[c]{.32\textwidth}
                    \centering
                    \includegraphics[width=\textwidth]{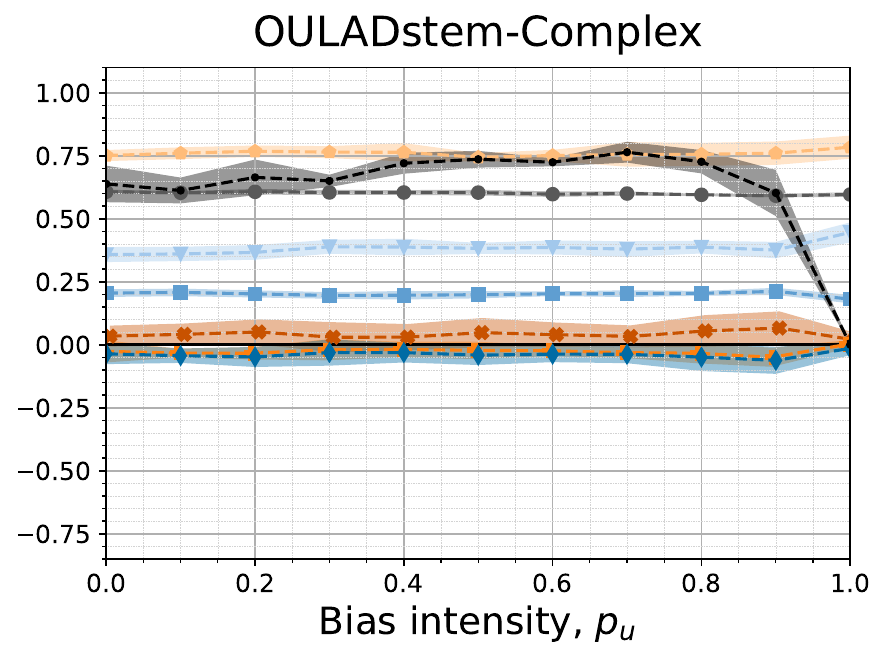}
                \end{minipage}
                \begin{minipage}[c]{.32\textwidth}
                    \centering
                    \includegraphics[width=\textwidth]{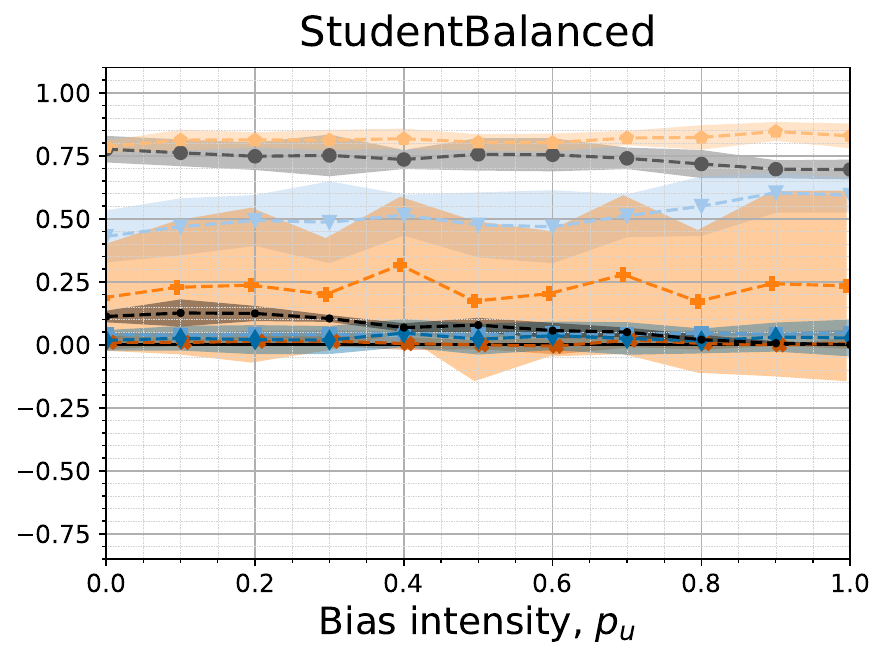}
                \end{minipage}
            \caption{Evolution of RF models performance when trained on datasets presenting increasing levels of \textbf{random selection}, as evaluated on unbiased data. $p_u = 1$ corresponds to the complete removal of the unprivileged group from the training data.}
            \label{fig:impact_randomSelect}
            \end{figure}

            \paragraph{Random selection}
            The results for random selection are presented in Figure \ref{fig:impact_randomSelect}. Regarding the reliance of models on the sensitive attribute, it is highest when the two groups are balanced in size and decreases as they become more imbalanced.

            \stepcounter{observ}
            \underline{Observation \arabic{resultSubsec}.\arabic{observ}} : \textbf{Under the WAE assumption, random selection on its own has no significant effect beyond the reduction of the dataset size, regardless of learning task complexity.}
            This is explained by the fact random selection does not change the within-group distributions. This observation is supported by the following results (Fig. \ref{fig:impact_randomSelect}):
            \begin{itemize}
                \item We see a negligible impact for OULADstem and OULADsocial, which are large datasets with a sufficiently low SPD (WAE assumption holds, see Section \ref{sec:framework}). Although still limited, the impact on accuracy is more important for Student, which is much smaller and has a higher between-group difference.
                \item The models trained on StudentBalanced ($SPD=0$) maintain (balanced) accuracy and FPR values closer to the fair baseline than those trained on Student, indicating that a lower SPD value in the original dataset leads to a lower effect of random selection. Despite the SPD value of Student only being $-0.057$, undersampling the underprivileged group reduces the overall base rate.
                The low fairness decrease that remains visible for StudentBalanced is due to the reduction in dataset size, which causes the model to predict the majority class more often. The results for the dataset size reduction can be found in Appendix \ref{app:bias_on_model}.
                \item The impact of random selection is negligible for OULADsocial-Complex and OULADstem-Complex even though they contain less predictive features, which makes for a more complex learning task.
            \end{itemize}
            
            \stepcounter{observ}
            \underline{Observation \arabic{resultSubsec}.\arabic{observ}} : \textbf{Random selection can exacerbate the effect of other bias types present in the dataset.} The small SPD disparity in Student can be seen as a form of historical bias (see \cite{baumann_biasOnDemand_2023}) which leads girls to perform better in school than boys. The effect of this bias on model accuracy, even if it was originally very low, is amplified by the random selection, as visible in Figure \ref{fig:impact_randomSelect}.

\begin{figure}[h]
            \centering
                \begin{minipage}[c]{.335\textwidth}
                    \centering
                    \includegraphics[width=\textwidth]{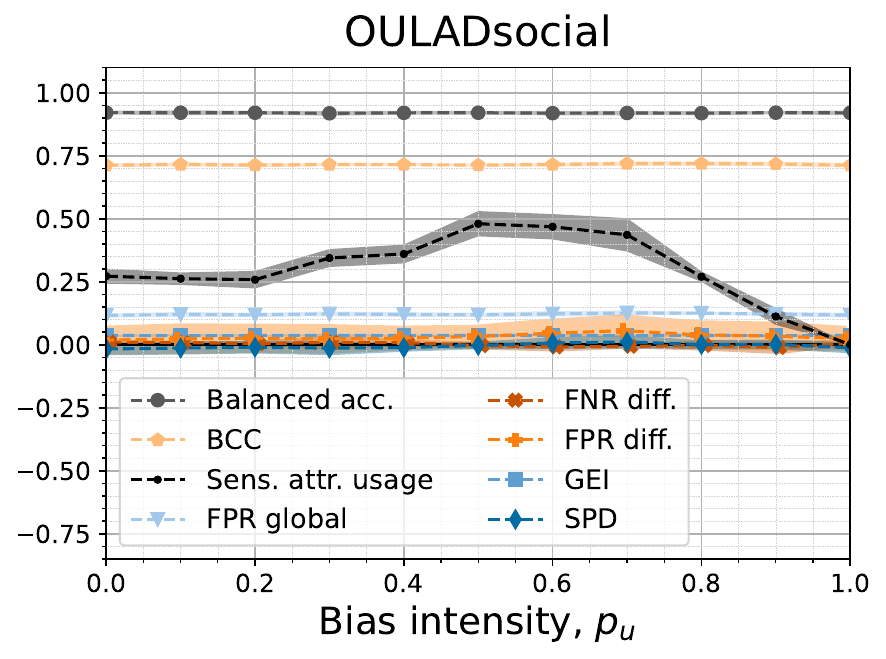}
                \end{minipage}
                \begin{minipage}[c]{.32\textwidth}
                    \centering
                    \includegraphics[width=\textwidth]{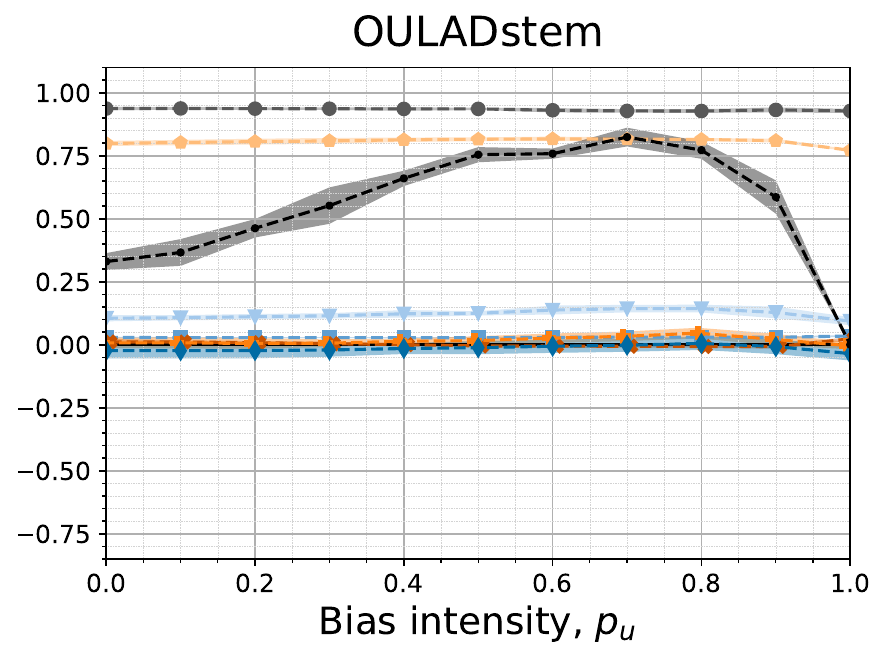}
                \end{minipage}
                \begin{minipage}[c]{.32\textwidth}
                    \centering
                    \includegraphics[width=\textwidth]{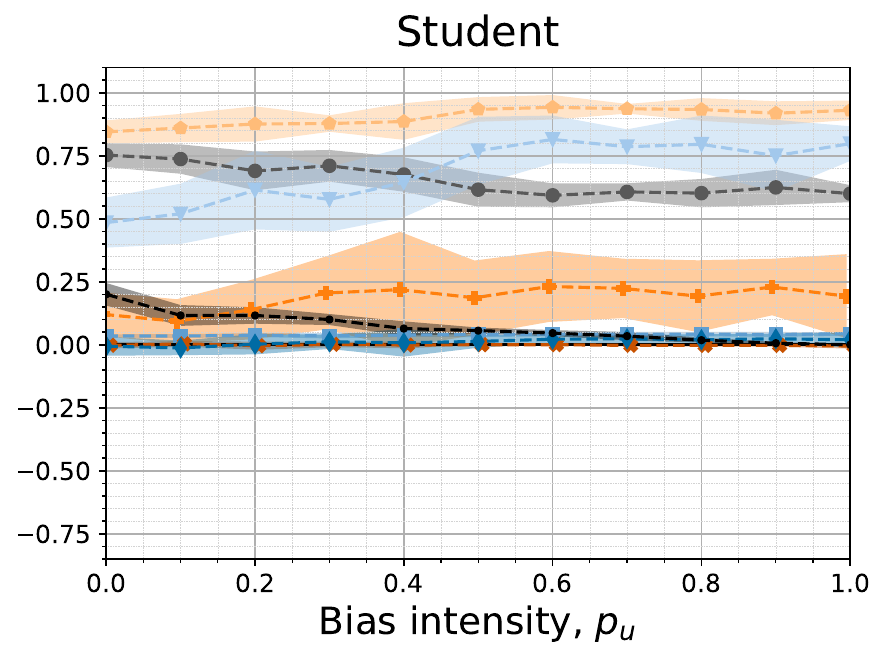}
                \end{minipage}
                \\
                \begin{minipage}[c]{.335\textwidth}
                    \centering
                    \includegraphics[width=\textwidth]{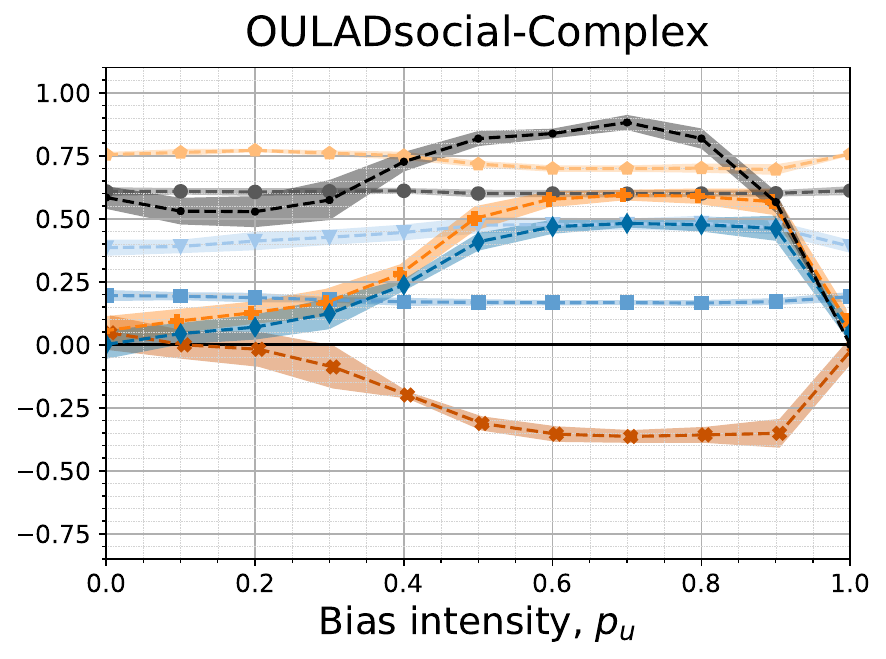}
                \end{minipage}
                \begin{minipage}[c]{.32\textwidth}
                    \centering
                    \includegraphics[width=\textwidth]{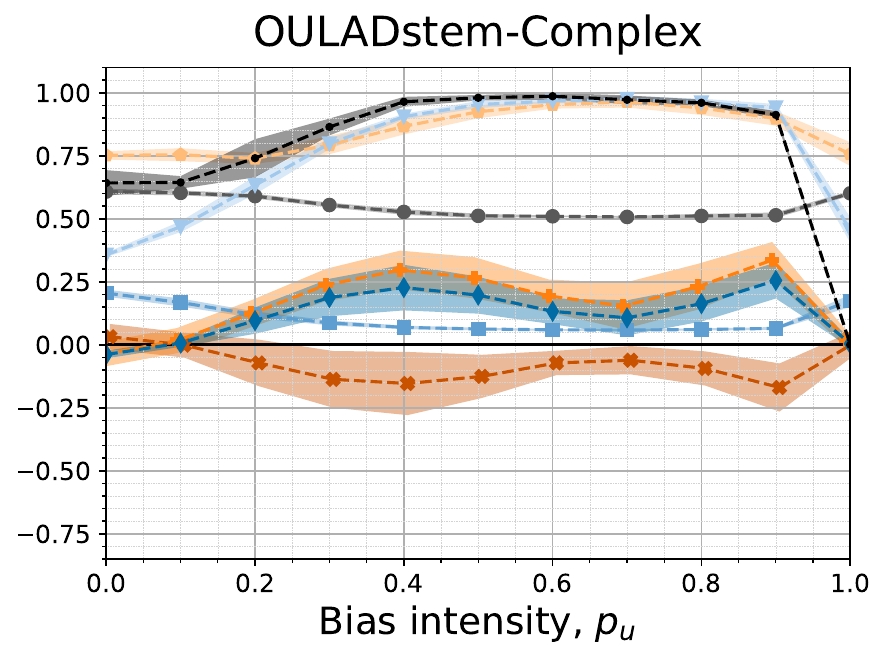}
                \end{minipage}
                \begin{minipage}[c]{.32\textwidth}
                    \centering
                    \includegraphics[width=\textwidth]{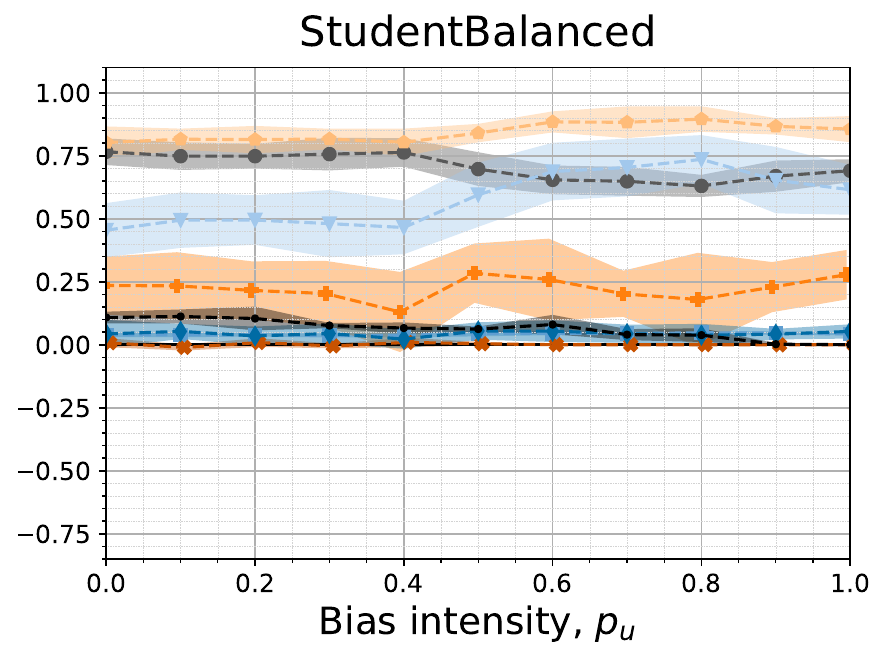}
                \end{minipage}
            \caption{Evolution of RF models performance when trained on datasets presenting increasing levels of \textbf{self-selection}, as evaluated on unbiased data. $p_u = 1$ corresponds to the complete removal of the unprivileged group from the training data.}
            \label{fig:impact_selfSelect}
            \end{figure}

            \paragraph{Self-selection}
            The results for self-selection can be seen in Figure \ref{fig:impact_selfSelect}. This type of selection bias has a more important impact than random selection on the sensitive attribute usage. This can be explained by the fact self-selection distorts the distribution of the unprivileged group.

            \stepcounter{observ}
            \underline{Observation \arabic{resultSubsec}.\arabic{observ}} : \textbf{The influence of self-selection on model performance is negligible for large datasets where the WAE assumption holds and with sufficiently predictive features}. It is however significant when the problem becomes too hard. The following results observable in Figure \ref{fig:impact_selfSelect} support this observation :
            \begin{itemize}
                \item For OULADsocial and OULADstem, the model can correctly learn the relevant patterns in training data despite the distribution distortion in the training set.
                \item For OULADsocial-Complex and OULADstem-Complex, where the non-sensitive features are significantly less predictive, the model reliance on the sensitive attribute does impact the fairness and accuracy performance of the model. In that case, there is a stronger increase in group unfairness when the unprivileged group is a minority (OULADsocial-Complex), while the impact on accuracy and individual fairness is less visible. The opposite trend is observed for OULADstem-Complex where the unprivileged group represents $82\%$ of the dataset.
                \item For both of the smaller datasets Student and StudentBalanced, self-selection leads to a reduction in fairness.
            \end{itemize}             

            \stepcounter{observ}
            \underline{Observation \arabic{resultSubsec}.\arabic{observ}} : \textbf{When self-selection does affect model performance, the complete removal of the unprivileged group can lead to a fairer and more accurate model if the WAE assumption holds and the number of remaining instances is sufficient.} We expect this result to extend to all subtypes of selection bias that only affect unprivileged individuals. Supporting this observation, Figure \ref{fig:impact_selfSelect} shows that models trained on OULADsocial-Complex or OULADstem-Complex are overall at least as good for $p_u=1$ than for the fair baseline ($p_u=0$). It is however not the case for Student and StudentBalanced.

            \begin{figure}[h]
            \centering
                \begin{minipage}[c]{.335\textwidth}
                    \centering
                    \includegraphics[width=\textwidth]{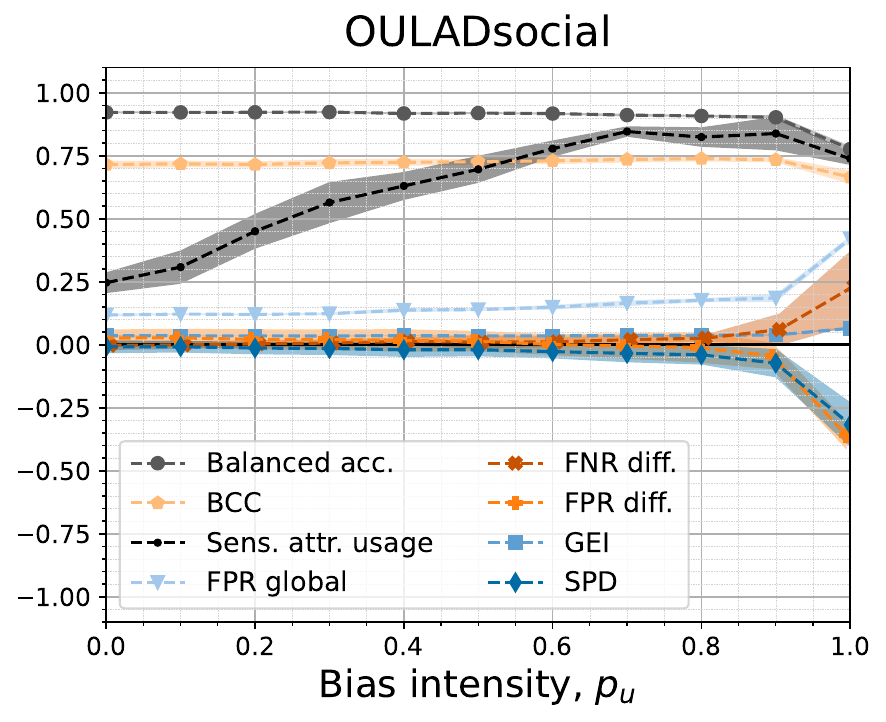}
                \end{minipage}
                \begin{minipage}[c]{.32\textwidth}
                    \centering
                    \includegraphics[width=\textwidth]{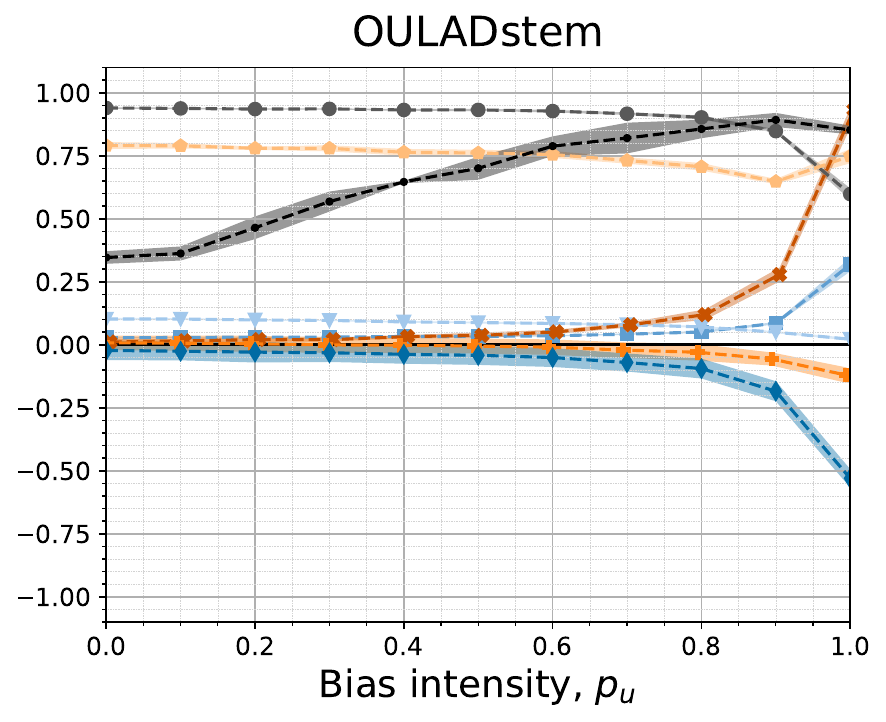}
                \end{minipage}
                \begin{minipage}[c]{.32\textwidth}
                    \centering
                    \includegraphics[width=\textwidth]{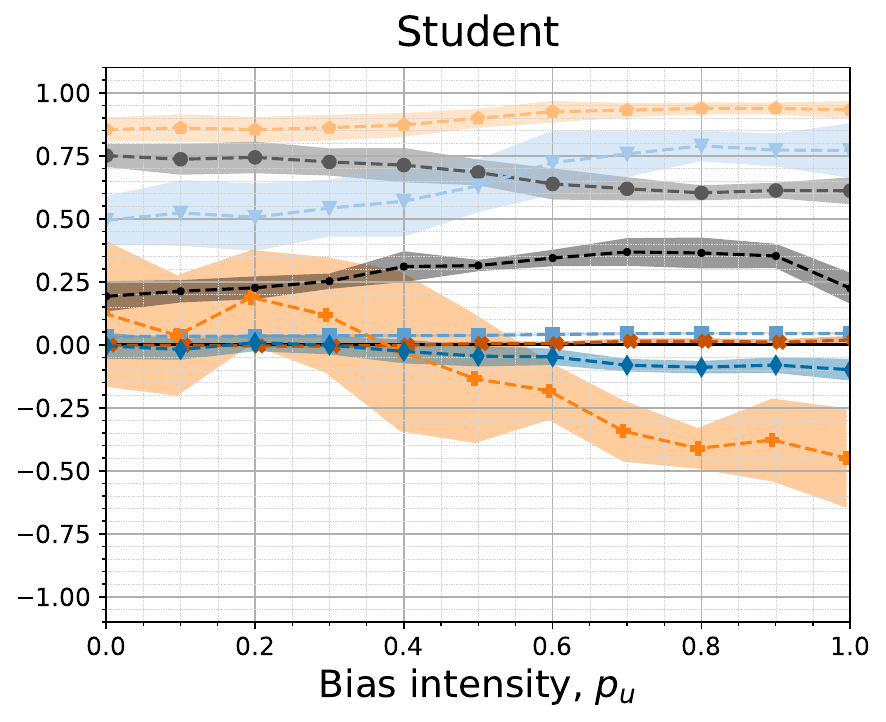}
                \end{minipage}
                \\
                \begin{minipage}[c]{.335\textwidth}
                    \centering
                    \includegraphics[width=\textwidth]{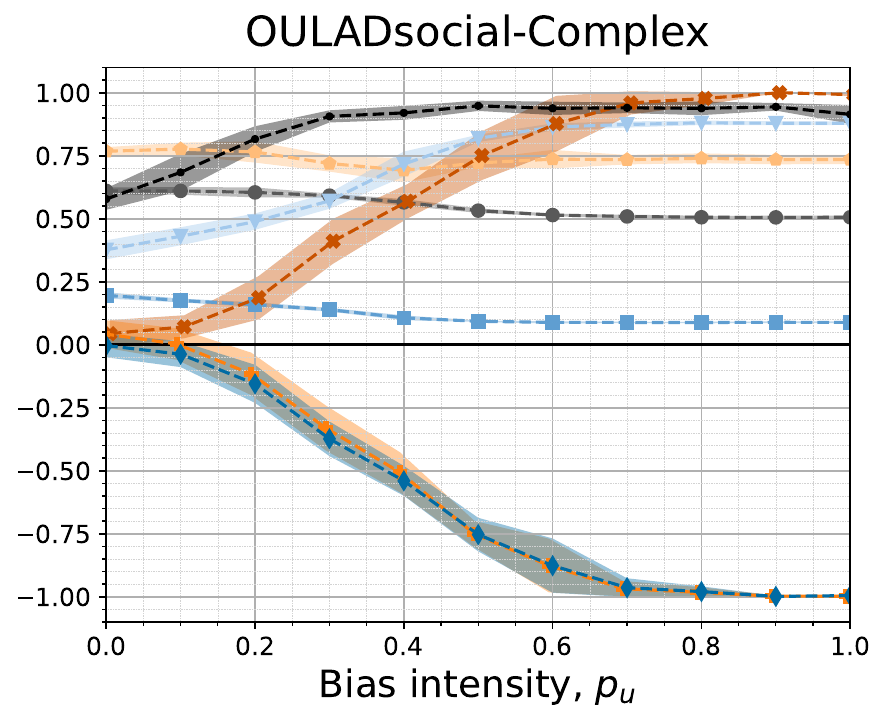}
                \end{minipage}
                \begin{minipage}[c]{.32\textwidth}
                    \centering
                    \includegraphics[width=\textwidth]{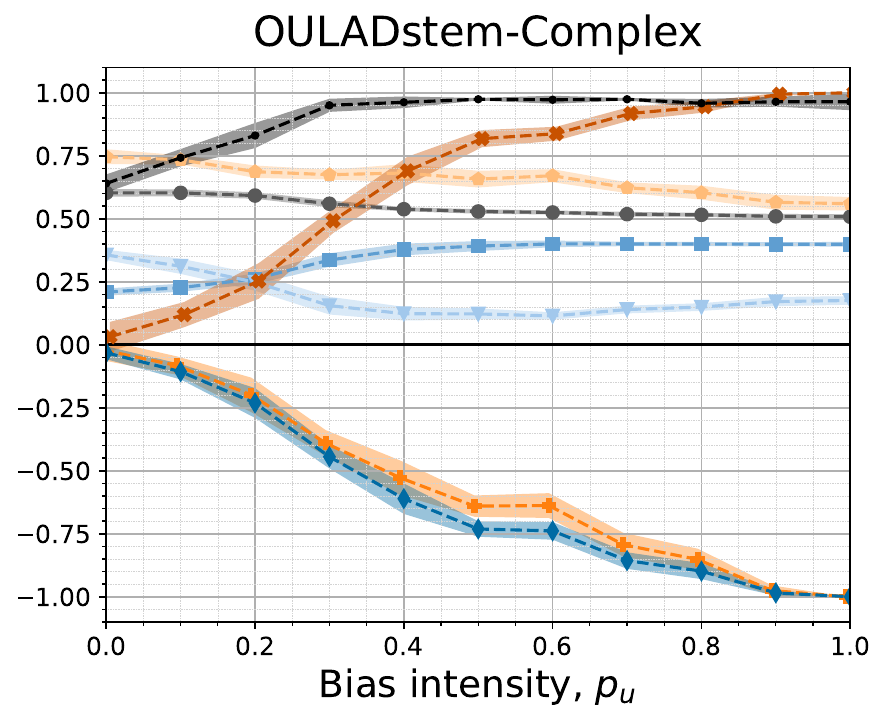}
                \end{minipage}
                \begin{minipage}[c]{.32\textwidth}
                    \centering
                    \includegraphics[width=\textwidth]{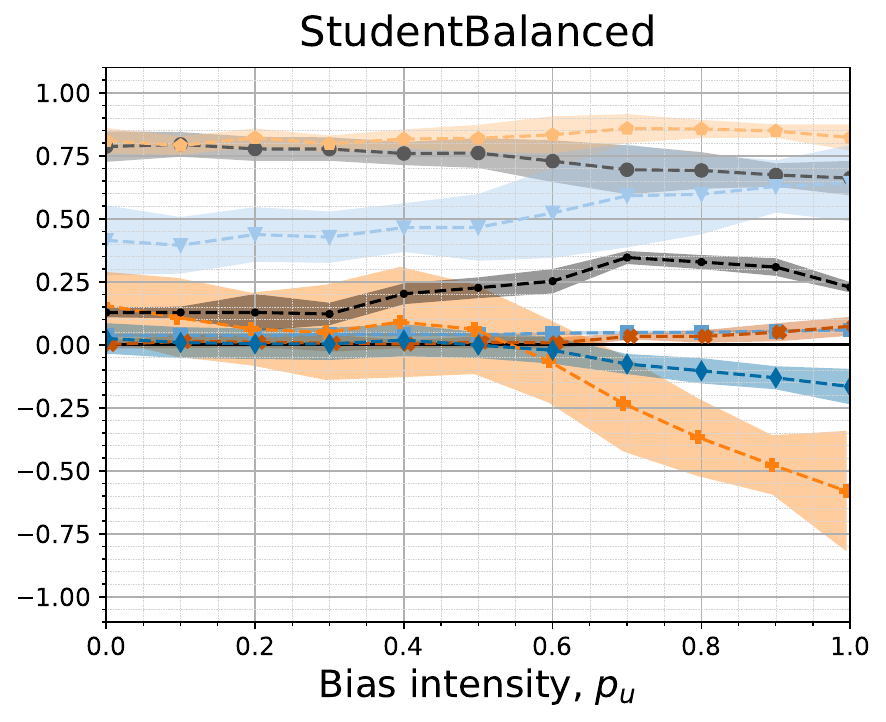}
                \end{minipage}
            \caption{Evolution of RF models performance when trained on datasets presenting increasing levels of \textbf{malicious selection}, as evaluated on unbiased data. When $p_u=1$, the members of the privileged (resp. unprivileged) group all receive positive (resp. negative) outcomes.}
            \label{fig:impact_maliSelect}
            \end{figure}

\begin{figure}[h]
            \centering
                \begin{minipage}[c]{.335\textwidth}
                    \centering
                    \includegraphics[width=\textwidth]{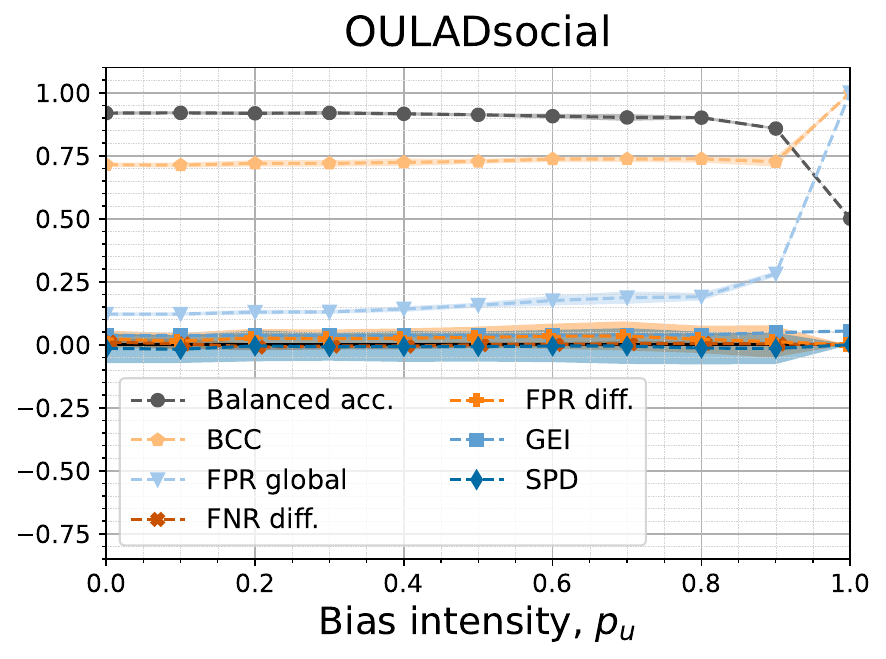}
                \end{minipage}
                \begin{minipage}[c]{.32\textwidth}
                    \centering
                    \includegraphics[width=\textwidth]{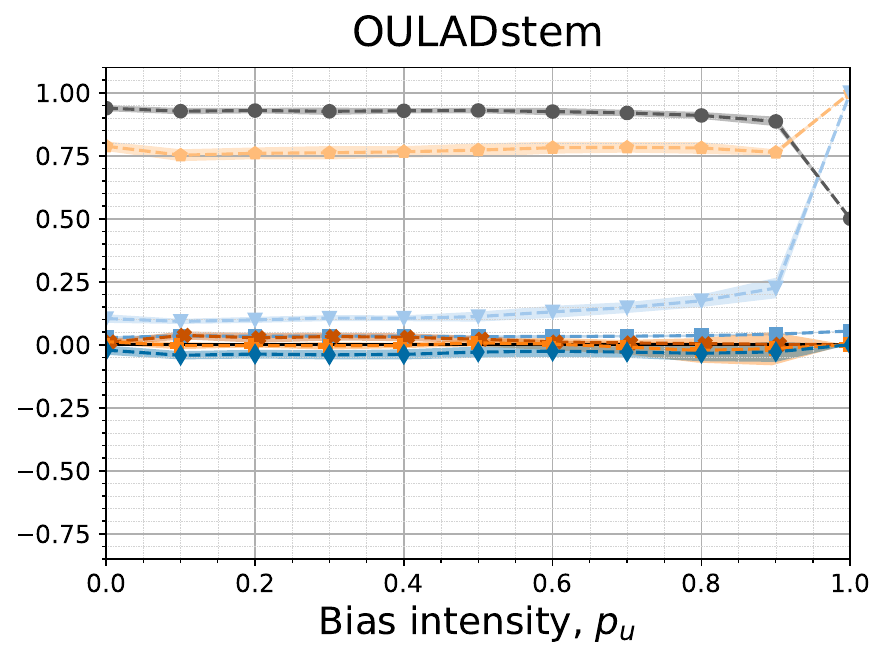}
                \end{minipage}
                \begin{minipage}[c]{.32\textwidth}
                    \centering
                    \includegraphics[width=\textwidth]{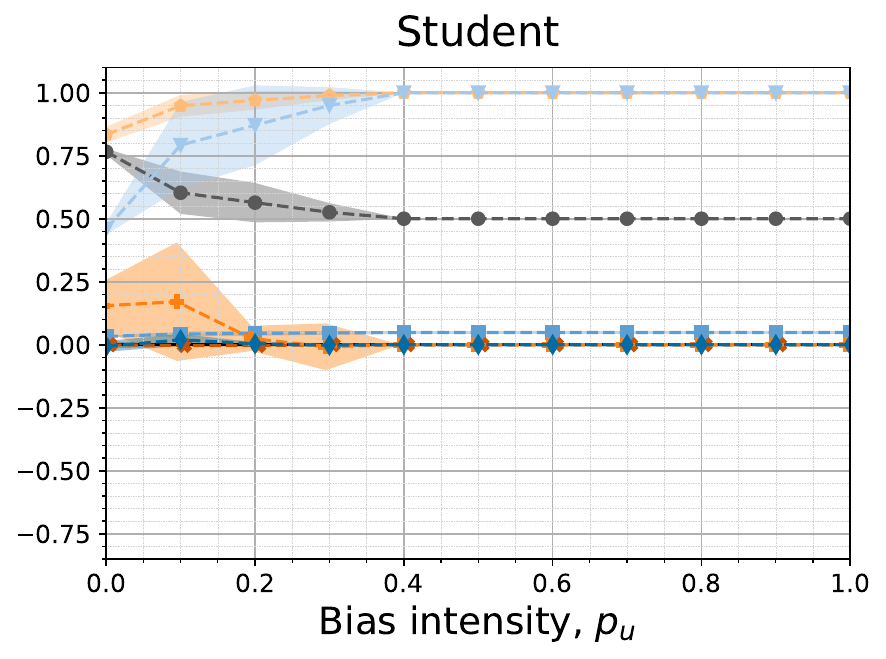}
                \end{minipage}
                \\
                \begin{minipage}[c]{.335\textwidth}
                    \centering
                    \includegraphics[width=\textwidth]{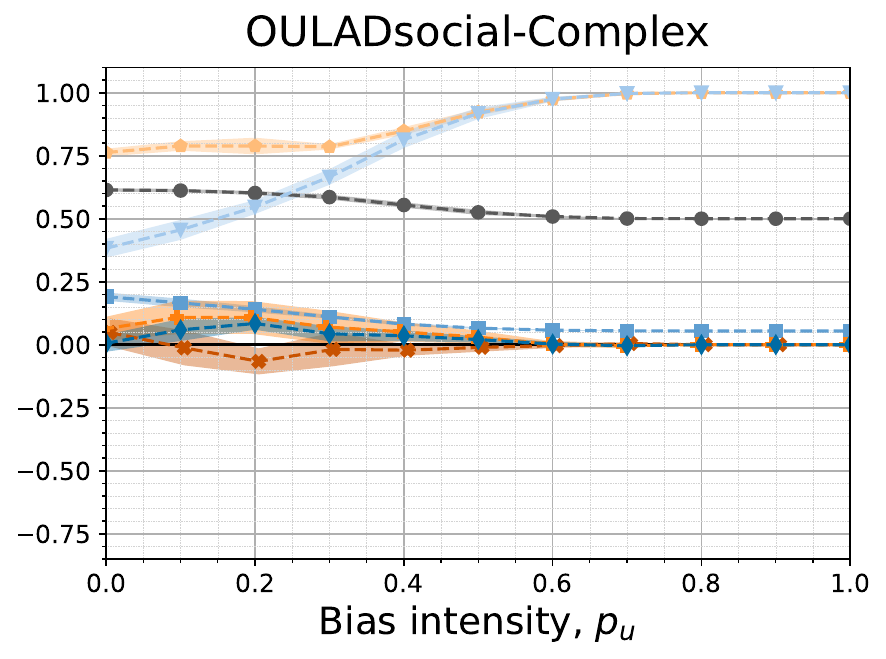}
                \end{minipage}
                \begin{minipage}[c]{.32\textwidth}
                    \centering
                    \includegraphics[width=\textwidth]{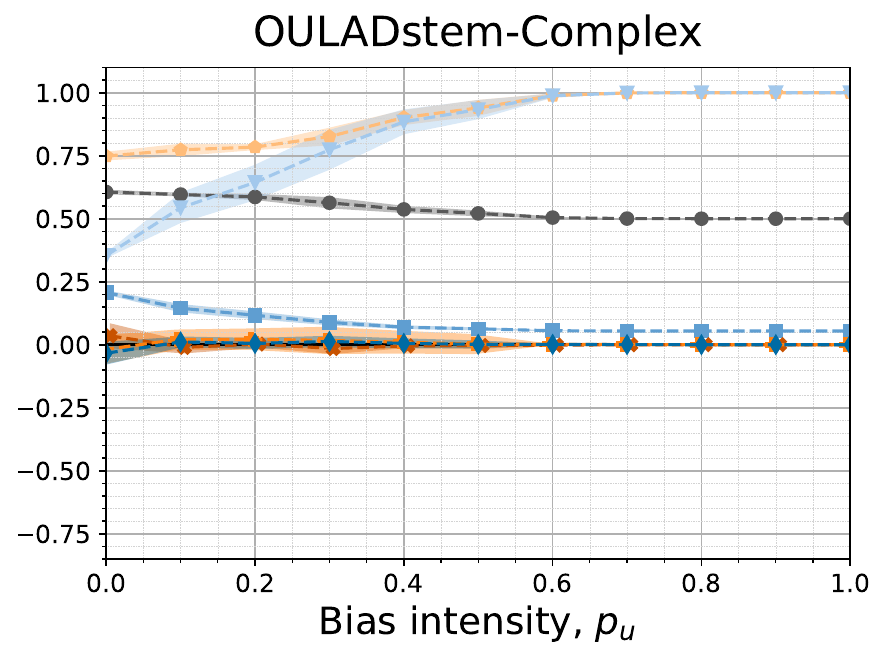}
                \end{minipage}
                \begin{minipage}[c]{.32\textwidth}
                    \centering
                    \includegraphics[width=\textwidth]{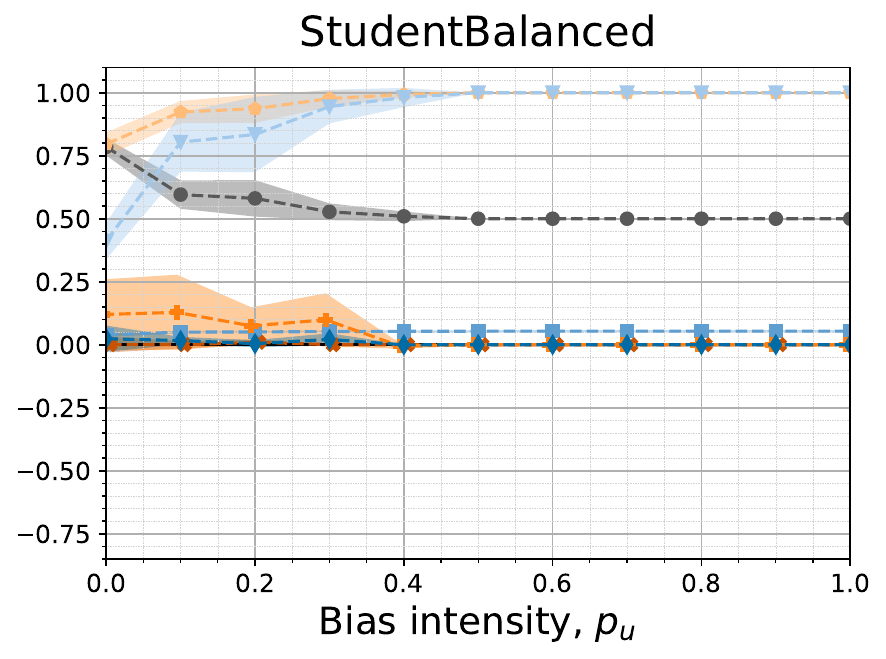}
                \end{minipage}
            \caption{Evolution of RF models performance trained on datasets from which the \textbf{unprivileged group is excluded} and presenting increasing levels of \textbf{malicious selection}, as evaluated on unbiased data. Models at $p_u = 0$ have been trained with the original dataset containing the two groups.}
            \label{fig:impact_maliciousNoUnpriv}
            \end{figure}

            \paragraph{Malicious selection} The results regarding malicious selection are shown in Figure \ref{fig:impact_maliSelect}. Since it affects both the privileged and unprivileged groups, this bias type has a bigger influence on the sensitive attribute usage than self-selection. Additionally, Figure \ref{fig:impact_maliciousNoUnpriv} present the results for models trained on data affected by malicious selection but from which the unprivileged group has been completely excluded.
            
            \stepcounter{observ}
            \underline{Observation \arabic{resultSubsec}.\arabic{observ}} : \textbf{When the features are sufficiently predictive, the impact of malicious selection on models is very limited, except for very high bias intensity. Larger datasets and lower SPD values improve the robustness of model, increasing the threshold of what should be considered "very high" intensity.} On the other hand, if the features aren't predictive enough to learn the relevant relationships, the reliance on the sensitive attribute, which becomes even higher, leads to significant increase in unfairness. This observation stems from the following results, visible in Figure \ref{fig:impact_maliSelect} :
            \begin{itemize}
                \item The impact of malicious selection is very limited for OULADstem and OULADsocial, except for extreme bias intensity.
                \item There is a direct negative impact on models trained with OULADsocial-Complex and OULADstem-Complex.
                \item For the smaller Student ($SPD=-0.057$) and StudentBalanced, the influence of malicious bias is also very limited up to a certain threshold, which is higher for StudentBalanced ($SPD=0$).
            \end{itemize}

            \stepcounter{observ}
            \underline{Observation \arabic{resultSubsec}.\arabic{observ}} :
             \textbf{In case of malicious selection, the absence of the unprivileged group in training prevents group discrimination. If the data quality and quantity is sufficient, removing this group has little effect on accuracy and individual fairness except for extreme bias intensity.} In case of a smaller dataset or insufficient data quality, the accuracy can be significantly reduced. The positive effect for group fairness is expected since the models must generalize information from the privileged group to the whole dataset. The following results observable in Figure \ref{fig:impact_maliciousNoUnpriv} allow to make these observations :
             \begin{itemize}
                 \item The values of group fairness metrics are close to zero for all datasets.
                 \item All metrics give values close to the fair baseline when models are trained on OULADsocial or OULADstem (except for the highest bias levels), but it is not the case for their complex versions.
                 \item Models trained on the smaller datasets Student or StudenBalanced are both directly highly impacted by malicious selection.
             \end{itemize}
  
        \paragraph{Discussion}
            Distinct manifestations of selection bias lead to different impact on model performance and their effect may be negligible under different conditions. Assumption on the kind of selection bias considered should thus be made explicit and results obtained for one subtype should be generalized with caution.
                
            When features are sufficiently predictive and there is enough data, fairness-agnostic models are able to learn the relevant prediction patterns despite distortion in the distribution of the training data, with the exception of extreme distortions which represent less realistic scenarios. Under these conditions, the reliance of models on the sensitive attribute can be influenced by the selection bias, but without impacting models accuracy and fairness performance. Data quality is thus an important factor to limit the effect of selection bias involving distribution distortions.

            Our results indicate that the approach suggested in \cite{ceccon_dataBiasProfile_2025} to reserve high-quality data about the unprivileged group for evaluation rather than training can be beneficial even in case of selection bias affecting both groups, at the condition that the remaining data about the privileged group is of sufficient quality. This strategy can also be beneficial when self-selection bias affects the model behavior.

    \subsection{Performance of bias mitigation methods}\label{sec:res_mitig}
                    
        We compare in this section the performance of the different bias mitigation methods evaluated on the unbiased test set representing the fair world, in accordance with our framework (see Section \ref{sec:framework_sec}). 
        We first discuss the efficiency of each method in relationship with the different bias types and present a summary of their performance in Table \ref{tab:method_performance}. We then highlight the main conclusions that can be drawn from these experiments. Following the results of Section \ref{sec:res_model}, we exclude random selection from this analysis since its impact is negligible and do not use Student since the WAE assumption does not hold for that dataset.
        
        We display the results in graphs that each present the evolution of a specific metric for RF models trained with an increasingly biased dataset. On each plot, the results for the unmitigated models (in black) are displayed for each training set bias level, ranging from 0 to 0.9, and are followed by their mitigated counterparts. 
        A bias level of 0 indicates that the model was trained with the original unbiased test set. The results for such models correspond to the unbiased baseline and are also indicated with a dotted horizontal line. 
        An ideal bias mitigation methods is thus one that can restore (or improve) that baseline value for all metrics. 
        
\begin{figure}[h]
    \centering
        \begin{minipage}[c]{.49\textwidth}
            \begin{flushright}
            \includegraphics[height=4.025cm]{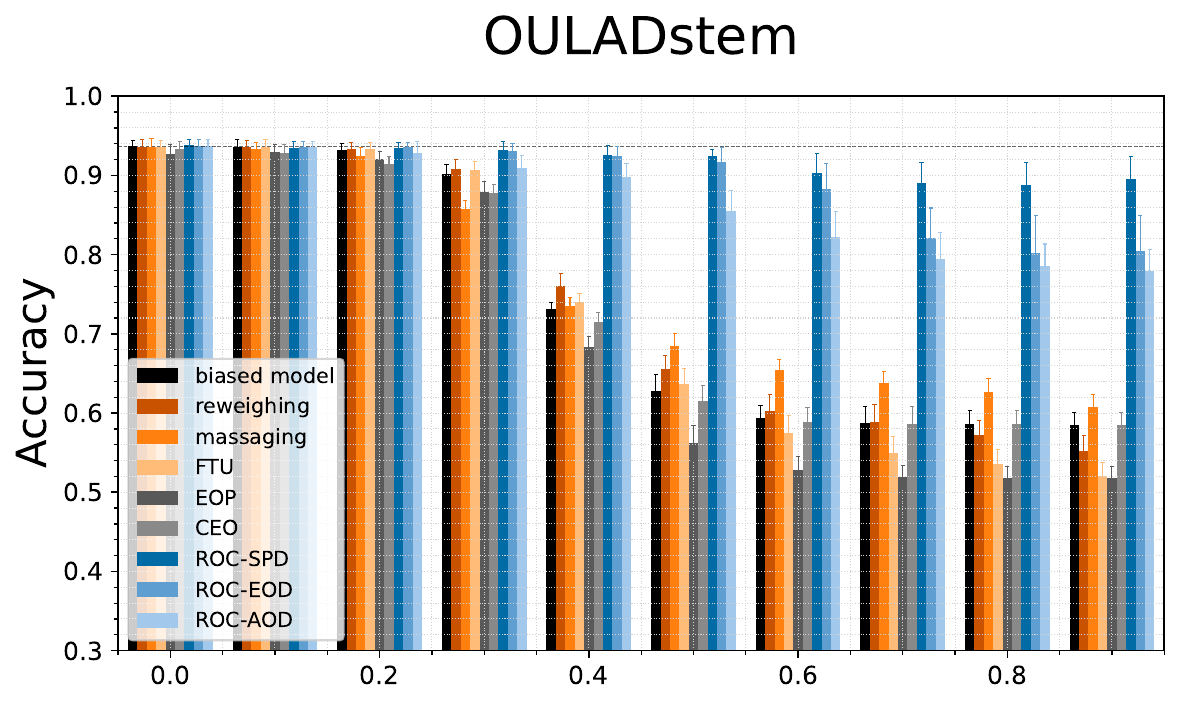}
            \end{flushright}
        \end{minipage}
        \begin{minipage}[c]{.49\textwidth}
            \begin{flushright}
            \includegraphics[height=4.025cm]{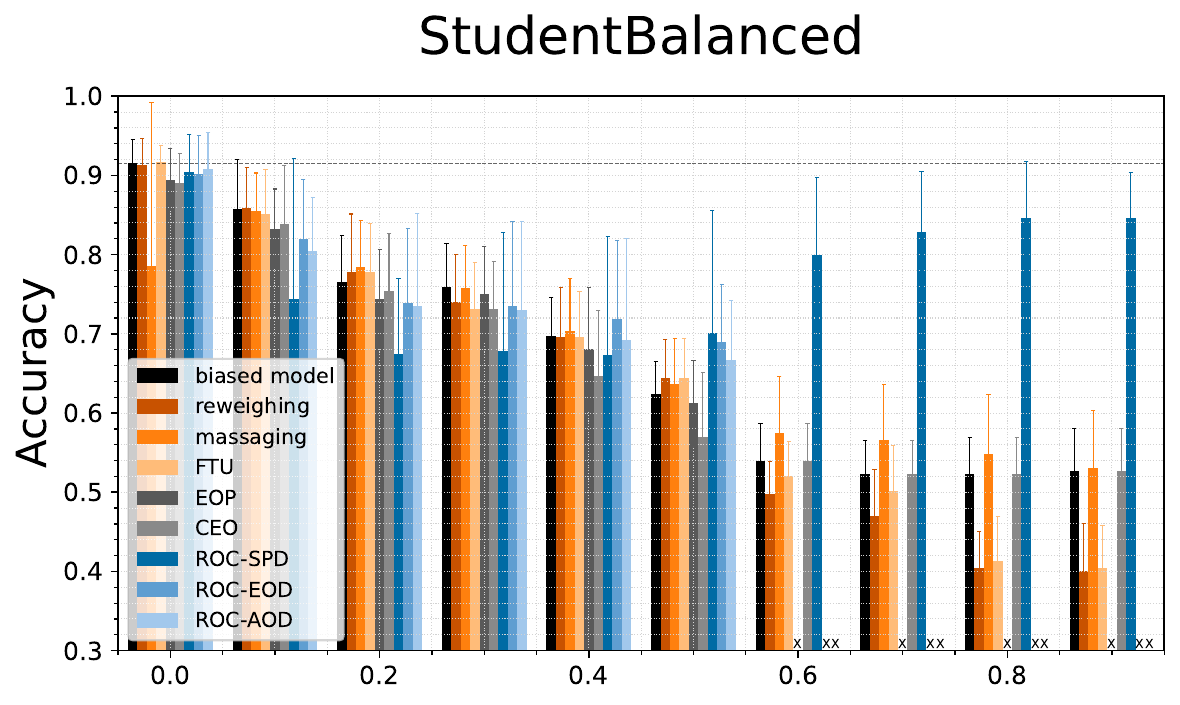}
            \end{flushright}
        \end{minipage}
        \\
         \begin{minipage}[c]{.49\textwidth}
            \begin{flushright}
            \includegraphics[height=3.6cm]{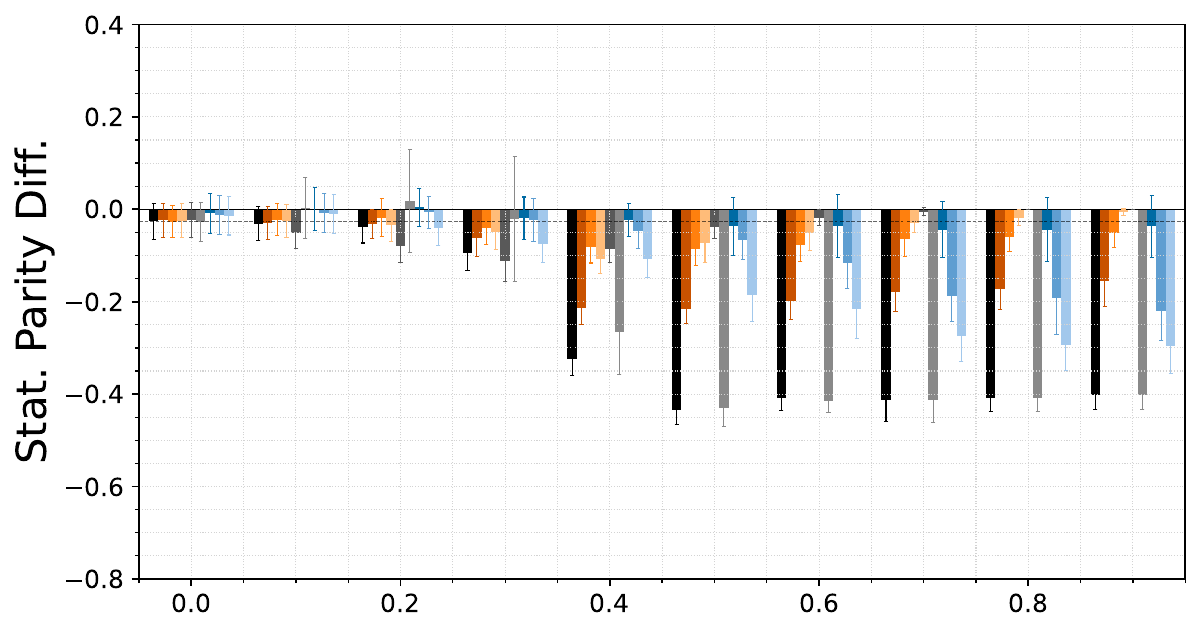}
            \end{flushright}
        \end{minipage}
        \begin{minipage}[c]{.49\textwidth}
            \begin{flushright}
            \includegraphics[height=3.6cm]{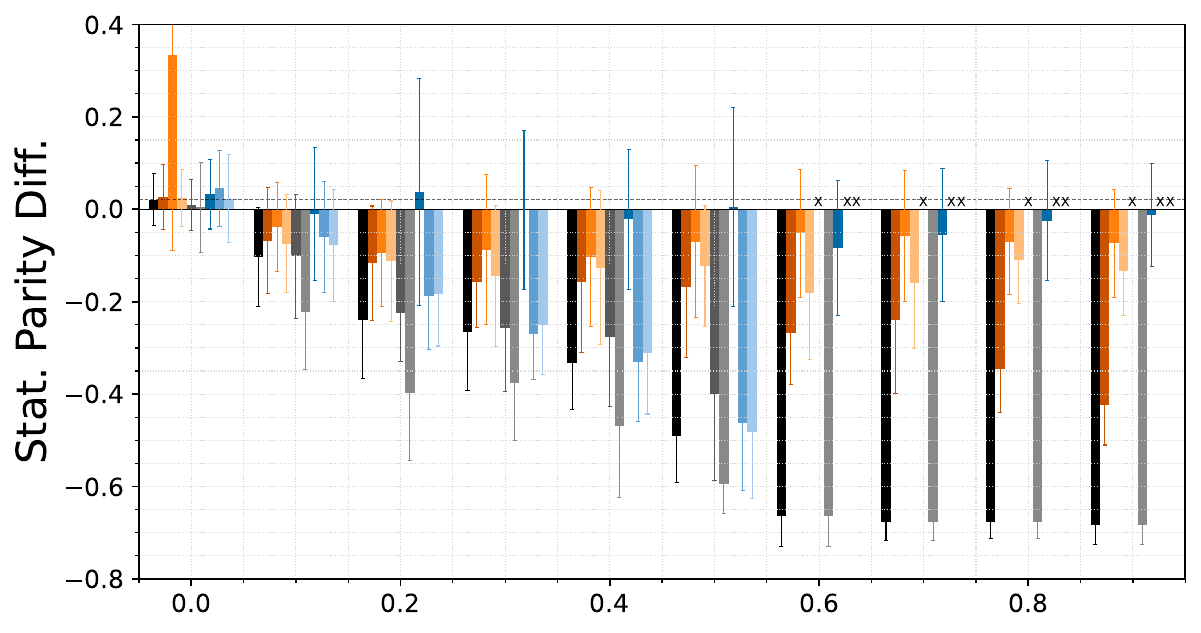}
            \end{flushright}
        \end{minipage}
         \\
        \begin{minipage}[c]{.49\textwidth}
            \begin{flushright}
            \includegraphics[height=3.55cm]{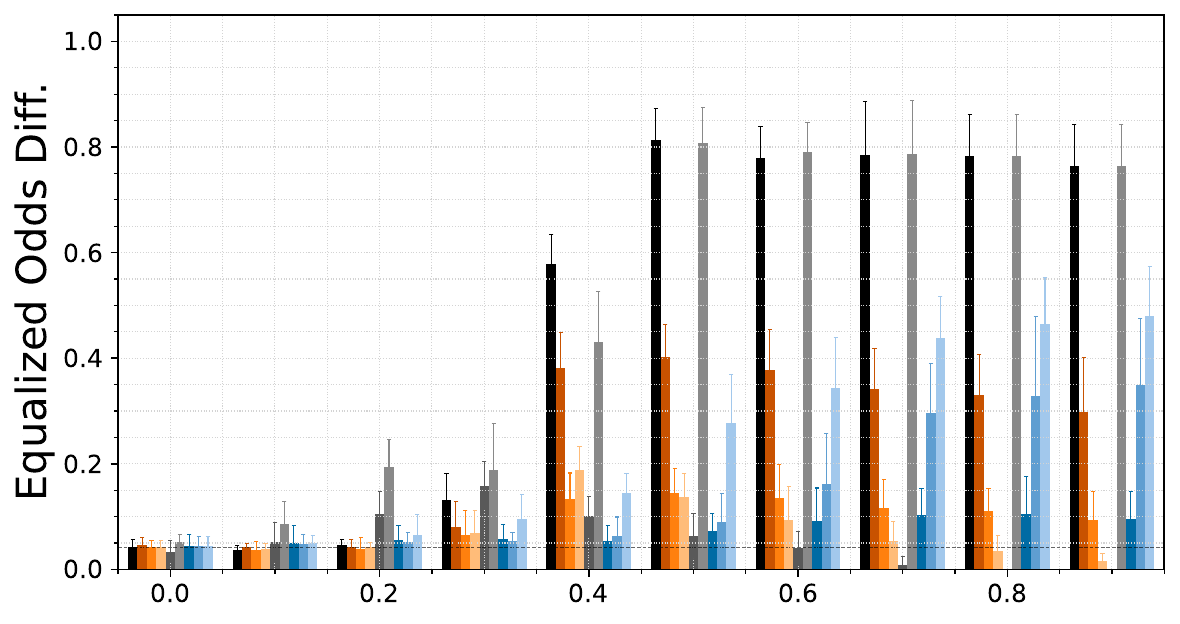}
            \end{flushright}
        \end{minipage}
        \begin{minipage}[c]{.49\textwidth}
            \begin{flushright}
            \includegraphics[height=3.55cm]{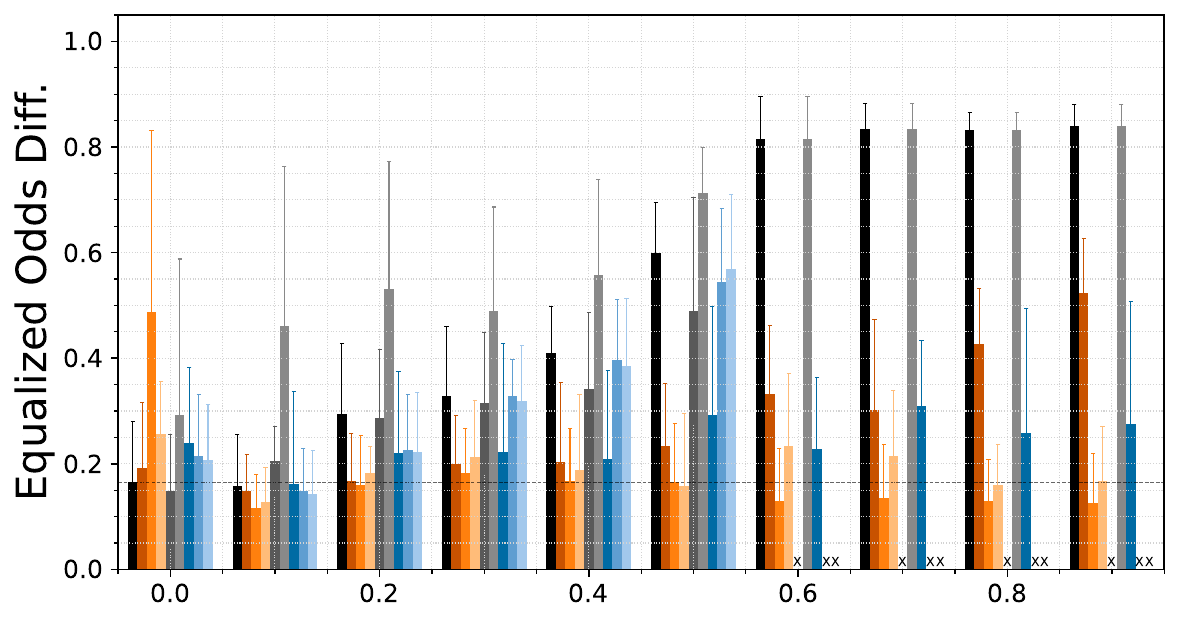}
            \end{flushright}
        \end{minipage}
        \\
        \begin{minipage}[c]{.49\textwidth}
            \begin{flushright}
            \includegraphics[height=3.85cm]{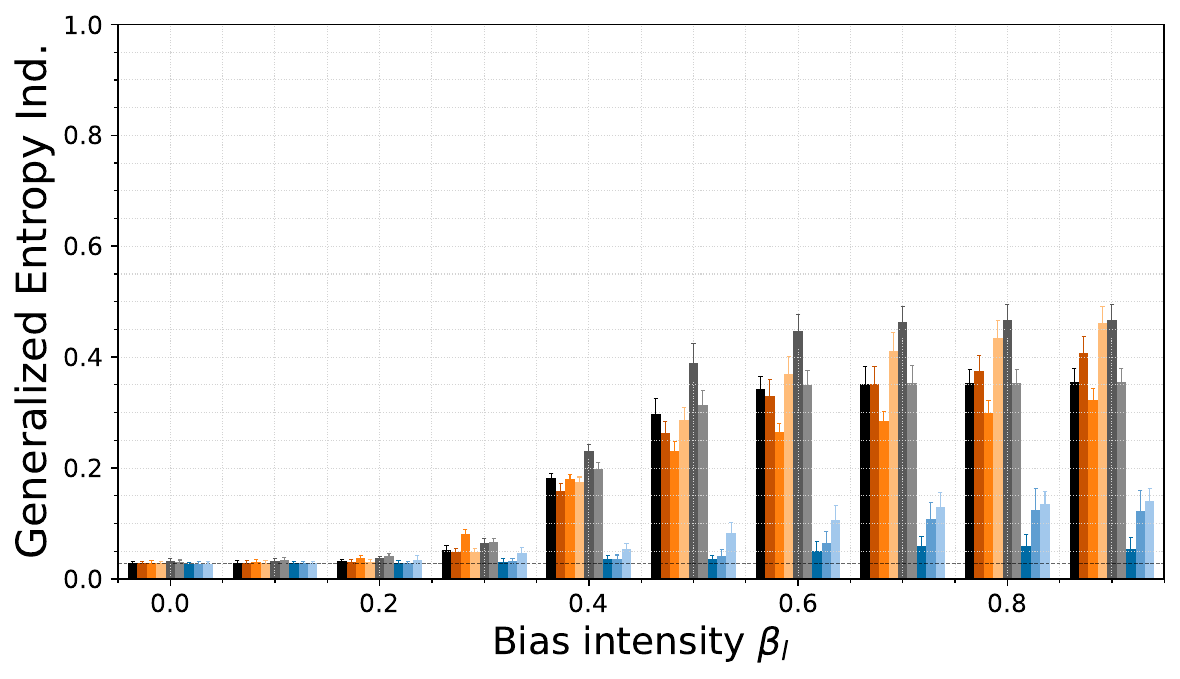}
            \end{flushright}
        \end{minipage}
        \begin{minipage}[c]{.49\textwidth}
            \begin{flushright}
            \includegraphics[height=3.85cm]{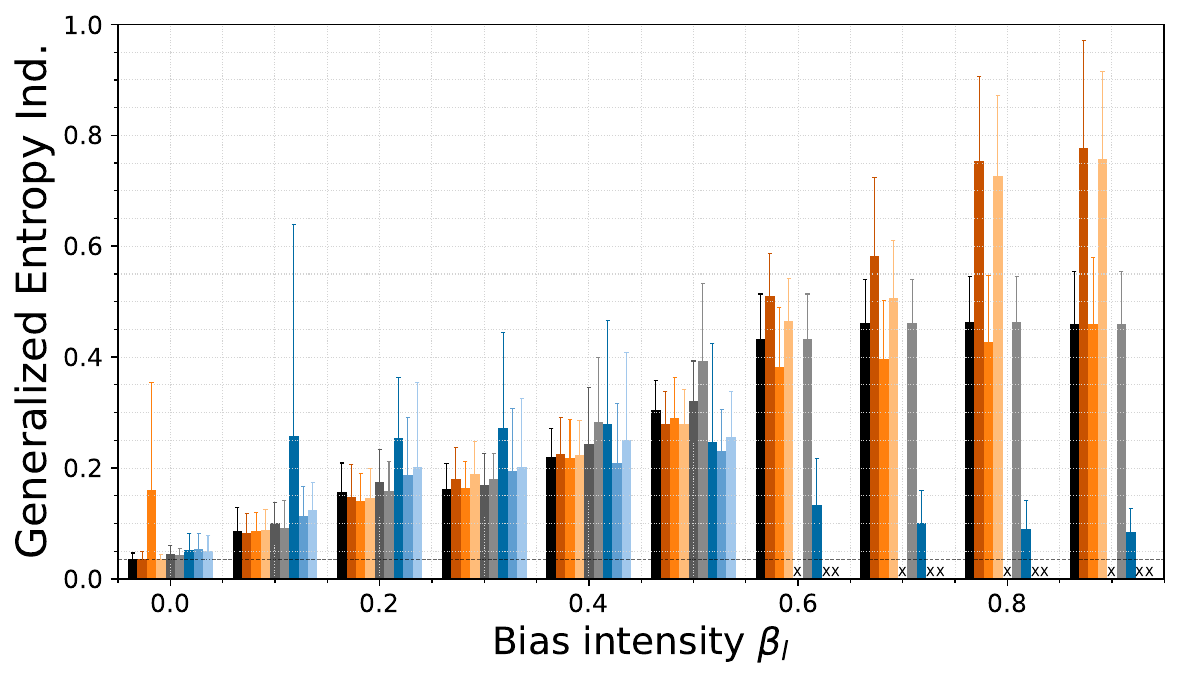}
            \end{flushright}
        \end{minipage}
    \caption{Evolution of accuracy and fairness metrics for RF models trained on data with increasing levels of \textbf{label bias} and evaluated on unbiased data. The fair baseline is indicated by a dashed horizontal black line.}\label{fig:comp_label}
    \end{figure}

\begin{figure}[h]
    \centering
        \begin{minipage}[c]{.49\textwidth}
            \begin{flushright}
            \includegraphics[height=4.025cm]{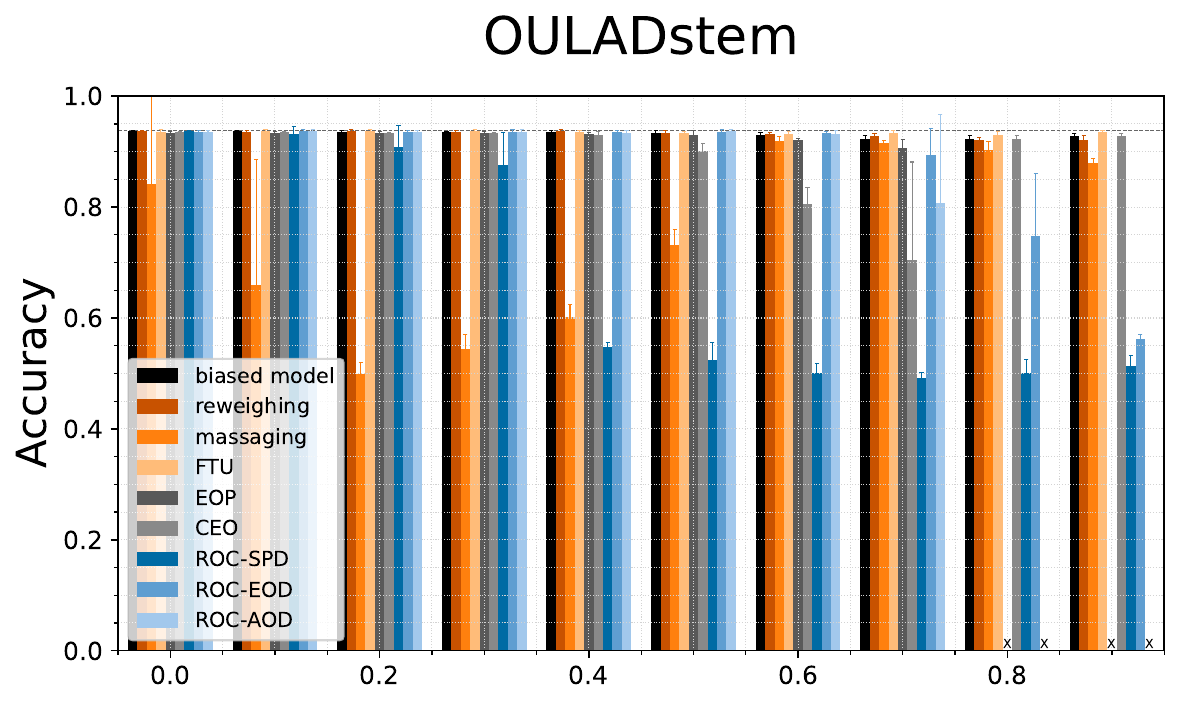}
            \end{flushright}
        \end{minipage}
        \begin{minipage}[c]{.49\textwidth}
            \begin{flushright}
            \includegraphics[height=4.025cm]{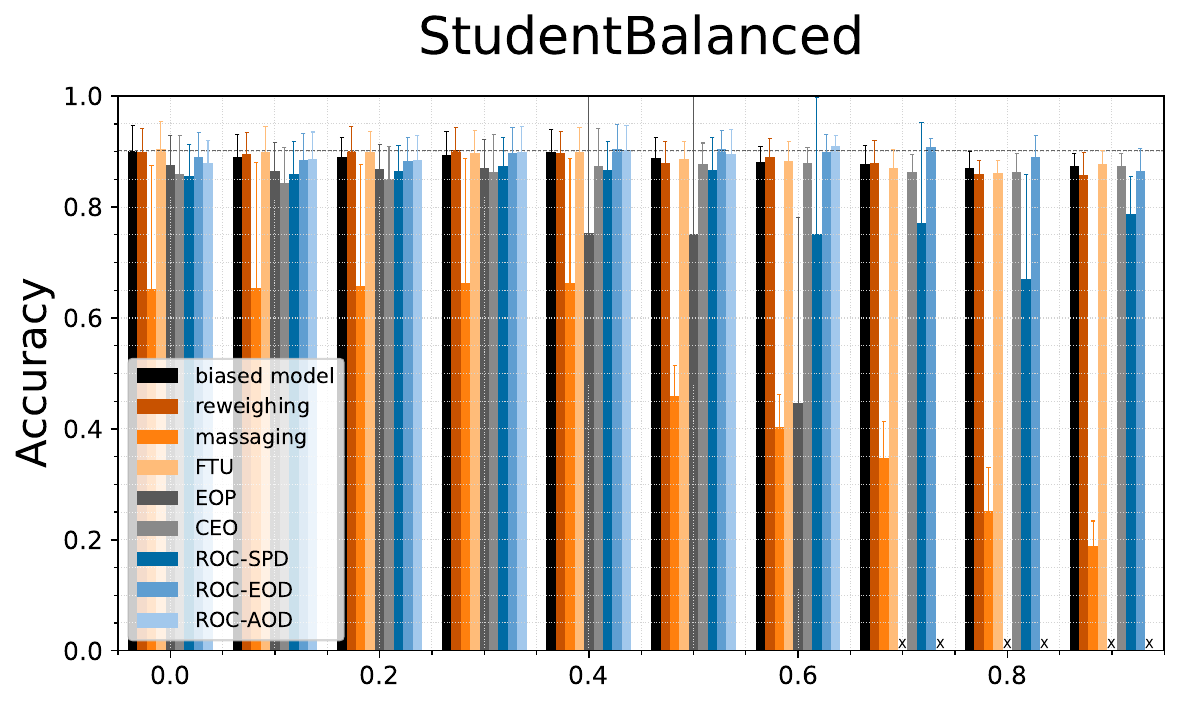}
            \end{flushright}
        \end{minipage}
        \\
         \begin{minipage}[c]{.49\textwidth}
            \begin{flushright}
            \includegraphics[height=3.6cm]{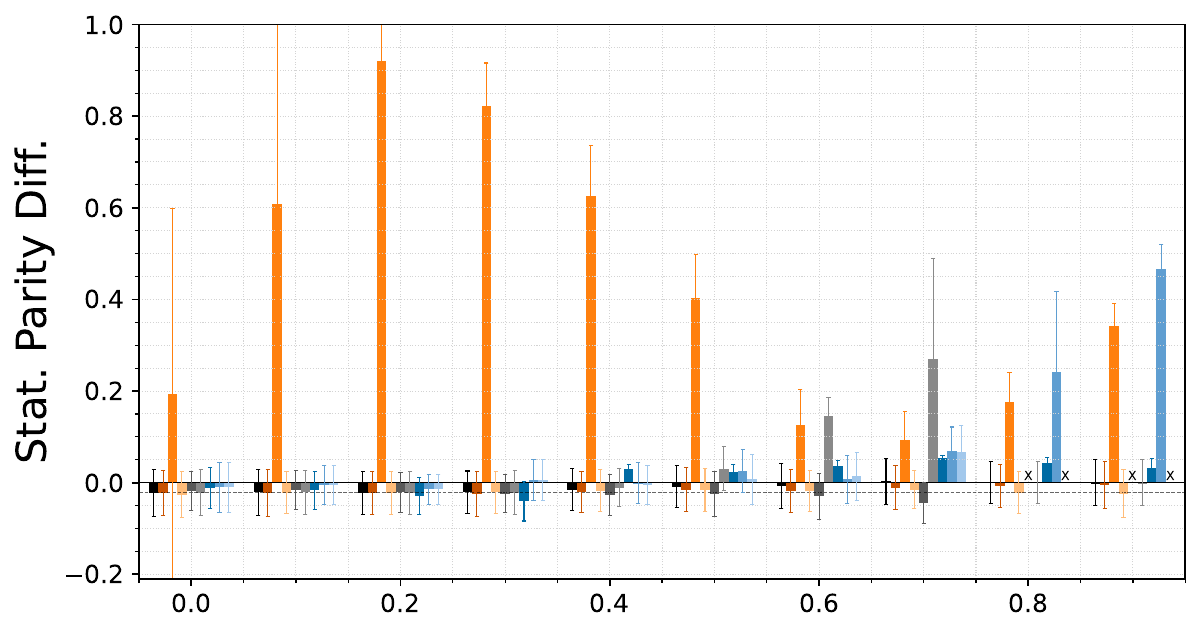}
            \end{flushright}
        \end{minipage}
        \begin{minipage}[c]{.49\textwidth}
            \begin{flushright}
            \includegraphics[height=3.6cm]{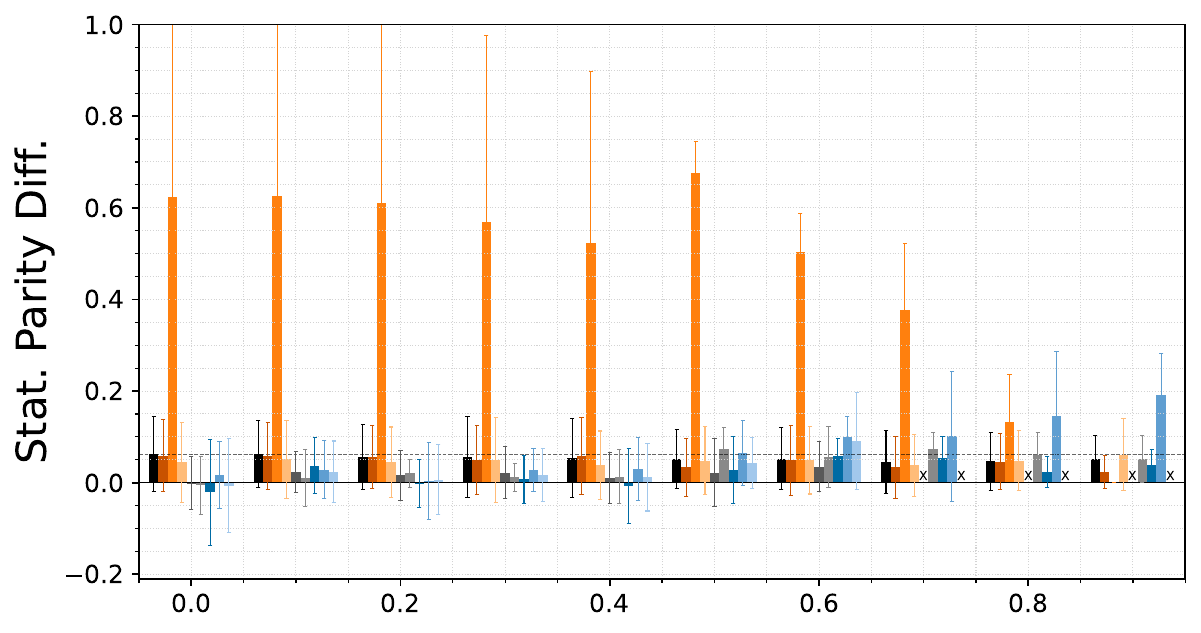}
            \end{flushright}
        \end{minipage}
        \\
        \begin{minipage}[c]{.49\textwidth}
            \begin{flushright}
            \includegraphics[height=3.55cm]{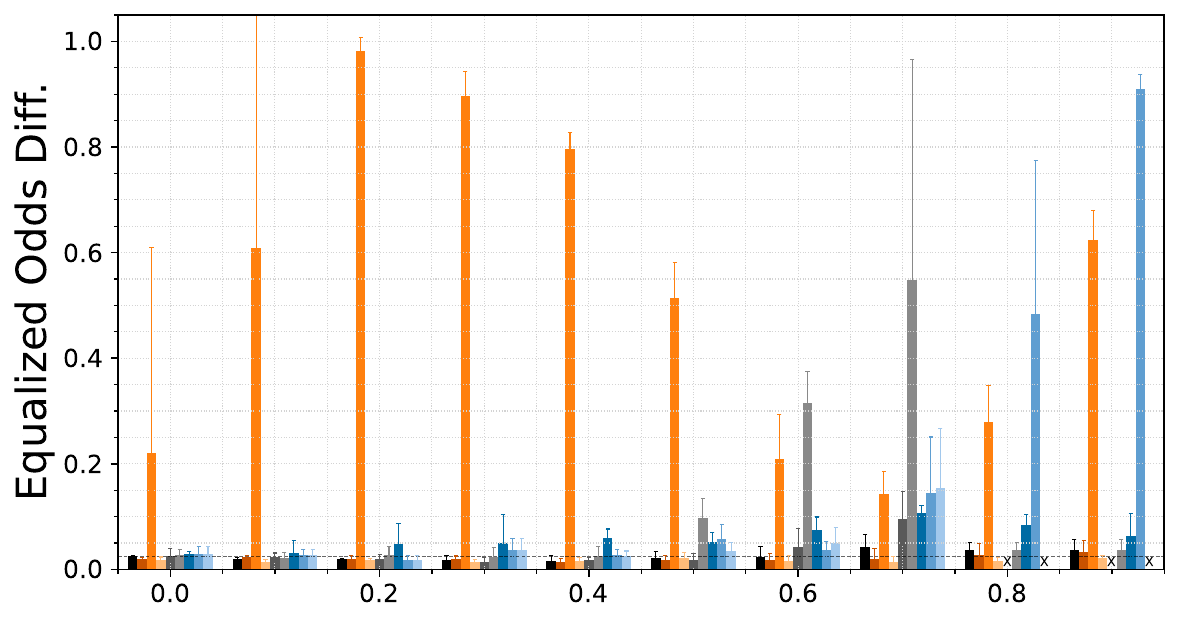}
            \end{flushright}
        \end{minipage}
        \begin{minipage}[c]{.49\textwidth}
            \begin{flushright}
            \includegraphics[height=3.55cm]{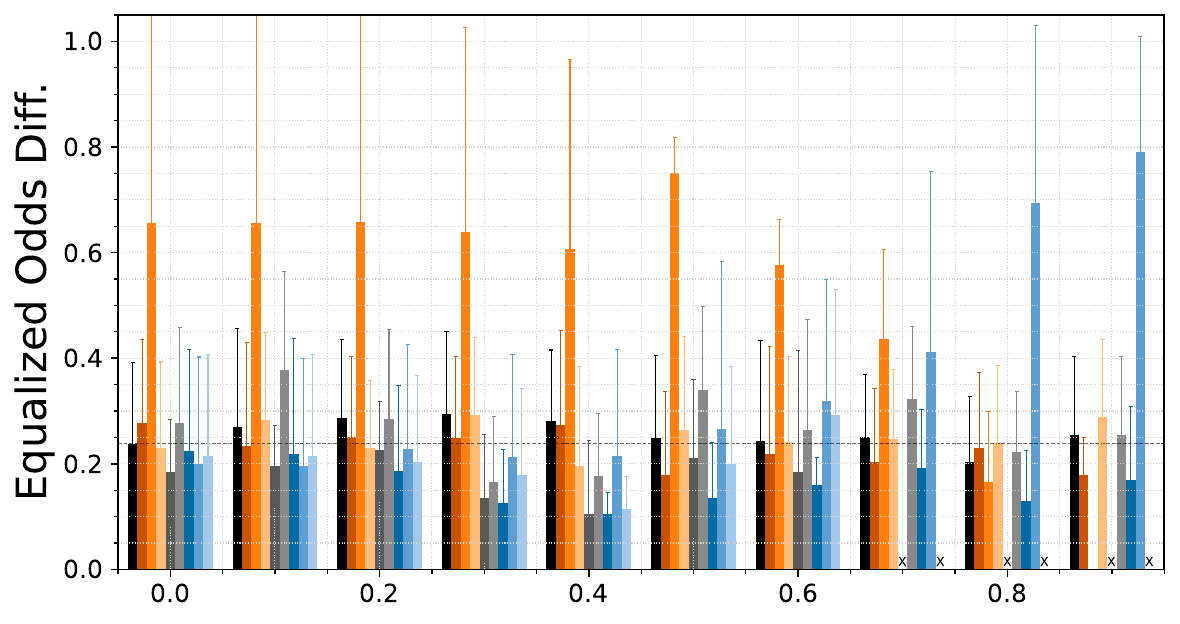}
            \end{flushright}
        \end{minipage}
        \\
        \begin{minipage}[c]{.49\textwidth}
            \begin{flushright}
            \includegraphics[height=3.85cm]{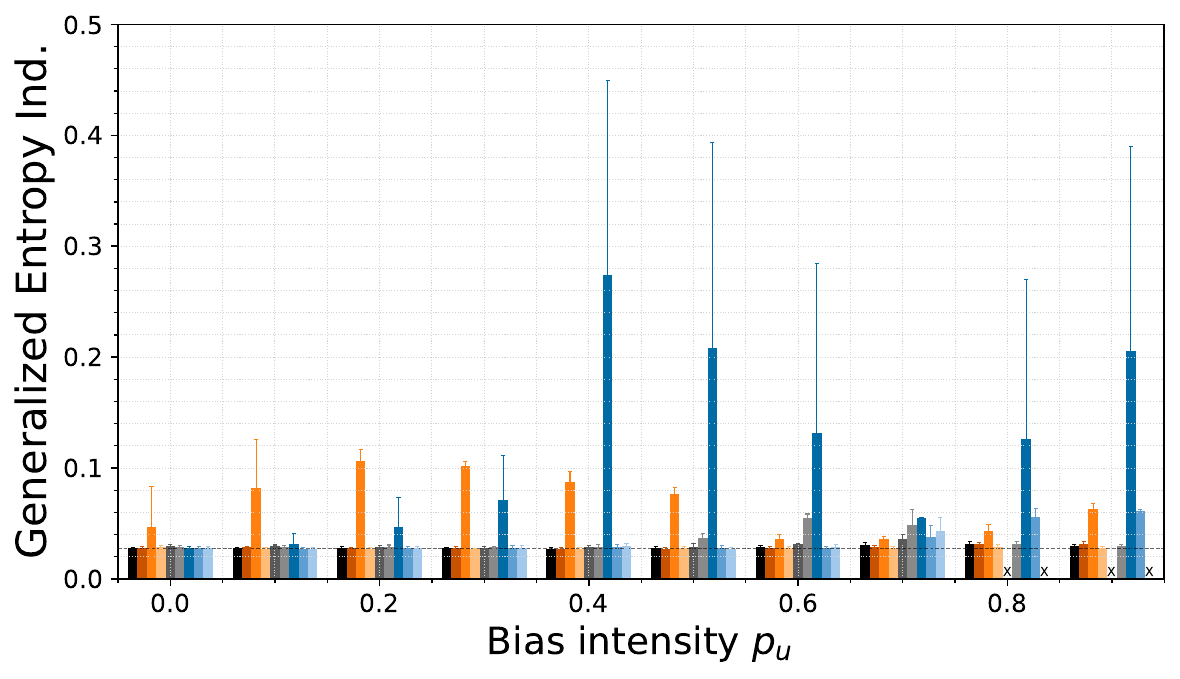}
            \end{flushright}
        \end{minipage}
        \begin{minipage}[c]{.49\textwidth}
            \begin{flushright}
            \includegraphics[height=3.85cm]{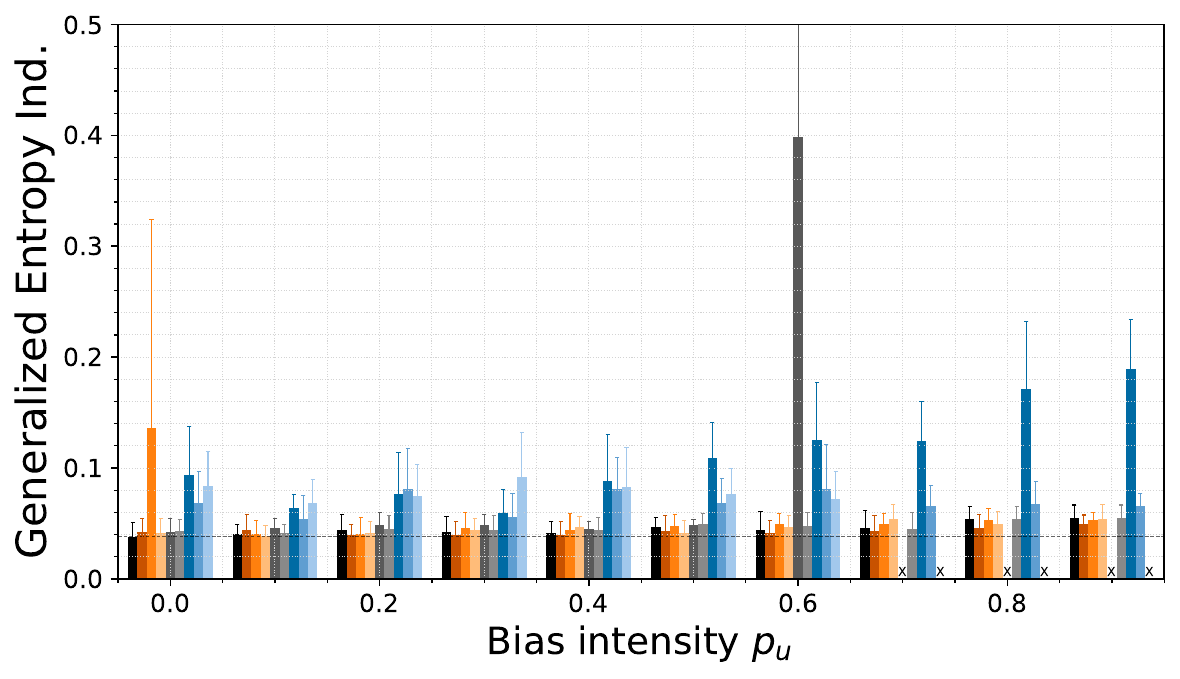}
            \end{flushright}
        \end{minipage}
    \caption{Evolution of accuracy and fairness metrics for RF models trained on data with increasing levels of \textbf{self-selection} and evaluated on unbiased data. The fair baseline is indicated by a dashed horizontal black line.}\label{fig:comp_selectLow}
    \end{figure}

\begin{figure}[h]
    \centering
        \begin{minipage}[c]{.49\textwidth}
            \begin{flushright}
            \includegraphics[height=4.025cm]{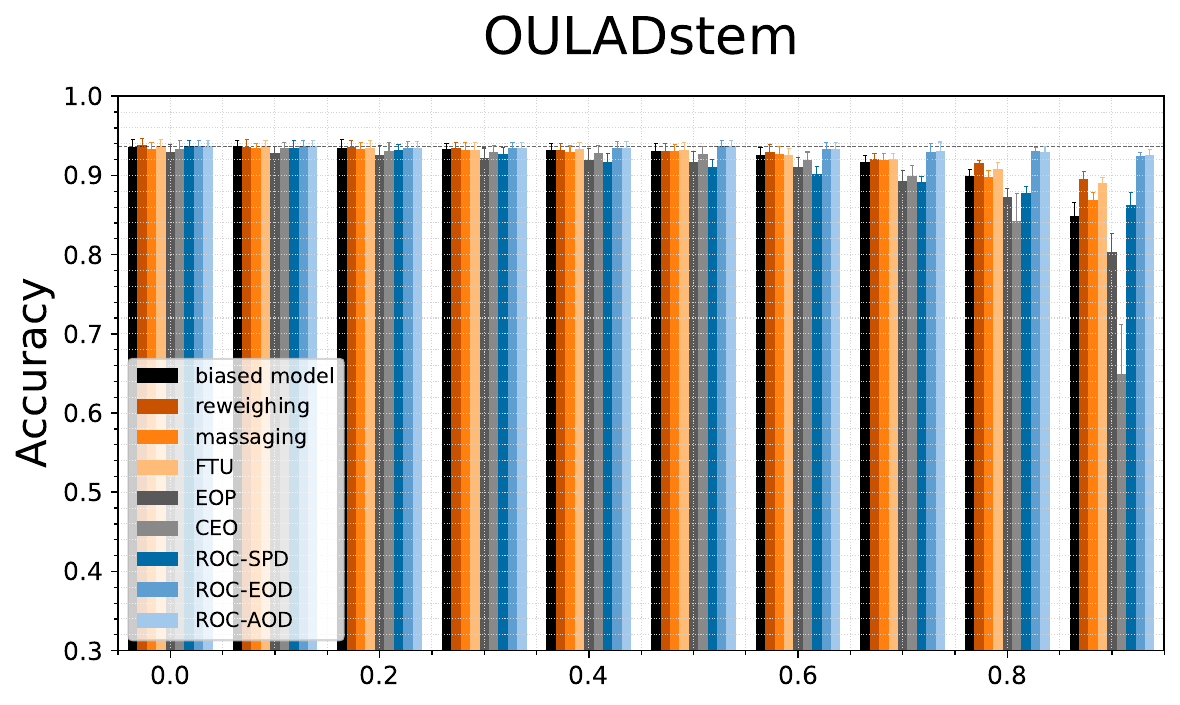}
            \end{flushright}
        \end{minipage}
        \begin{minipage}[c]{.49\textwidth}
            \begin{flushright}
            \includegraphics[height=4.025cm]{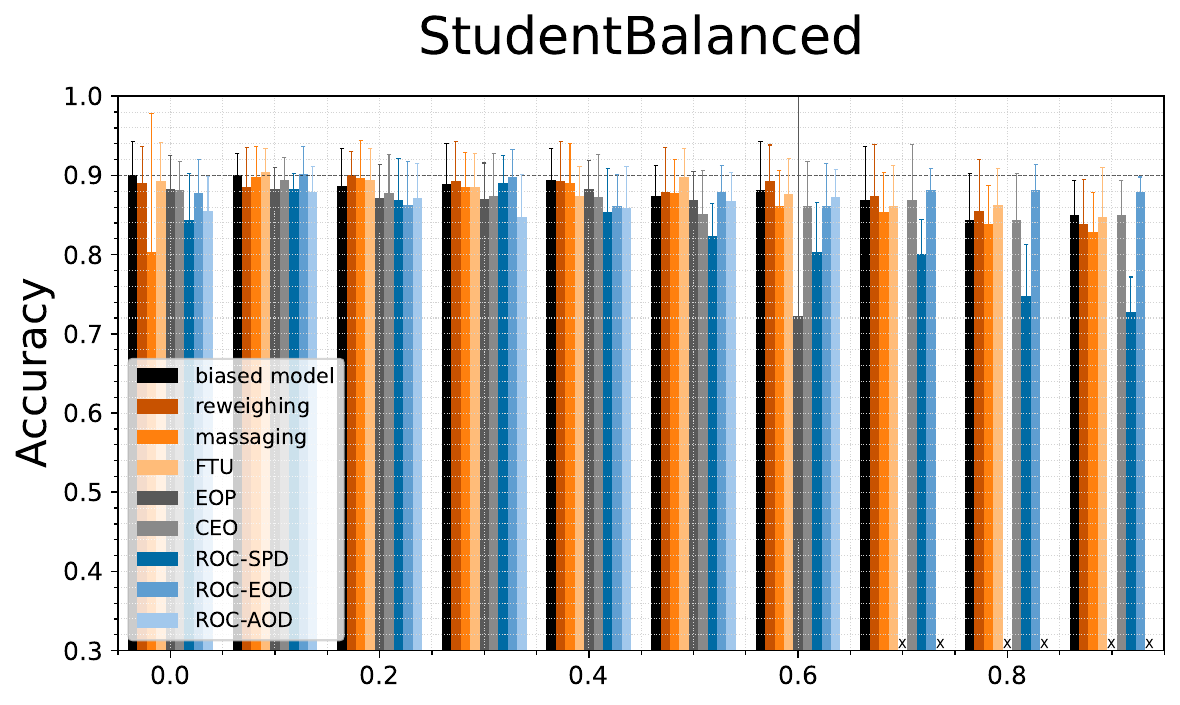}
            \end{flushright}
        \end{minipage}
        \\
         \begin{minipage}[c]{.49\textwidth}
            \begin{flushright}
            \includegraphics[height=3.55cm]{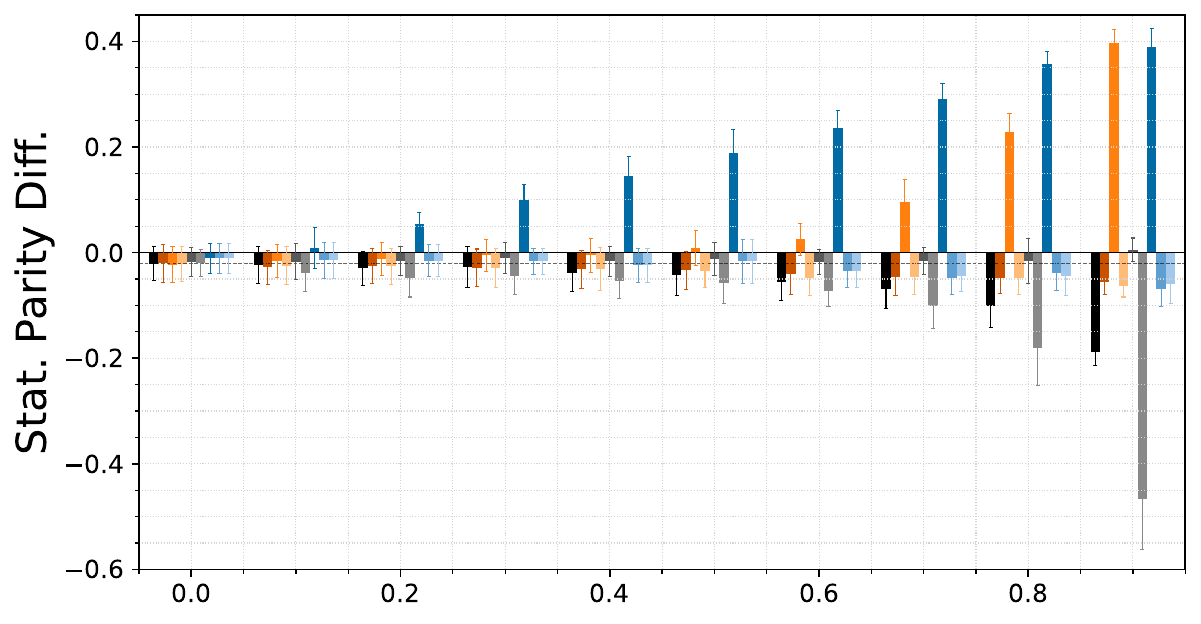}
            \end{flushright}
        \end{minipage}
        \begin{minipage}[c]{.49\textwidth}
            \begin{flushright}
            \includegraphics[height=3.55cm]{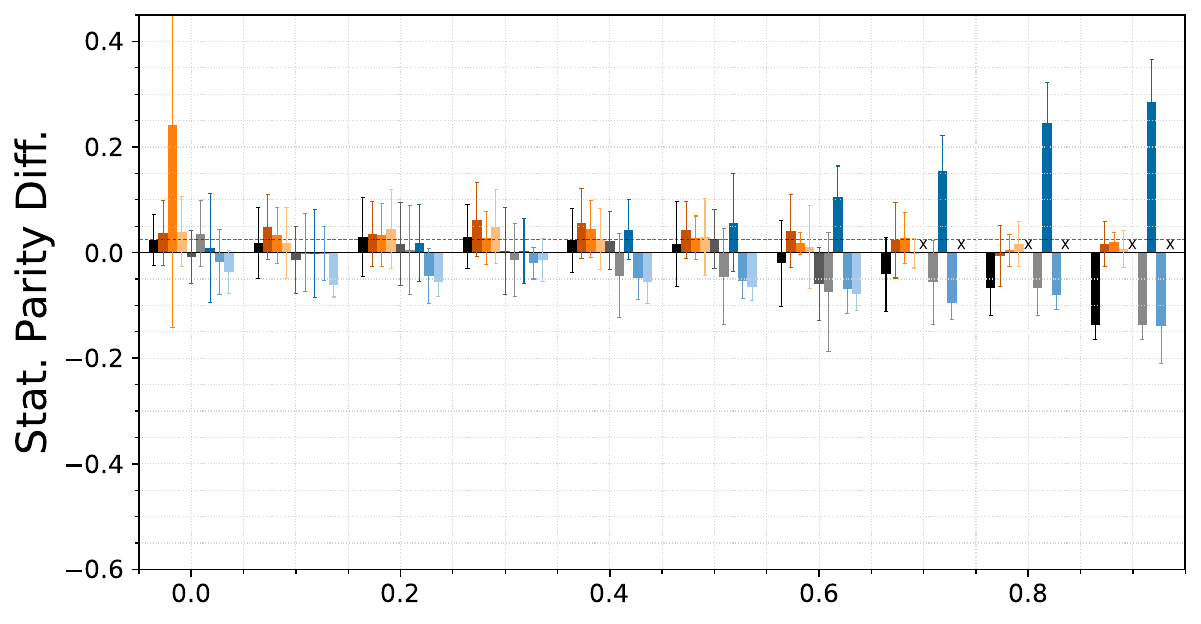}
            \end{flushright}
        \end{minipage}
        \\
        \begin{minipage}[c]{.49\textwidth}
            \begin{flushright}
            \includegraphics[height=3.55cm]{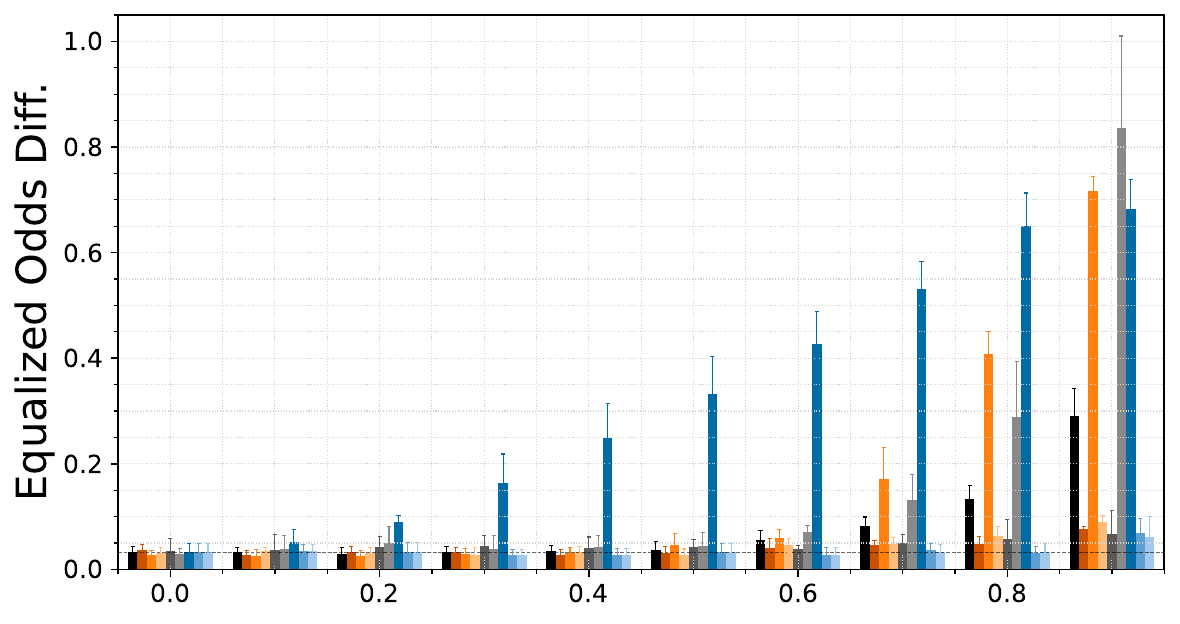}
            \end{flushright}
        \end{minipage}
        \begin{minipage}[c]{.49\textwidth}
            \begin{flushright}
            \includegraphics[height=3.55cm]{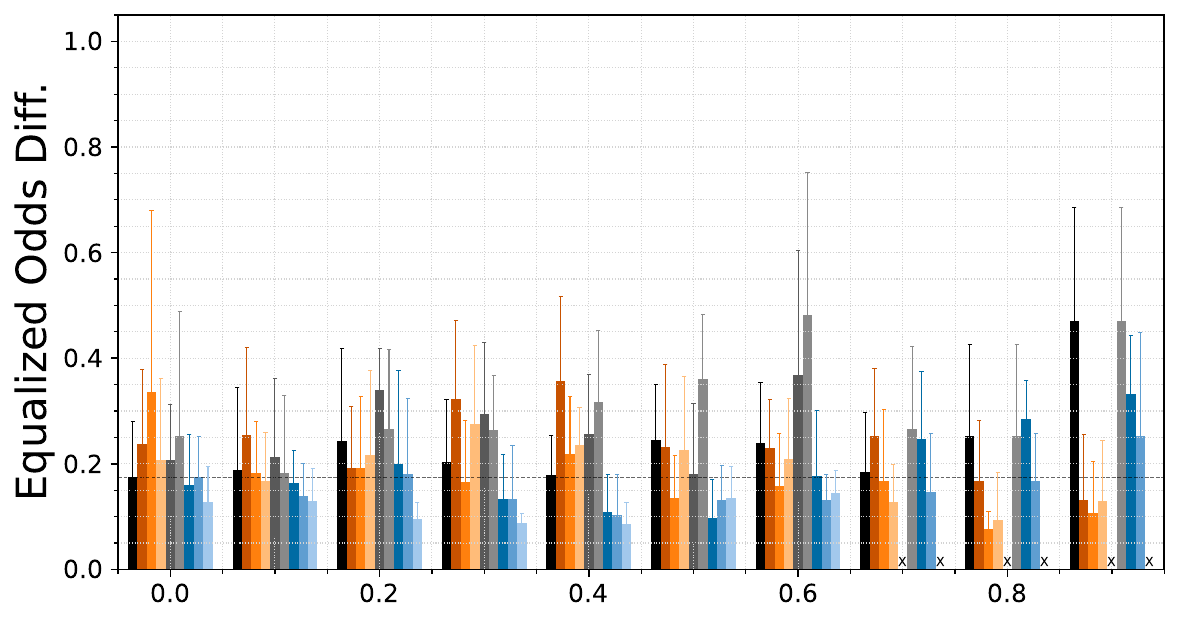}
            \end{flushright}
        \end{minipage}
        \\
        \begin{minipage}[c]{.49\textwidth}
            \begin{flushright}
            \includegraphics[height=3.85cm]{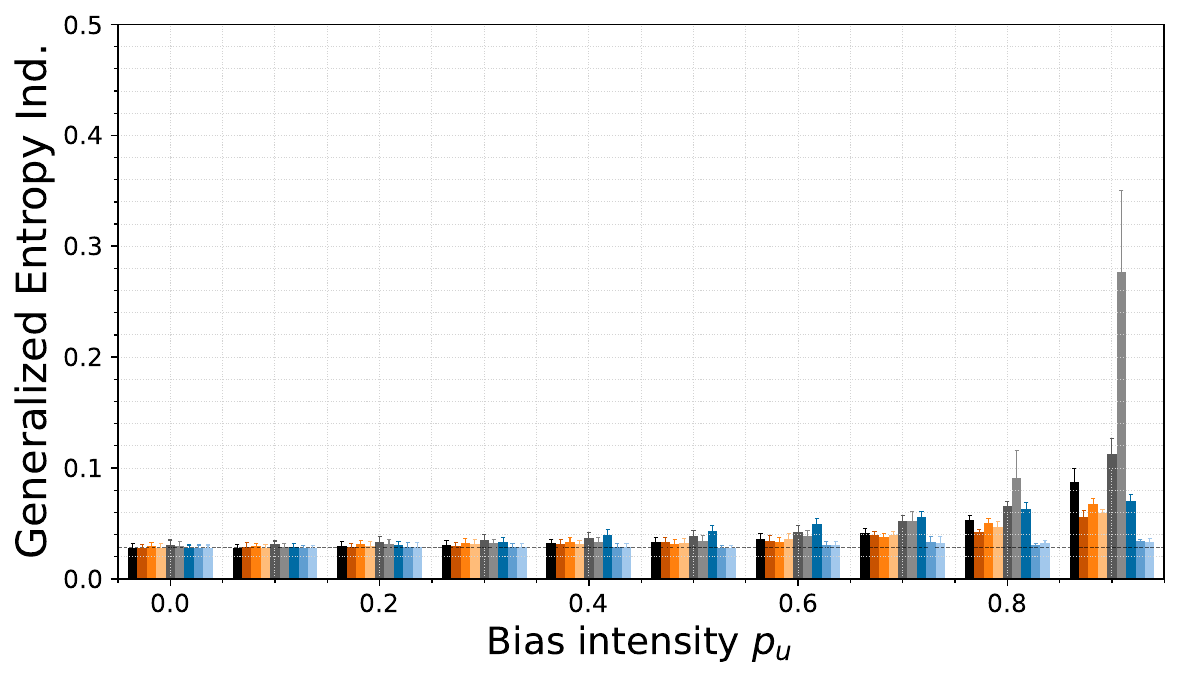}
            \end{flushright}
        \end{minipage}
        \begin{minipage}[c]{.49\textwidth}
            \begin{flushright}
            \includegraphics[height=3.85cm]{figures/comp/studentBalanced_selectDoubleProp_GenEntropyIndex.pdf}
            \end{flushright}
        \end{minipage}
    \caption{Evolution of accuracy and fairness metrics for RF models trained on data with increasing levels of \textbf{malicious selection} and evaluated on unbiased data. The fair baseline is indicated by a dashed horizontal black line.}\label{fig:comp_selectDouble}
    \end{figure}

        The results for label bias are displayed in Figure \ref{fig:comp_label}, for self-selection in Figure \ref{fig:comp_selectLow} and for malicious selection in Figure \ref{fig:comp_selectDouble}. Each figure presents the results for OULADstem and Student, and for the four evaluation metrics presenting the most interesting results. Fairness is indicated by values of accuracy close to 1 and values of EqOd, SPD and GEI\footnote{Extremely high values of GEI are not displayed to maintain readability, but can be consulted in the experiment directory.} close to 0. The results for OULADsocial and BCC can be found in Appendix \ref{app:mitig_comp}. If a method failed to produce a result for over half of the folds, it is indicated by a cross. Note that the scales are not identical for all plots.

        \begin{table}[h]
            \begin{tabular}{|c|c|c|c|c|c|c|c|c|}
            \hline
            Bias &     Reweighing &  Massaging  &   FTU     &    EOP    &    CEO    &   ROC-SPD    & ROC-EqOp & ROC-AvOd \\\hline
            Label &   \ding{51}   & \ding{51}  & \ding{51} & \ding{55} & \ding{55} & \ding{51}\ding{51} & \ding{81} & \ding{81}   \\
Malicious select. & \ding{51}\ding{51} & \ding{55}\ding{55} & \ding{51}\ding{51} & \ding{55} & \ding{55}\ding{55} & \ding{55}\ding{55} & \ding{81} & \ding{81} \\
Self-select. & \ding{51}\ding{51} & \ding{55}\ding{55} & \ding{51}\ding{51} & \ding{55} & \ding{55}\ding{55} & \ding{55}\ding{55} & \ding{55}\ding{55} & \ding{55} \\\hline
            \end{tabular}
            \caption{Overview of methods ability to mitigate each bias type. \ding{51} indicates an overall positive impact (excluding results for the highest and less realistic bias levels). The symbol is doubled when the mitigated results are close to the fair baseline. \ding{55} indicate a mostly negative or neutral effect. The symbol is doubled when the mitigated models introduce significant unfairness. \ding{81} indicates results that may be positive or negative depending on the dataset and bias level.}
            \label{tab:method_performance}
        \end{table}

        \subsubsection{Reweighing} Because of its design, reweighing should theoretically be efficient at reverting selection bias, even though it aims to achieve a biased evaluation of SPD=0 on the training set. On the other hand, while the method is not designed to correct label bias perturbations, that bias type does not perturb the measurement of its objective.
        Looking at Figures \ref{fig:comp_selectLow} and \ref{fig:comp_selectDouble}, we see that reweighing has indeed a very positive impact on both subtypes of selection bias. In all scenarios, it either improves accuracy or maintains it close to the unmitigated value. Despite its use of biased SPD evaluation, SPD values are kept low. There is a light negative impact for low and medium bias levels in the case of StudentBalanced, but SPD values are otherwise close to or lower than the fair baseline.
        We assume that the effect of selection bias on SPD could be mitigated by the fairness-agnostic training algorithm since reweighting did not perturb the labels. We also note an absence of trade-off between all evaluation metrics, since EqOd and individual fairness are either maintained or improved by reweighting, with only a few exceptions (mainly regarding EqOd in the malicious scenario for StudentBalanced).
        
        For label bias, the effect of reweighing is mostly positive, although several other methods provide better results. Its impact on accuracy and GEI is also influenced by the bias level and dataset, with a negative impact in case of higher bias intensity.

        \subsubsection{Massaging} This method should theoretically perform well to mitigate label bias since it flips training labels to optimize SPD, which is robust to that bias type. It does have an overall positive impact, although not as significant as ROC-SPD, its post-processing equivalent. Massaging significantly improves SPD and EqOd in all situations (except in the outlier case of StudentBalanced at x=0\footnote{We consider the results for massaging trained on StudentBalanced with no bias as an outlier caused by three folds where the discrimination was reversed and the unprivileged group did not receive any negative outcome.}), but its results on accuracy are moderate at best and slightly negative in a few instances. The results for individual fairness depend on the dataset and bias level, even though they are mostly positive.
        
        In the case of selection bias, massaging gives very poor results because modifying the accurate labels, especially to achieve a biased statistical parity goal, corresponds to the introduction of label bias in the training set.
        In the case of malicious selection, the preprocessed models show "reverse" unfairness that increase with the bias level for the larger dataset OULADstem.
        The steady increase in group unfairness is caused by the optimization of an increasingly biased evaluation of SPD, which leads to an over-correction of the existing unfairness. This negative effect is however rather limited for the StudentBalanced dataset.
        For self-selection, massaging introduces high "reverse" unfairness in all situations. The negative effect is stronger in this case and is caused by more than the biased metric evaluation. 
        The method only changes the labels of the unprivileged individuals from negative to positive and of the privileged ones from positive to negative, which can only increase, or at best maintain, the unfairness caused by the self-selection bias.
    
        \subsubsection{FTU} This method gives overall good performances for all biasing scenarios and is oftentimes among the best performing methods. Only when label bias affects most of the dataset is it detrimental to accuracy and GEI. 
        FTU is efficient in our scenarios because the label is independent from the sensitive attribute in the original unbiased data. Those results confirm and go beyond those in \cite{baumann_biasOnDemand_2023, favier_biasTheoretical_2023}, where FTU is considered a good option to address label bias under that independence condition. If the correlation between the sensitive attribute and other features had been higher, we expect FTU to have been less efficient \cite{pedreshi_discrimination-aware_2008}.
    
        \subsubsection{EOP}
        Since EOP tries to satisfy EqOd as evaluated on the biased training set, it is expected be vulnerable to label bias, as well as selection bias, although to a lesser degree, as seen in Section \ref{sec:biasVSunbiased}. 
        The results show that EOP provides some improvement for EqOd and SPD in several instances, but its impact on EqOd is mostly negative for lower bias levels. As expected, the results for selection bias are overall better regarding EqOd and SPD values, although there are several situations where the bias is increased. For all bias types, in the few cases in which EqOd and SPD are improved, it is usually at the expense of accuracy and individual fairness, meaning that EOP does not allow to retrieve the fair distribution.
        
        With the implementation based on predicted binary labels, EOP fails to compute a post-processed model in the corner cases where the unmitigated predictions for the validation set lead to one of the group not receiving any positive or negative predictions. An implementation based on predicted scores, as also presented in \cite{hardt_equalityOpp_2016}, would be more robust to these corner cases.
    
        \subsubsection{CEO} Similarly to EOP,  the CEO method should also be impacted by both label and selection bias since it relies on a fairness notion related to EqOd. When applied on label bias, CEO has an overall negative or neutral impact. 
        In the case of malicious selection, the steadily increasing negative impact on EqOd, SPD and accuracy suggests that it is caused by the realization of a fairness objective that is increasingly skewed by the selection bias, like with SPD, but in this case not associated with the reversal of the discrimination direction. The results for self-selection show more variations depending on the dataset and bias level, but CEO never leads to improvements compared to the unmitigated results. The high bias increase for OULADstem stops before the highest bias values, at which point there is no or almost no unprivileged individuals with negative outcome in the validation dataset used to postprocess the models. 

        \subsubsection{ROC-SPD} Since ROC-SPD changes predicted labels to optimize SPD, it can be seen as the post-processing equivalent of massaging and should also theoretically be efficient against label bias. It is indeed the best performing method for moderate and high label bias levels, although in StudentBalanced it is detrimental to accuracy when the bias level is low. Individual fairness metrics can also be impacted negatively in some cases (high bias levels in OULADstem and low bias level in StudentBalanced).

        These overall positive results do not translate to either subtype of selection bias.
        For self-selection, the amount of unfairness introduced by ROC-SPD depends on the bias level, dataset and choice of fairness metric. The effect on accuracy is particularly negative.
        ROC-SPD and massaging use the same mechanism of changing label values to optimize a biased evaluation of SPD, but the results of the two methods show significant differences. When the data is only perturbed by selection bias, massaging changes a number of training labels with previously accurate values, directly introducing label bias in the training set. Since selection bias has only a limited effect on model's predictions, as seen in Section \ref{sec:res_model}, ROC-SPD needs to modify a much lower number of predicted labels, likely including some inaccurate ones. This difference explains why the negative impact of ROC-SPD is much lower than that of massaging in the case of self-selection, where both models are only able to increase the existing discrimination because of the direction in which the labels are changed.
        For malicious selection, ROC-SPD leads to increasing "reverse" unfairness that can be linked to the optimization of biased SPD value, like is the case for massaging. As our results show, that excessive correction is more impactful when performed as a post-processing step. 
        
        \subsubsection{ROC-EqOp} This method is designed to revert label bias, but should lead to predictions corresponding to biased equality of opportunity. The results for the label bias scenario show that it is indeed influenced by that bias, while still improving or maintaining EqOd and SPD for all datasets. The impact on the other metrics is heterogeneous across datasets and bias levels.
       
        In the case of selection bias, the results vary according to the bias subtype, the dataset and the biasing intensity, often having little effect and sometimes introducing significant new bias. Only for the malicious selection in OULADstem is the effect significantly positive, and so for all metrics. On the other hand, in the case of self-selection, ROC-EqOp highly increases unfairness for both EqOd and SPD in all datasets. It is because, like massaging, ROC methods only change label in a direction that reinforce the discrimination caused by self-selection.
        
        Since this method relies on EqOp, which is computed with binary predicted labels, it fails to produce a post-processed model when the unmitigated predictions for the validation set do not contain any positive label for one of the groups.
    
        \subsubsection{ROC-AvOd} 
        This technique is really close to ROC-EqOp, as they both enforce a relaxed version of EqOd, which explains why the results of these two methods are close to each other. The differences, mostly visible in the results for label bias, are explained by the fact AvOd is a less strict constraint on both TPR anf FPR, while EqOp is a stronger constraint on TPR only.
        This is also the reason why ROC-AvOd fails to produce a mitigated model in more situations than ROC-EqOp, since AvOd cannot be computed when one of the group in the validation set does not present one of the label, be it positive or negative.
       
        \subsubsection{Discussion}
        
        \paragraph{\textbf{Different bias (sub)types require different bias mitigation methods to be successfully mitigated.}}
        It is apparent in our empirical results that the type of bias present in the training and validation data influences the efficiency of bias mitigation methods, with performance changing for label and selection bias. This complements the results in \cite{baumann_biasOnDemand_2023, favier_biasTheoretical_2023, wick_unlocking_2019}.
        We can also observe that \textbf{different kinds of selection bias can lead to very different results}. Understanding the bias present in the training dataset and taking it into account when choosing a bias mitigation method is thus important to lead to more effective fairness interventions. 
        
        \textbf{To select an appropriate method for a given bias type, both the fairness criterion they use and the assumptions on which their design is based need to be considered}. In that regard, the following conclusions can be drawn from our experiments :
        \begin{itemize}
            \item Reweighing can work well for different types of selection bias, while methods based on both SPD and relabeling fail to recover the unbiased distribution.
            \item Methods based on EqOd or its relaxed variations are affected by both label and selection bias, which leads to the mitigated models still retaining some bias. 
            \item Methods based on relabeling in only one direction, for example from positive to negative label for the privileged group and negative to positive for the unprivileged group, can exclusively correct the specific types of bias they have been designed for. If the data exhibits a bias type that goes against their underlying assumption, the methods can increase the existing bias, as is the case for massaging and ROC that are designed for malicious selection and largely increase unfairness in the presence of self-selection.
        \end{itemize}
        
        In order to provide a more comprehensive and applicable set of guidelines, further research is needed about other sources of bias, more complex scenarios involving the combination of different bias types, techniques to detect them in a given dataset, as well as additional bias mitigation methods and their potential combination.
        
        \paragraph{\textbf{The efficiency of bias mitigation methods also varies according to other dataset characteristics.}} Our results suggest that relevant characteristics include the proportion of unprivileged and favored groups, the base rate, and the bias intensity, with for example the same method decreasing fairness for lower bias levels while improving it for higher ones. A limitation of our work is that the three datasets we use differ from each other on several parameters. A more controlled experimental setting would thus be necessary to determine the respective impact of each factor and which methods would be more efficient in a specific situation.

        \paragraph{\textbf{Bias mitigation does not lead to a fairness-accuracy trade-off under the fair world assumption.}}
        The results in this section empirically confirm that there is no trade-off between fairness and accuracy when their measure is unbiased, as several mitigation methods could successfully improve both.
        Since the underlying assumption of most fairness interventions is that the training data is not representative of the desired outcome because of unwanted biases, the evaluation performed using that data should also be considered as unrepresentative of the desired model performance. It follows that optimizing the trade-off between biased evaluation of accuracy and fairness does not provide any guarantee that the resulting model will give predictions in accordance with the fair criterion that should be enforced. It can even be detrimental to the achievement of both goal, since it forces the model predictions to stay close to the biased available dataset, which is what a high accuracy represents when it is measured with a biased test set. 
        We thus argue that the reliance on biased metrics and the fairness-accuracy trade-off framework in fairness research has the potential to hinder the development and use of efficient fairness interventions.

        \paragraph{\textbf{Under our fair world assumptions, there is no trade-off between individual and group fairness, nor between statistical parity and equalized odds.}} Several mitigation methods could simultaneously improve all the fairness metrics we used, whether they were based on individual or group fairness. 
        Furthermore, statistical parity and equalized odds always improved together, as long as they were not measured using a biased ground truth. This result is rather intuitive when considering the fair world framework \cite{favier_biasTheoretical_2023} introduced in Section \ref{sec:method}. If a method is able to correctly mitigate the bias present in the training set, the resulting model gives predictions that should differ from a test set presenting the same bias, but that are a closer to the fair world distribution. Hence, all evaluation measures should improve when the model is evaluated using a test set that represents that fair distribution.\\
 
        With the new hindsight gained from fair evaluation on distinct bias types, our work highlights several elements that should be taken into account when choosing bias mitigation methods. On the one hand, a better understanding of the different factors and their influence on fairness interventions would allow to develop better guidelines on which methods should or should not be used in a specific situation. On the other hand, our results, and future work in the same direction, help provide pointers towards development of new bias mitigation methods that would be more robust to training data bias.

\section{Conclusion}\label{sec:ccl}
    
    In this work, we study the influence of label bias and different manifestations of selection bias on the evaluation of classification models, on their behavior, and on the efficiency of several bias mitigation methods. Doing so, we introduce a novel biasing and evaluation framework to model biased versions of a fair world by introducing controlled bias in an existing real-world dataset that is sufficiently fair. In addition to studying specific bias types, this allows to train models using biased data while evaluating their performance on a test considered to be unbiased.
    
    Our results point towards several factors that influence wether bias in the training data leads to significant or negligible impact on models. They also show an absence of trade-off between fairness and accuracy, and between group and individual fairness, when the evaluation is performed using a test set that does not exhibit unwanted bias. We further discuss how label and selection biases each influence the values of fairness metrics. We thus encourage the use of fairer evaluation methods rather than the traditional use of a test set exhibiting the same bias as the train set.
    Our experiments also empirically confirm that the performance of bias mitigation methods are influenced by the type of bias present in the data. We discuss these results for each of the methods studied. Our findings call for future work to study other and more complex types of bias injections, as well as other factors that impact the efficiency of mitigation methods.


\bibliographystyle{ACM-Reference-Format}
\bibliography{bibliography}

\appendix

\newpage
\section{Additional results comparing fair and biased evaluations}\label{app:biasVSunbiased}

    \begin{figure}[h]
    \centering
        \begin{minipage}[c]{.45\textwidth}
            \centering
            \includegraphics[width=\textwidth]{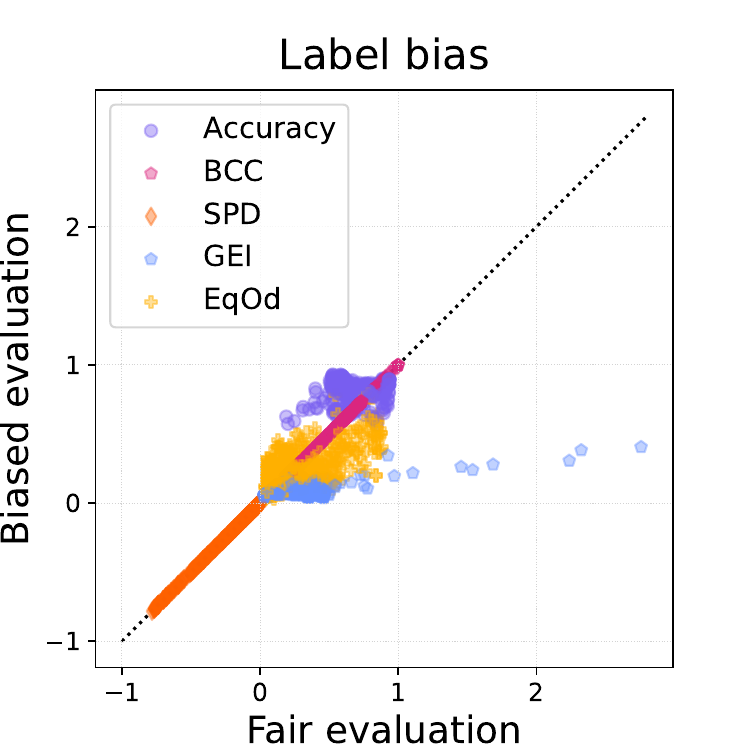}
        \end{minipage}
        \begin{minipage}[c]{.45\textwidth}
            \centering
            \includegraphics[width=\textwidth]{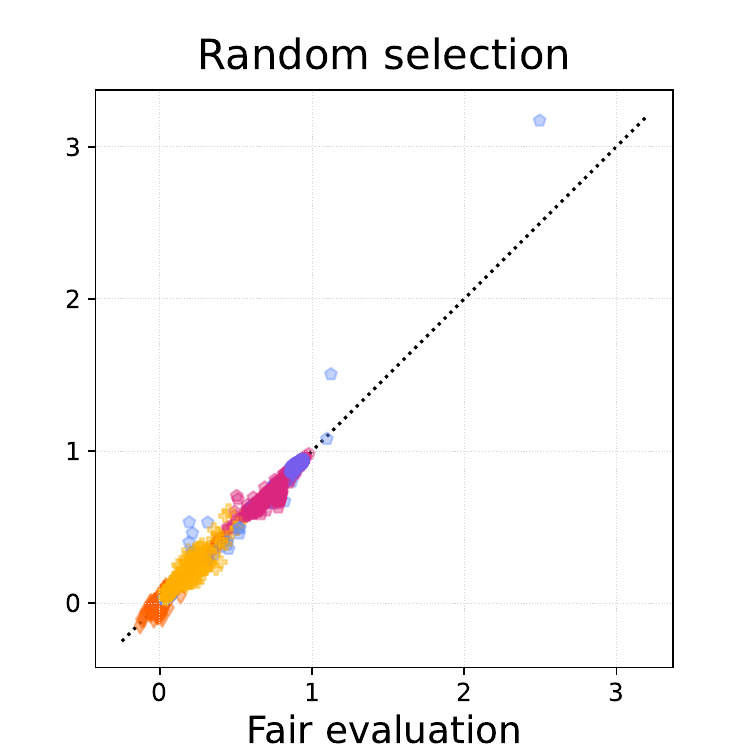}
        \end{minipage}
        \\
        \begin{minipage}[c]{.45\textwidth}
            \centering
            \includegraphics[width=\textwidth]{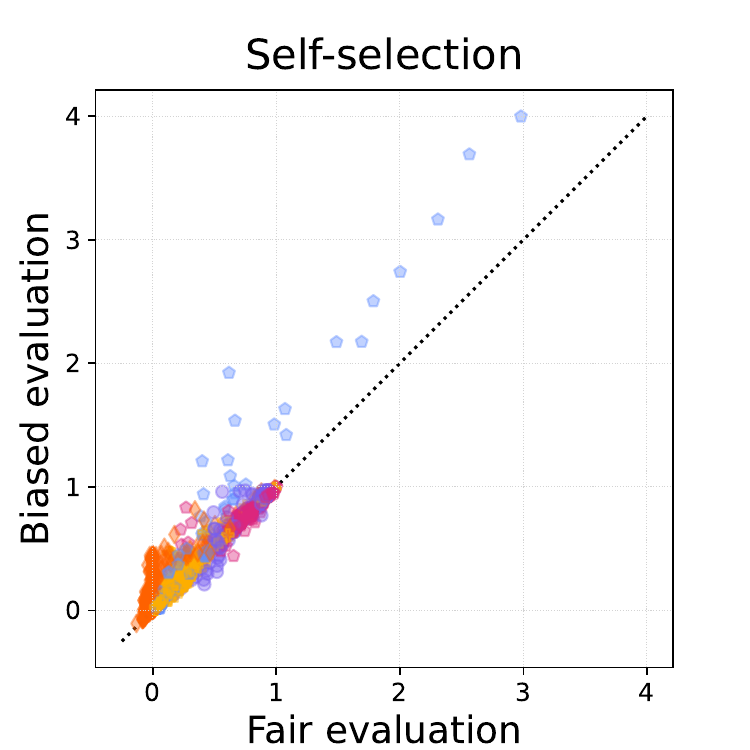}
        \end{minipage}
        \begin{minipage}[c]{.45\textwidth}
            \centering
            \includegraphics[width=\textwidth]{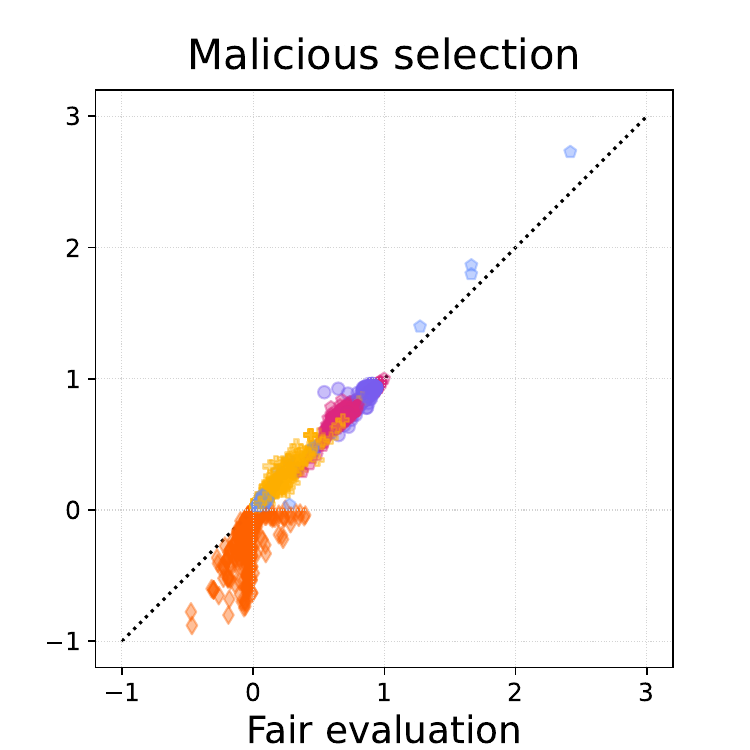}
        \end{minipage}
     \caption{Comparison of metric results evaluated on a biased test set (Biased evaluation) or a fair test set (Fair evaluation). Points that fall outside of the x=y diagonal indicate that the metric measurement is skewed by the bias present in the unfair test set. Are included the results for all four datasets with bias intensities ranging from 0 to 0.9, the three fairness-agnostic training algorithms, the eight bias mitigation procedures, and unmitigated models.}
    \label{fig:BvsF_appendix}
    \end{figure}
\newpage
    \begin{figure}[H]
    \centering
        \begin{minipage}[c]{.35\textwidth}
            \centering
            \includegraphics[width=\textwidth]{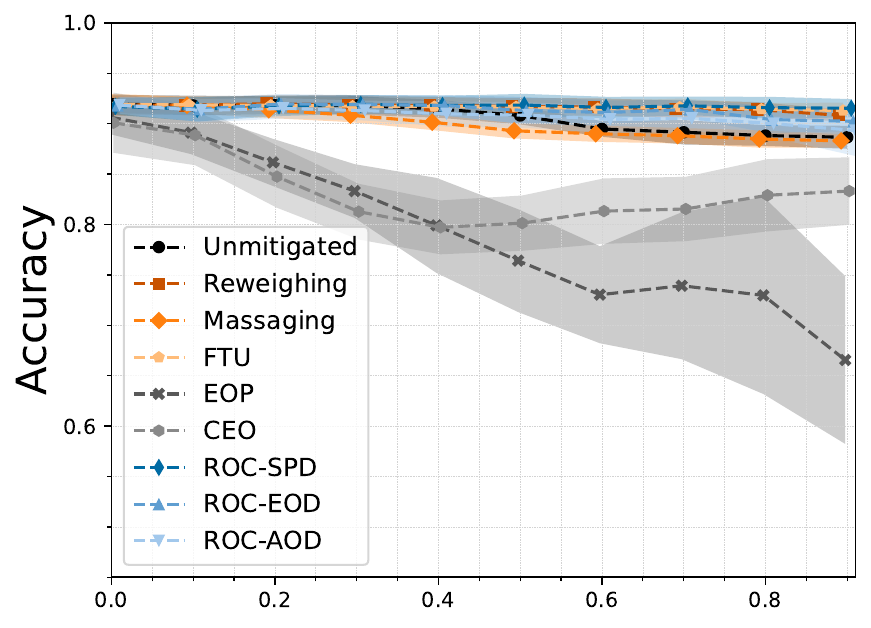}
        \end{minipage}
        \begin{minipage}[c]{.35\textwidth}
            \centering
            \includegraphics[width=\textwidth]{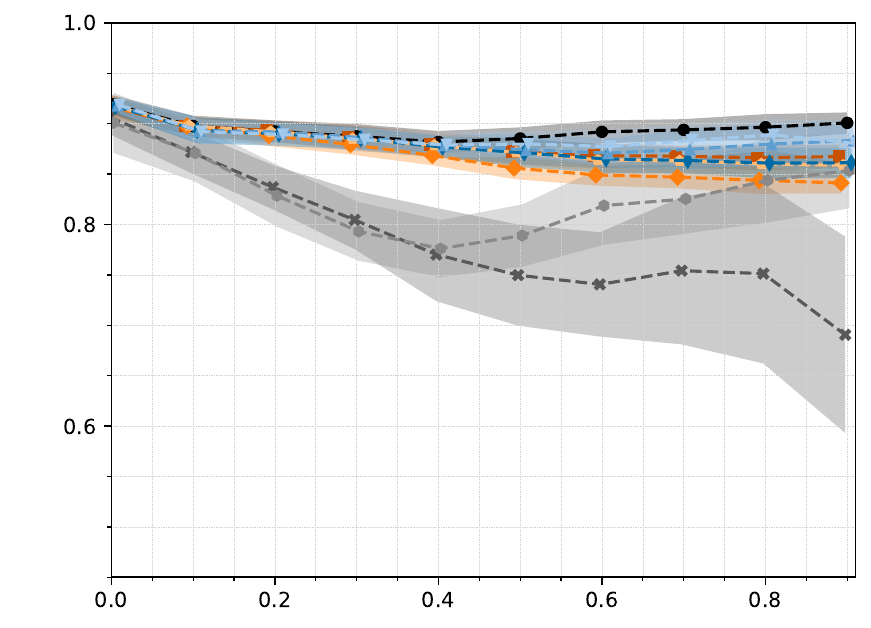}
        \end{minipage}
        \\
        \begin{minipage}[c]{.35\textwidth}
            \centering
            \includegraphics[width=\textwidth]{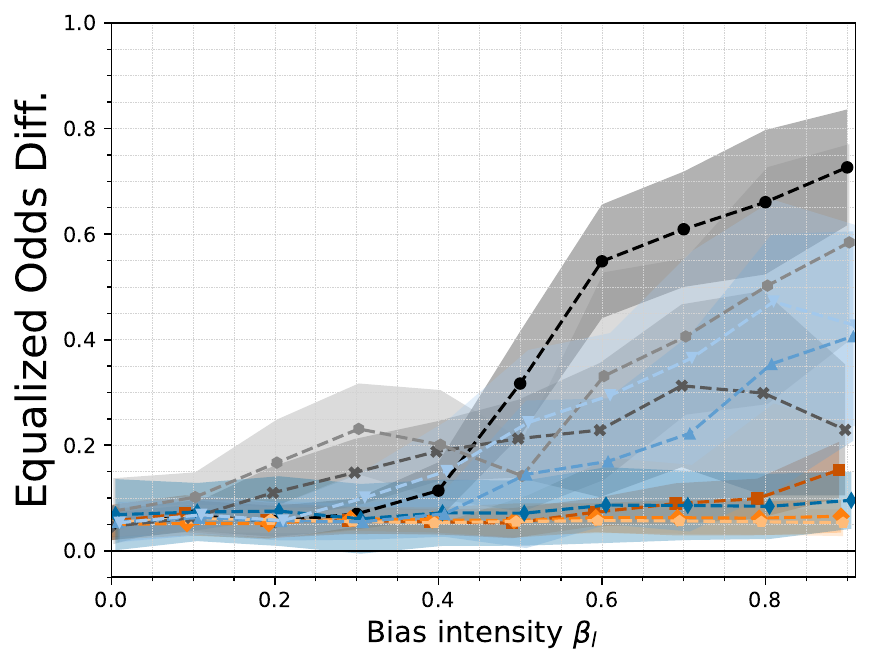}
            \subcaption{Fair evaluation}
        \end{minipage}
        \begin{minipage}[c]{.35\textwidth}
            \centering
            \includegraphics[width=\textwidth]{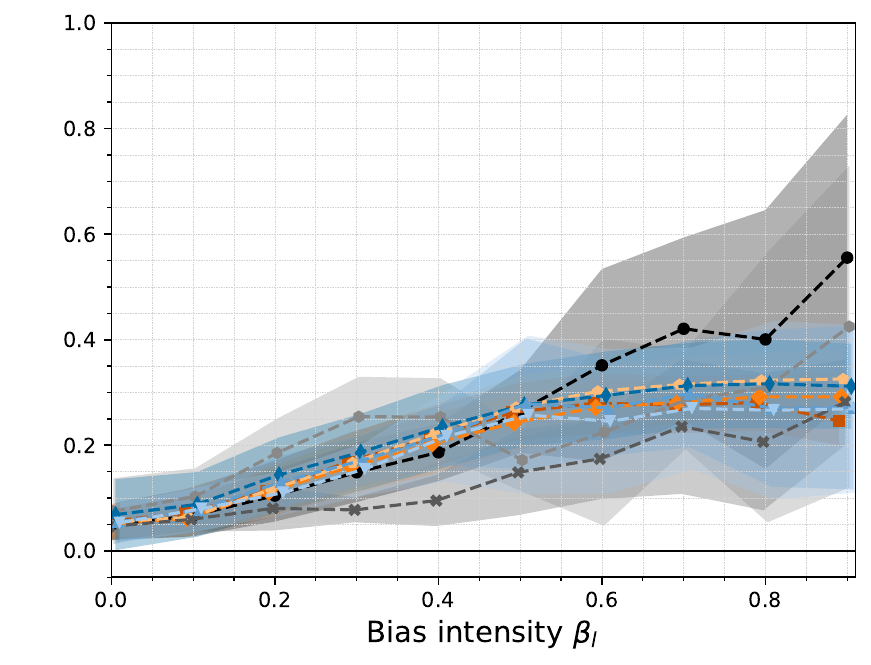}
            \subcaption{Biased evaluation}
        \end{minipage}
    \caption{Fair and biased evaluation of Accuracy and EqOd for increasing levels of \textbf{label bias} injected in \textbf{OULADsocial} training sets.}
    \label{fig:trad_label_social}
    \end{figure}

    \begin{figure}[H]
    \centering
        \begin{minipage}[c]{.35\textwidth}
            \centering
            \includegraphics[width=\textwidth]{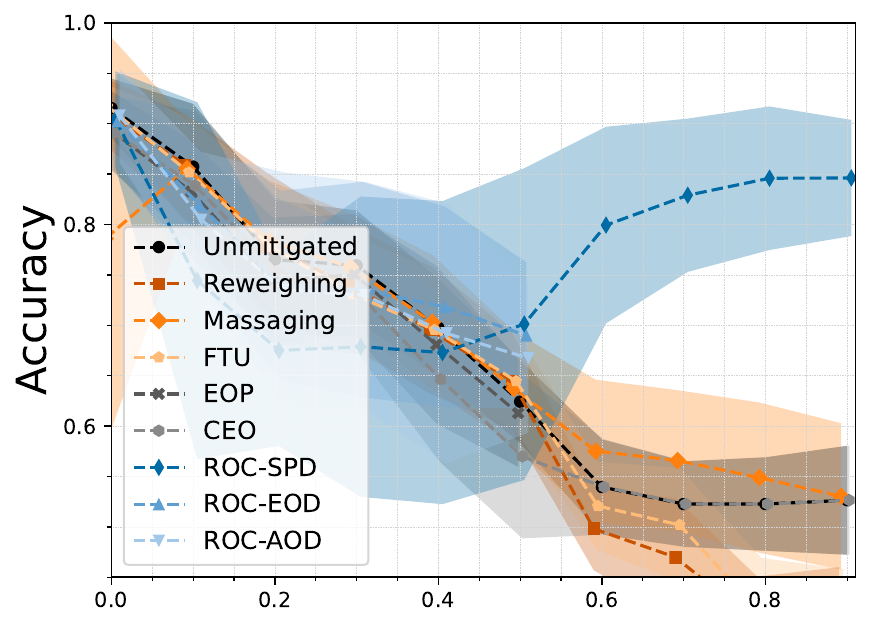}
        \end{minipage}
        \begin{minipage}[c]{.35\textwidth}
            \centering
            \includegraphics[width=\textwidth]{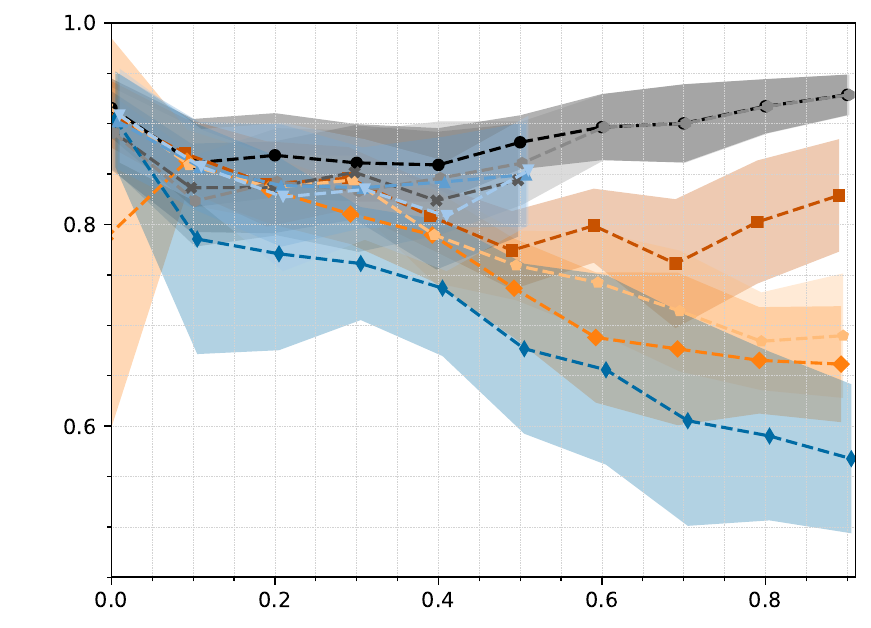}
        \end{minipage}
        \\
        \begin{minipage}[c]{.35\textwidth}
            \centering
            \includegraphics[width=\textwidth]{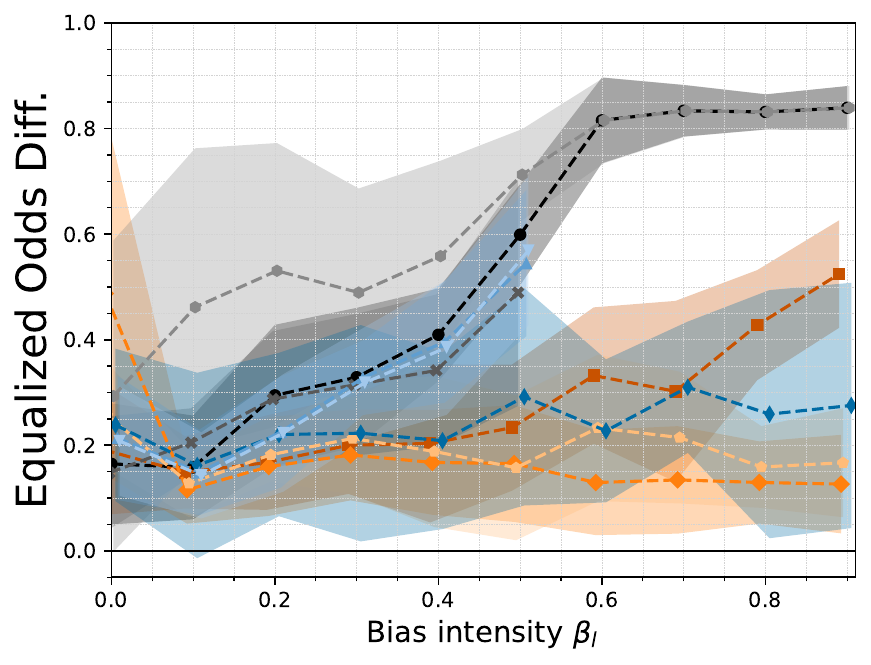}
            \subcaption{Fair evaluation}
        \end{minipage}
        \begin{minipage}[c]{.35\textwidth}
            \centering
            \includegraphics[width=\textwidth]{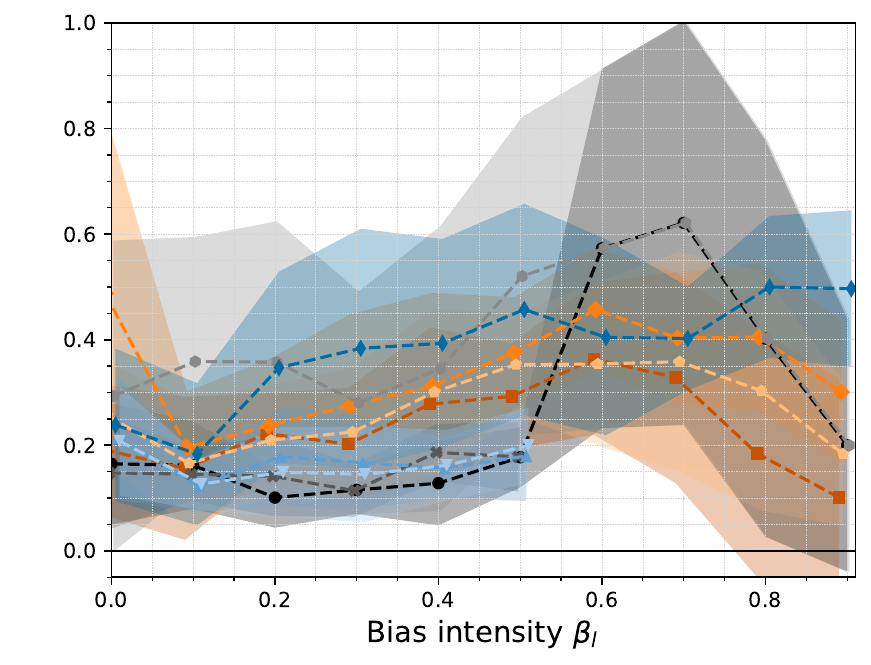}
            \subcaption{Biased evaluation}
        \end{minipage}
    \caption{Fair and biased evaluation of Accuracy and EqOd for increasing levels of \textbf{label bias} in \textbf{StudentBalanced} training sets.}
    \label{fig:trad_label_studentBal}
    \end{figure}
\newpage
    \begin{figure}[H]
    \centering
        \begin{minipage}[c]{.35\textwidth}
            \centering
            \includegraphics[width=\textwidth]{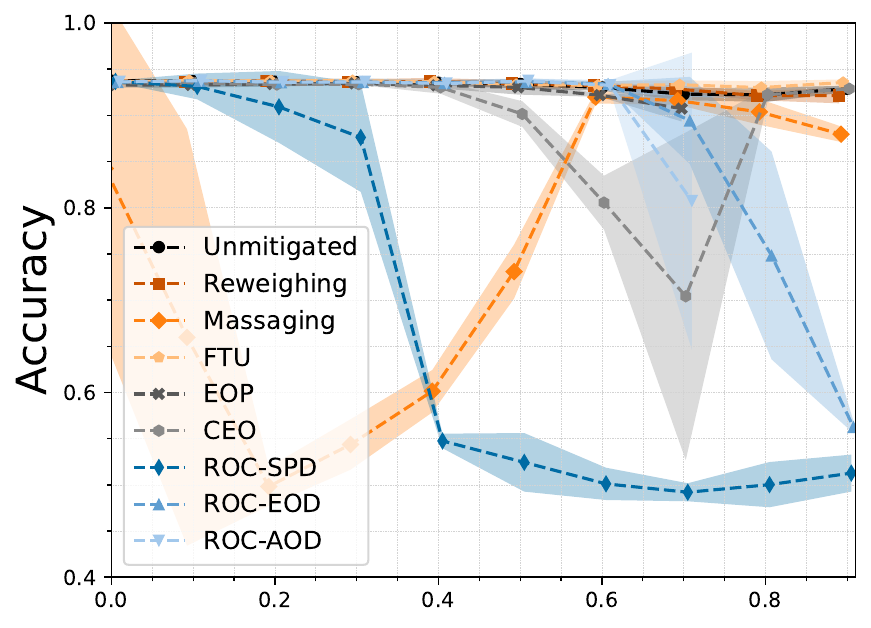}
        \end{minipage}
        \begin{minipage}[c]{.35\textwidth}
            \centering
            \includegraphics[width=\textwidth]{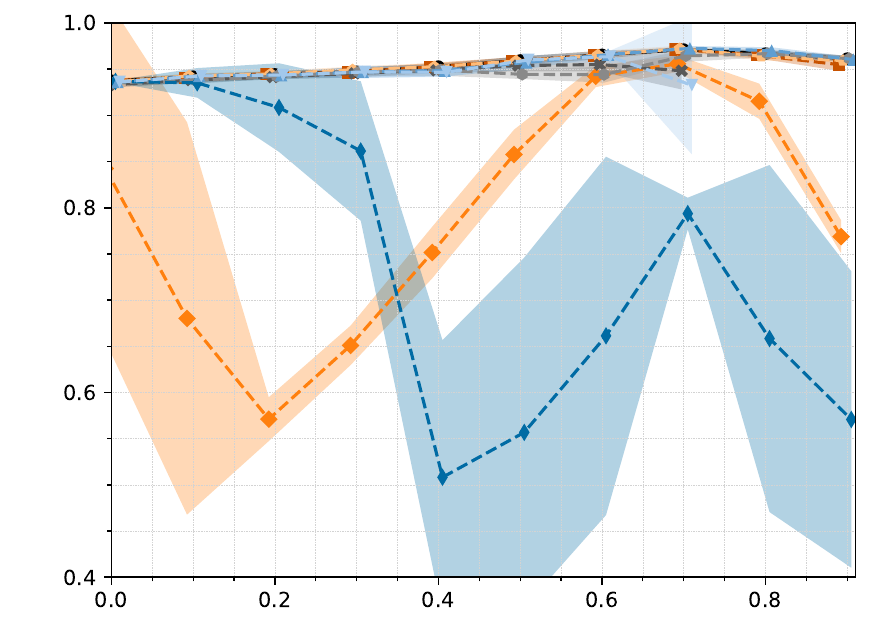}
        \end{minipage}
        \\
        \begin{minipage}[c]{.35\textwidth}
            \centering
            \includegraphics[width=\textwidth]{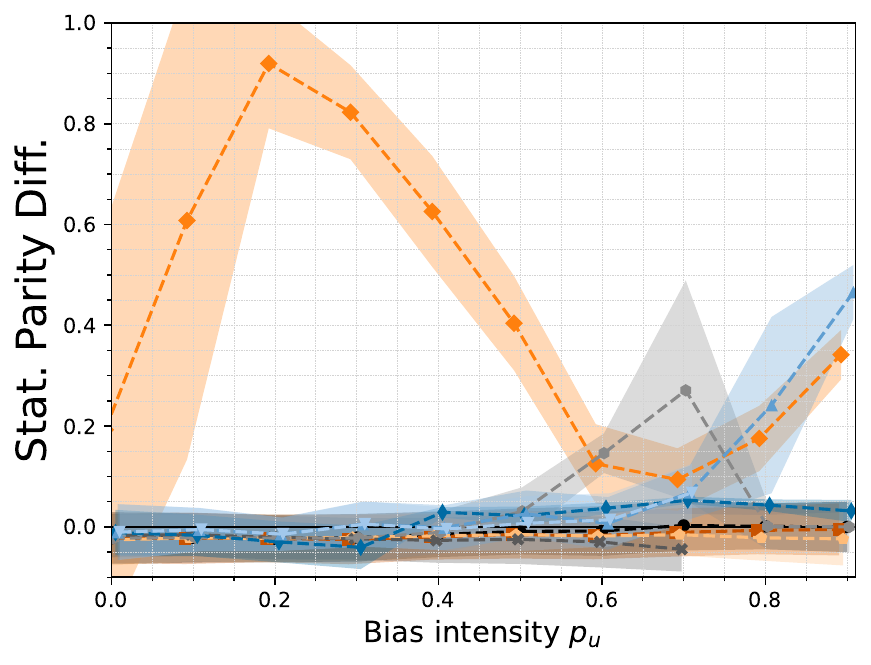}
            \subcaption{Fair evaluation}
        \end{minipage}
        \begin{minipage}[c]{.35\textwidth}
            \centering
            \includegraphics[width=\textwidth]{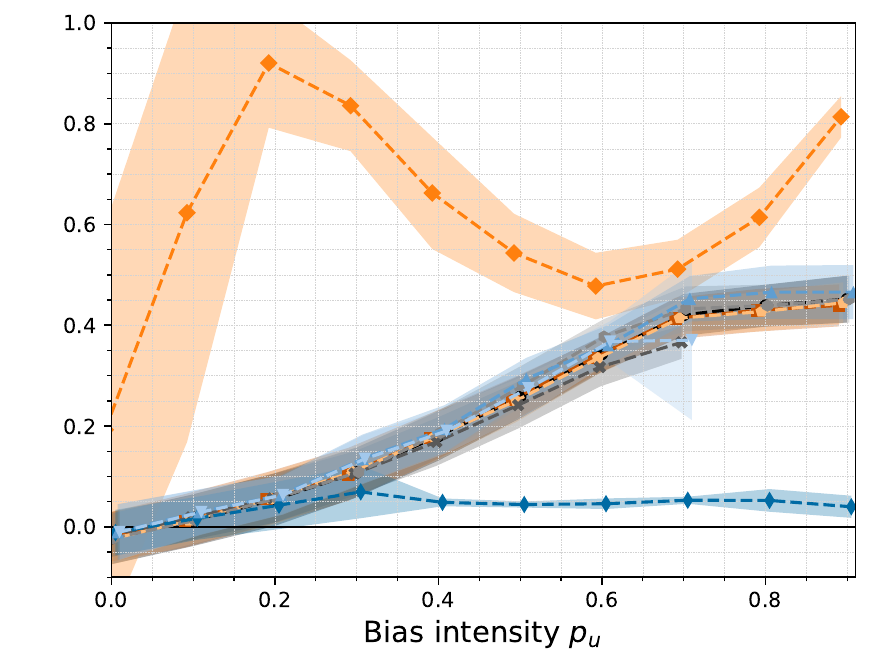}
            \subcaption{Biased evaluation}
        \end{minipage}
    \caption{Fair and biased evaluation of Accuracy and SPD for increasing levels of \textbf{self-selection} injected in \textbf{OULADstem} training sets.}
    \label{fig:trad_selectLow_stem}
    \end{figure}

    \begin{figure}[H]
    \centering
        \begin{minipage}[c]{.35\textwidth}
            \centering
            \includegraphics[width=\textwidth]{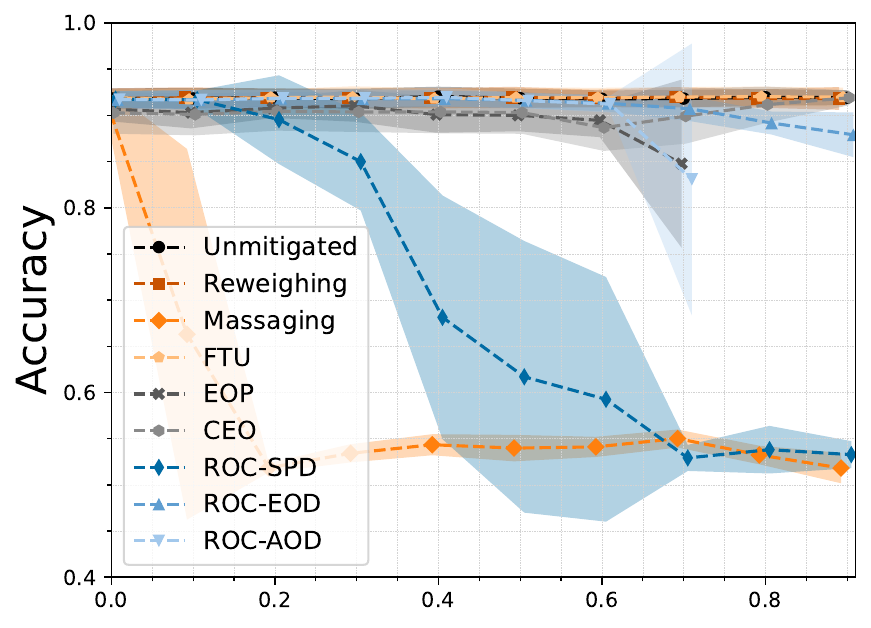}
        \end{minipage}
        \begin{minipage}[c]{.35\textwidth}
            \centering
            \includegraphics[width=\textwidth]{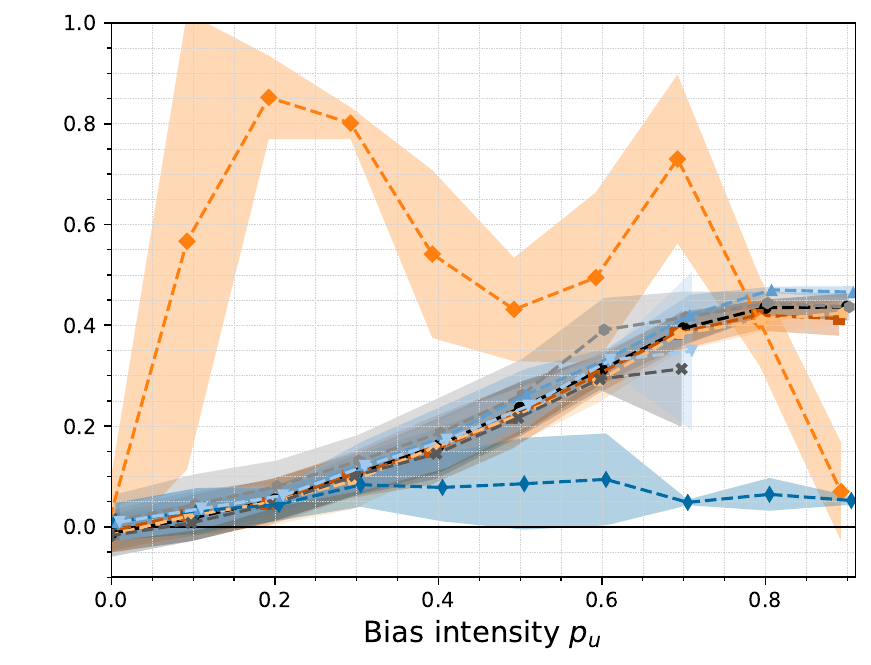}
        \end{minipage}
        \\
        \begin{minipage}[c]{.35\textwidth}
            \centering
            \includegraphics[width=\textwidth]{figures/tradeoff/tradeoff_selectLow_acc_OULADsocial_RF_FairTest.pdf}
            \subcaption{Fair evaluation}
        \end{minipage}
        \begin{minipage}[c]{.35\textwidth}
            \centering
            \includegraphics[width=\textwidth]{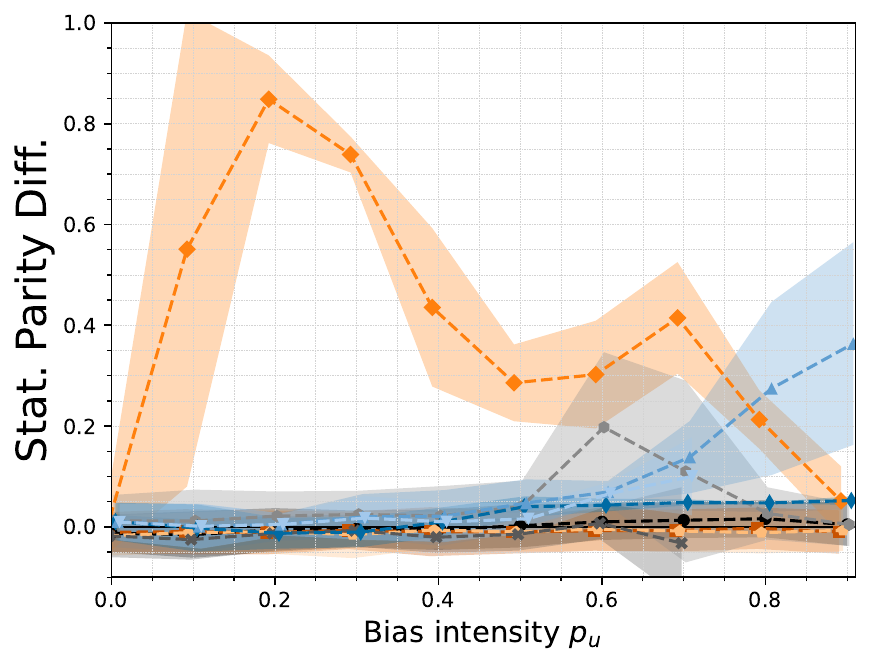}
            \subcaption{Biased evaluation}
        \end{minipage}
    \caption{Fair and biased evaluation of Accuracy and SPD for increasing levels of \textbf{self-selection} in \textbf{OULADsocial} training sets.}
    \label{fig:trad_selectLow_social}
    \end{figure}
\newpage
    \begin{figure}[H]
    \centering
        \begin{minipage}[c]{.35\textwidth}
            \centering
            \includegraphics[width=\textwidth]{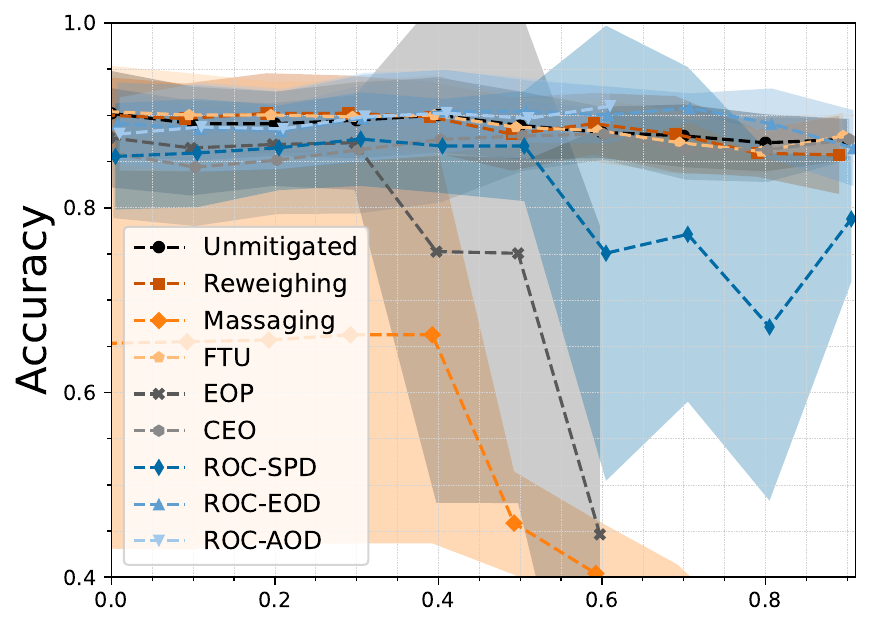}
        \end{minipage}
        \begin{minipage}[c]{.35\textwidth}
            \centering
            \includegraphics[width=\textwidth]{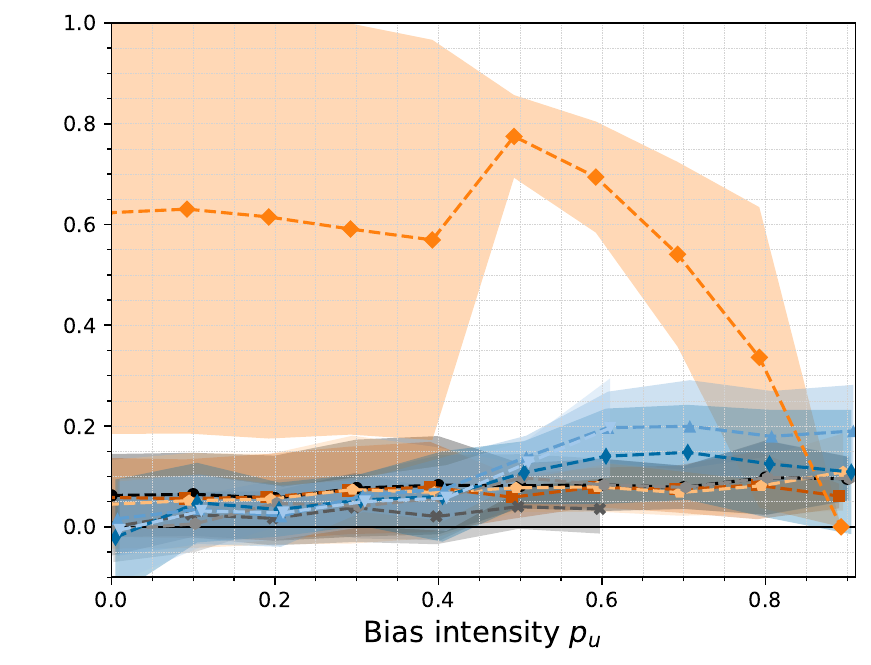}
        \end{minipage}
        \\
        \begin{minipage}[c]{.35\textwidth}
            \centering
            \includegraphics[width=\textwidth]{figures/tradeoff/tradeoff_selectLow_acc_studentBalanced_RF_FairTest.pdf}
            \subcaption{Fair evaluation}
        \end{minipage}
        \begin{minipage}[c]{.35\textwidth}
            \centering
            \includegraphics[width=\textwidth]{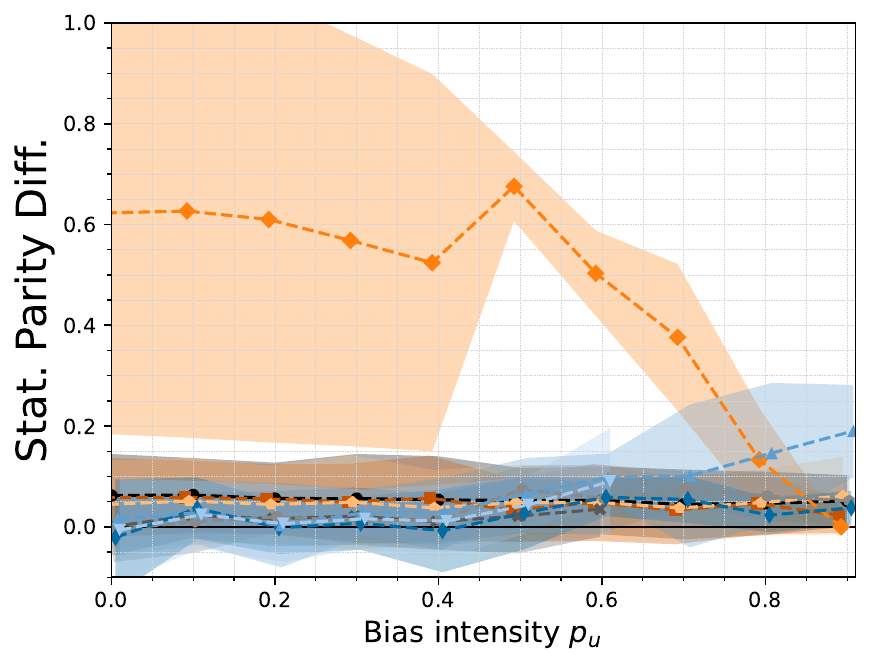}
            \subcaption{Biased evaluation}
        \end{minipage}
    \caption{Fair and biased evaluation of Accuracy and SPD for increasing levels of \textbf{self-selection} in \textbf{StudentBalanced} training sets.}
    \label{fig:trad_selectLow_studentBal}
    \end{figure}

    \begin{figure}[H]
    \centering
        \begin{minipage}[c]{.36\textwidth}
            \centering
            \includegraphics[width=\textwidth]{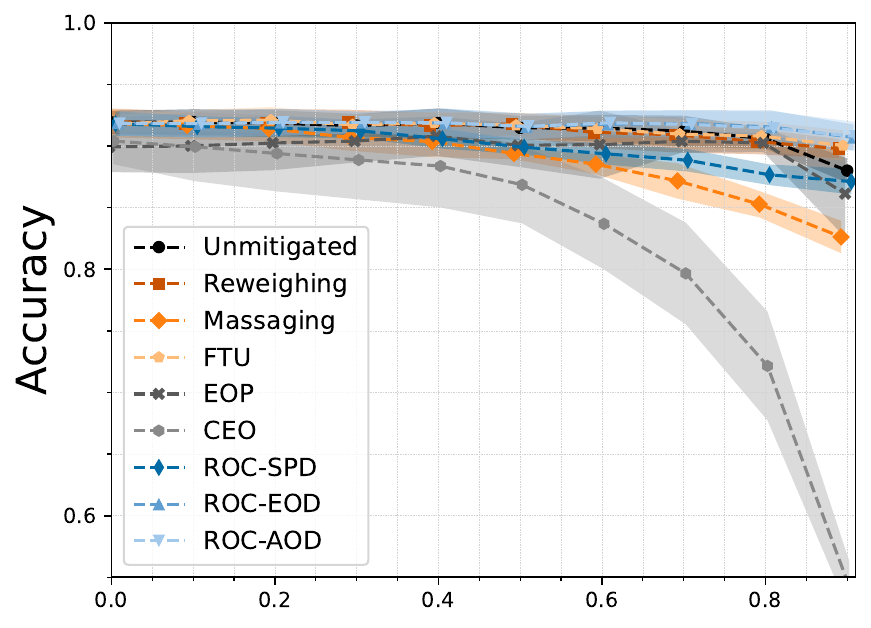}
        \end{minipage}
        \begin{minipage}[c]{.36\textwidth}
            \centering
            \includegraphics[width=\textwidth]{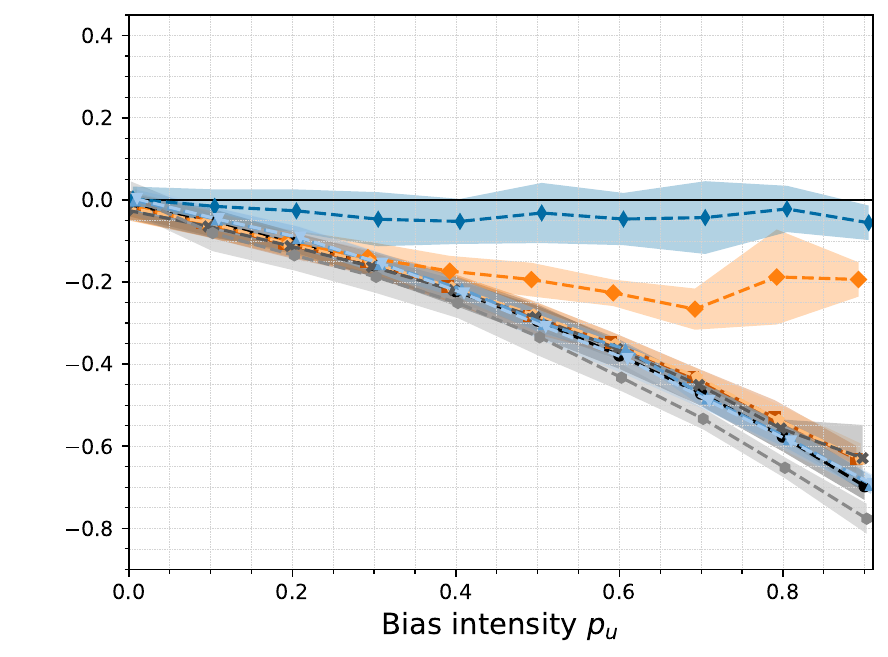}
        \end{minipage}
        \\
        \begin{minipage}[c]{.36\textwidth}
            \centering
            \includegraphics[width=\textwidth]{figures/tradeoff/tradeoff_selectDoubleProp_acc_OULADsocial_RF_FairTest.pdf}
            \subcaption{Fair evaluation}
        \end{minipage}
        \begin{minipage}[c]{.36\textwidth}
            \centering
            \includegraphics[width=\textwidth]{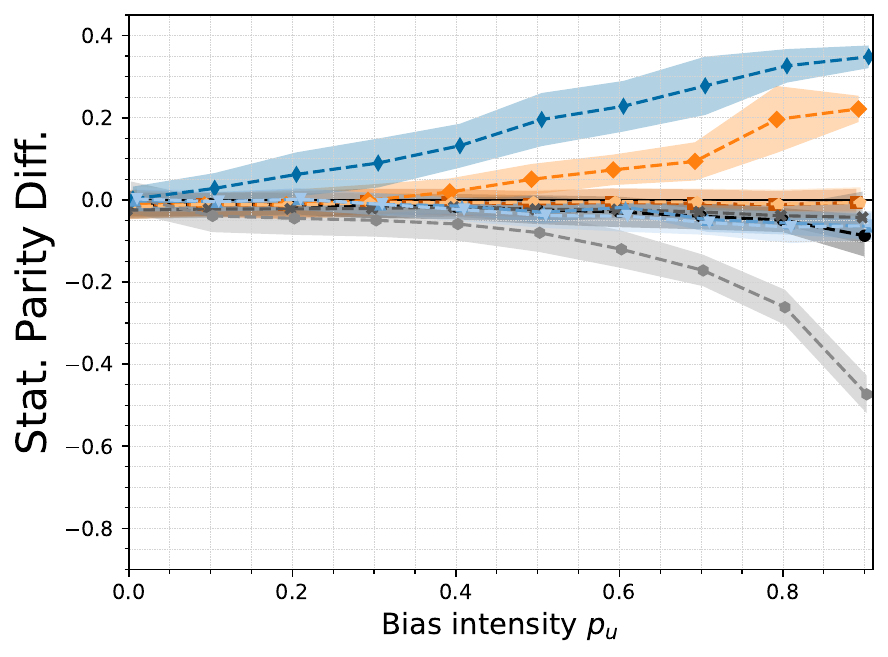}
            \subcaption{Biased evaluation}
        \end{minipage}
    \caption{Fair and biased evaluation of Accuracy and SPD for increasing levels of \textbf{malicious selection} in \textbf{OULADsocial} training sets.}
    \label{fig:trad_selectDouble_stem}
    \end{figure}
\newpage
    \begin{figure}[H]
    \centering
        \begin{minipage}[c]{.36\textwidth}
            \centering
            \includegraphics[width=\textwidth]{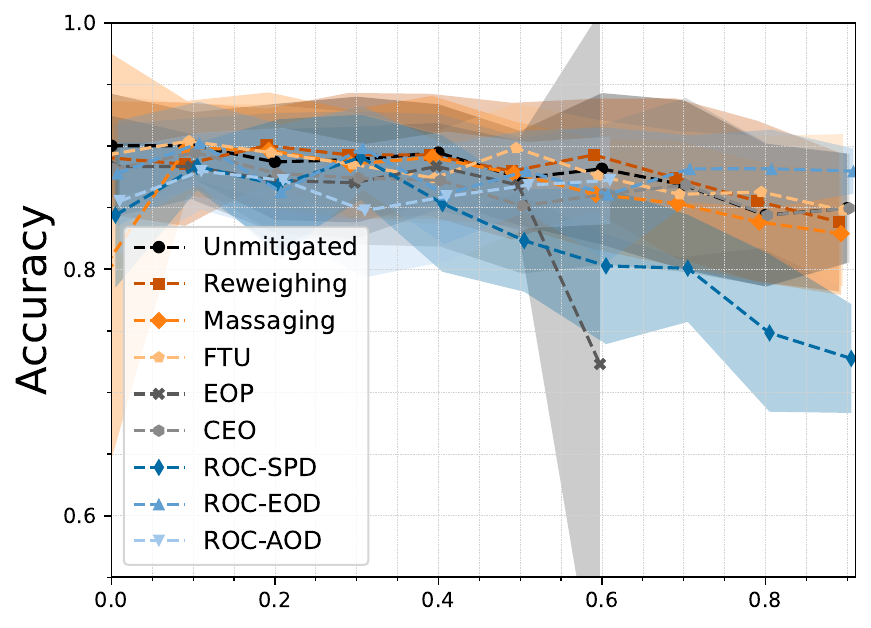}
        \end{minipage}
        \begin{minipage}[c]{.36\textwidth}
            \centering
            \includegraphics[width=\textwidth]{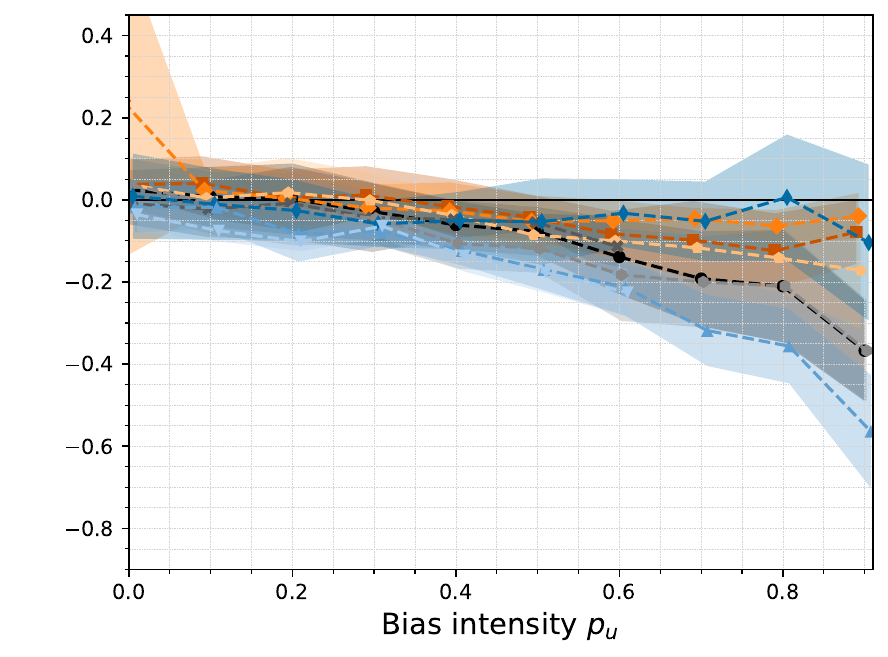}
        \end{minipage}
        \\
        \begin{minipage}[c]{.36\textwidth}
            \centering
            \includegraphics[width=\textwidth]{figures/tradeoff/tradeoff_selectDoubleProp_acc_studentBalanced_RF_FairTest.pdf}
            \subcaption{Fair evaluation}
        \end{minipage}
        \begin{minipage}[c]{.36\textwidth}
            \centering
            \includegraphics[width=\textwidth]{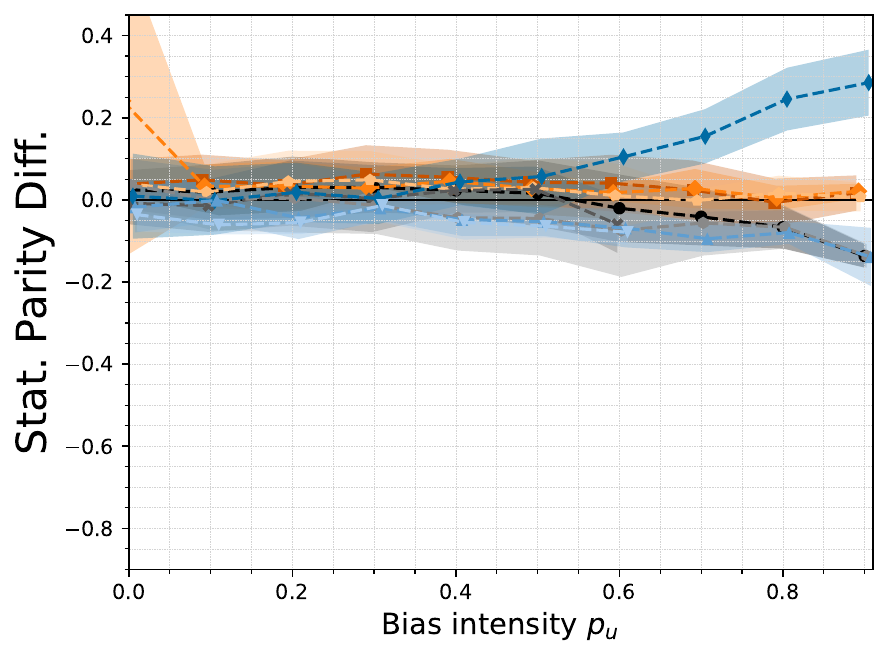}
            \subcaption{Biased evaluation}
        \end{minipage}
    \caption{Fair and biased evaluation of Accuracy and SPD for increasing levels of \textbf{malicious selection} in \textbf{StudentBalanced} training sets.}
    \label{fig:trad_selectDouble_studentBal}
    \end{figure}

\section{Additional results on bias impact on unmitigated models}\label{app:bias_on_model}

    \begin{figure}[H]
        \centering
        \includegraphics[width=0.5\linewidth]{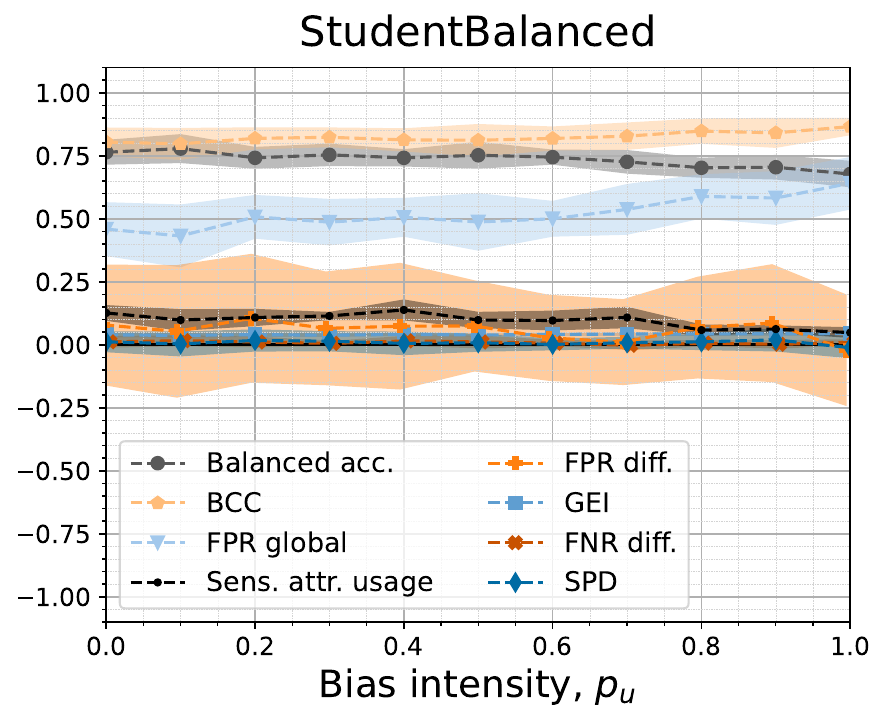}
        \caption{Effect of randomly \textbf{undersampling the whole training set} on RF models. When $p_u = 1$, the number of individuals removed corresponds to the size of the unprivileged group.}
        \label{fig:enter-label}
    \end{figure}

\section{Additional results on performance of bias mitigation methods}\label{app:mitig_comp}

    \begin{figure}[H]
        \begin{minipage}[c]{.33\textwidth}
            \begin{flushright}
            \includegraphics[height=4.3cm]{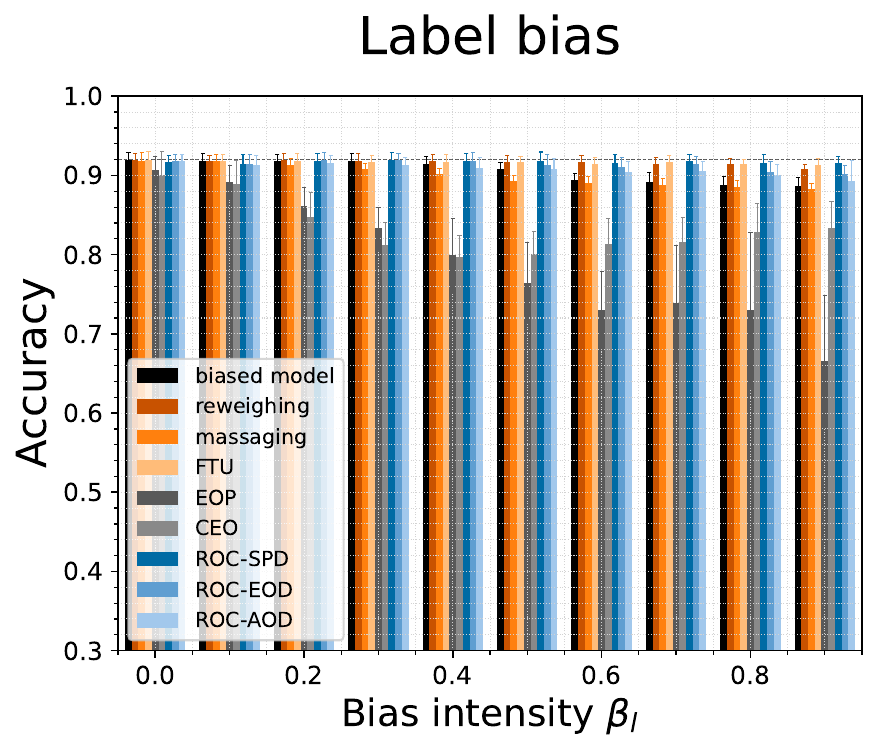}
            \end{flushright}
        \end{minipage}
        \begin{minipage}[c]{.33\textwidth}
            \begin{flushright}
            \includegraphics[height=4.3cm]{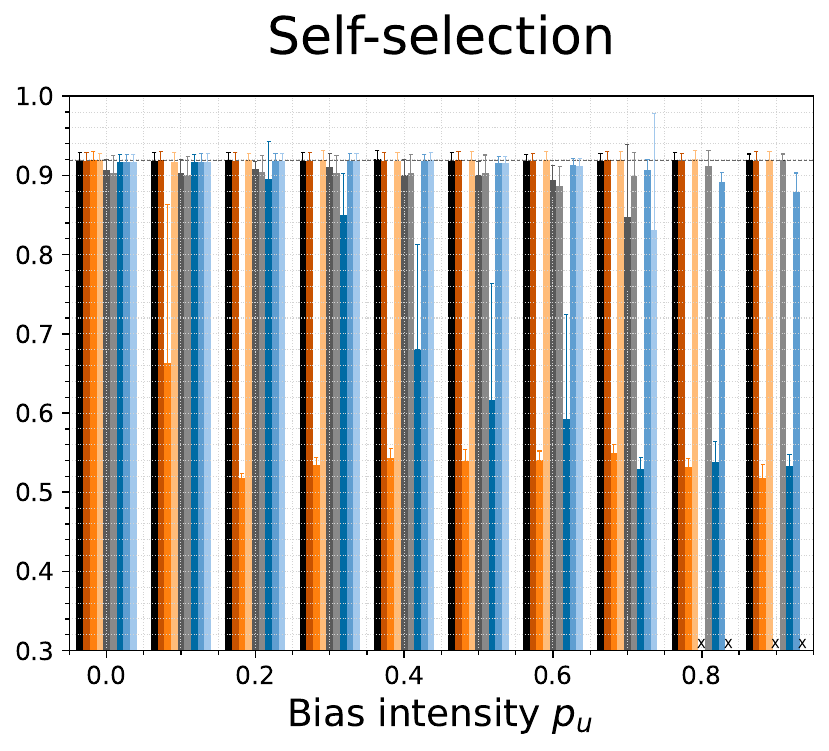}
            \end{flushright}
        \end{minipage}
        \begin{minipage}[c]{.33\textwidth}
            \begin{flushright}
            \includegraphics[height=4.3cm]{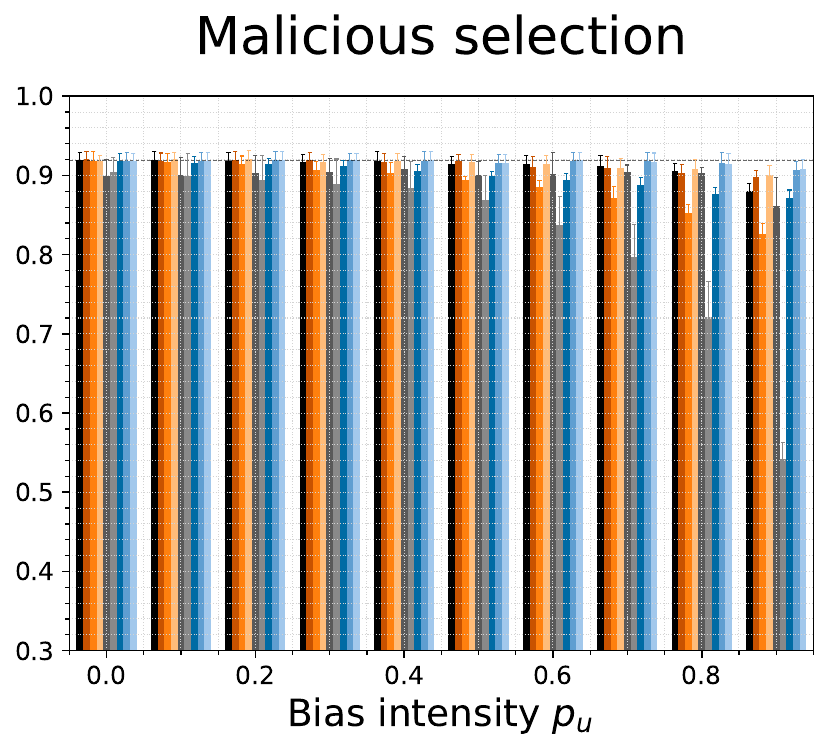}
            \end{flushright}
        \end{minipage}
        \\
        \begin{minipage}[c]{.33\textwidth}
            \begin{flushright}
            \includegraphics[height=3.8cm]{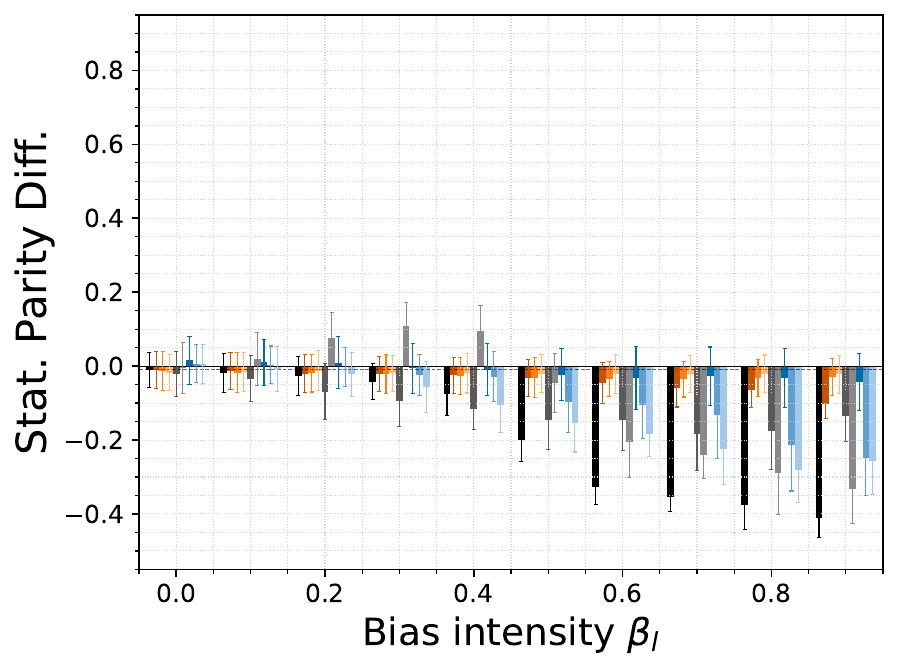}
            \end{flushright}
        \end{minipage}
        \begin{minipage}[c]{.33\textwidth}
            \begin{flushright}
            \includegraphics[height=3.8cm]{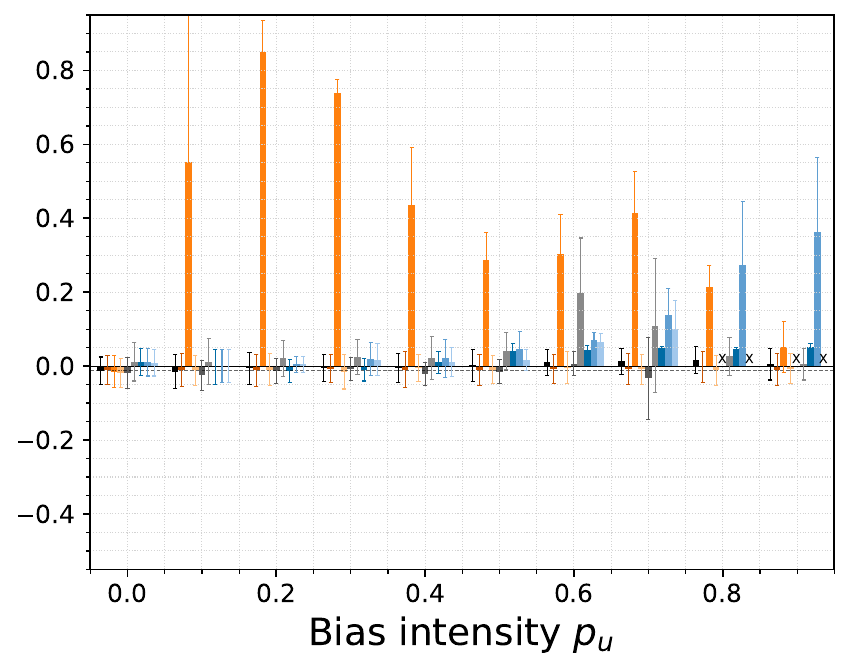}
            \end{flushright}
        \end{minipage}
        \begin{minipage}[c]{.33\textwidth}
            \begin{flushright}
            \includegraphics[height=3.8cm]{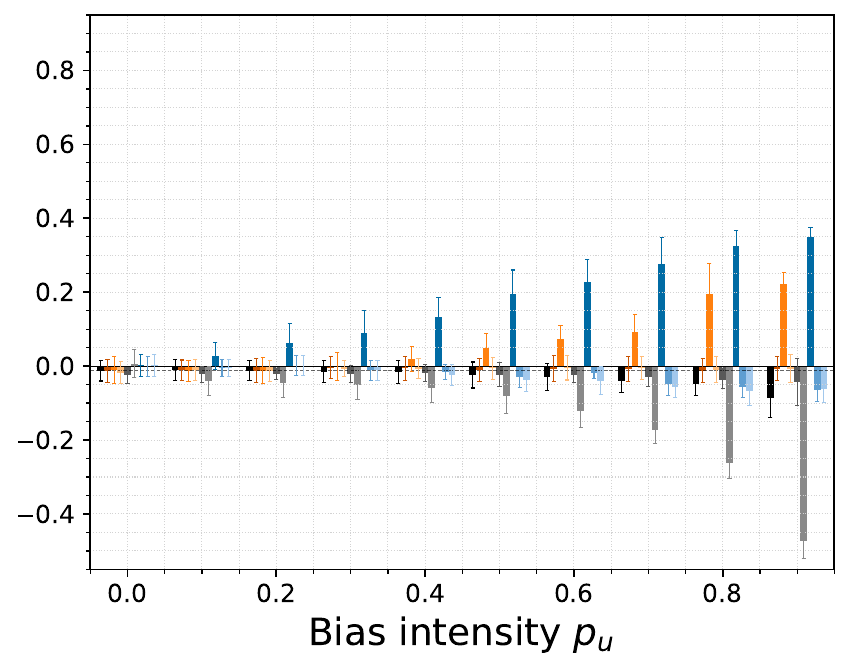}
            \end{flushright}
        \end{minipage}
        \\
        \begin{minipage}[c]{.33\textwidth}
            \begin{flushright}
            \includegraphics[height=3.8cm]{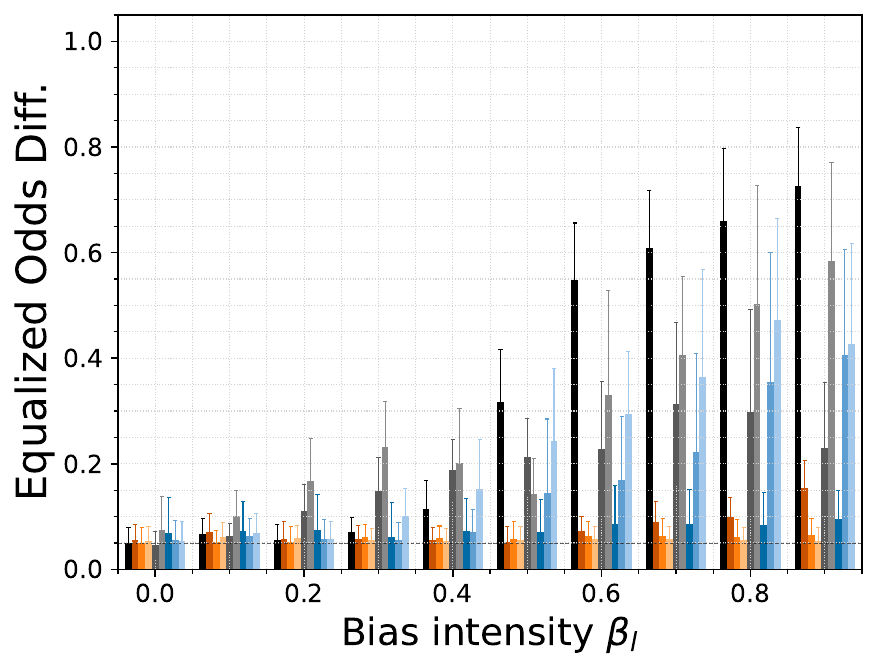}
            \end{flushright}
        \end{minipage}
        \begin{minipage}[c]{.33\textwidth}
            \begin{flushright}
            \includegraphics[height=3.8cm]{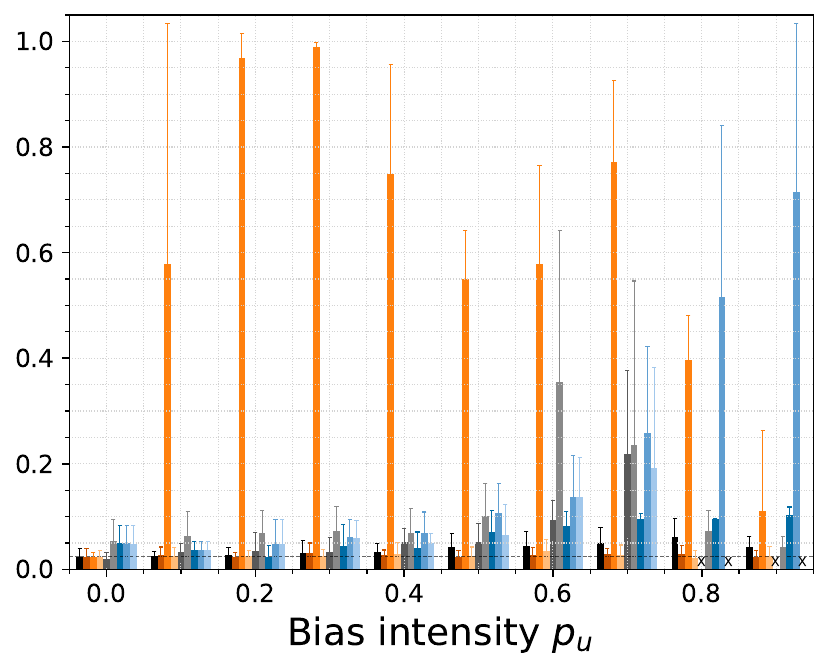}
            \end{flushright}
        \end{minipage}
        \begin{minipage}[c]{.33\textwidth}
            \begin{flushright}
            \includegraphics[height=3.8cm]{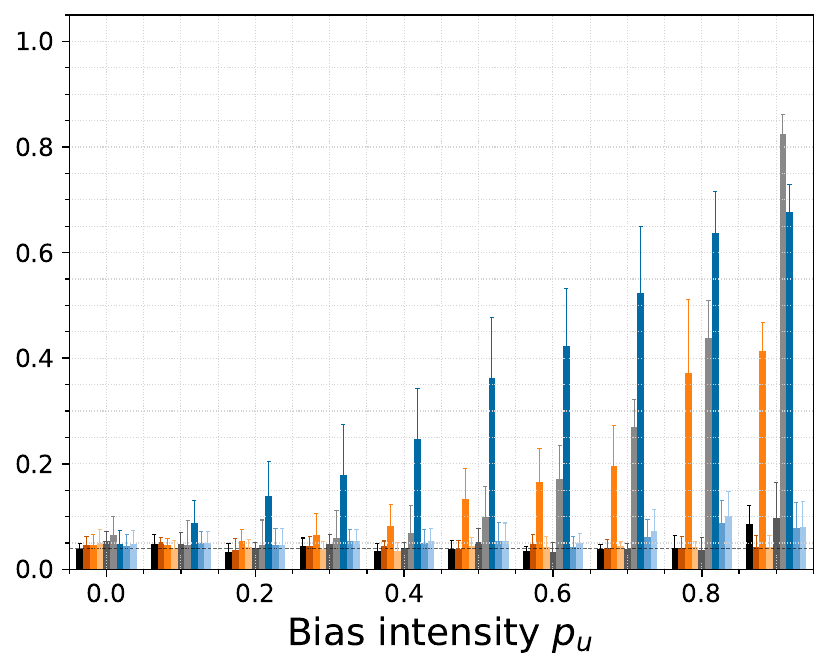}
            \end{flushright}
        \end{minipage}
        \\
        \begin{minipage}[c]{.33\textwidth}
            \begin{flushright}
            \includegraphics[height=3.8cm]{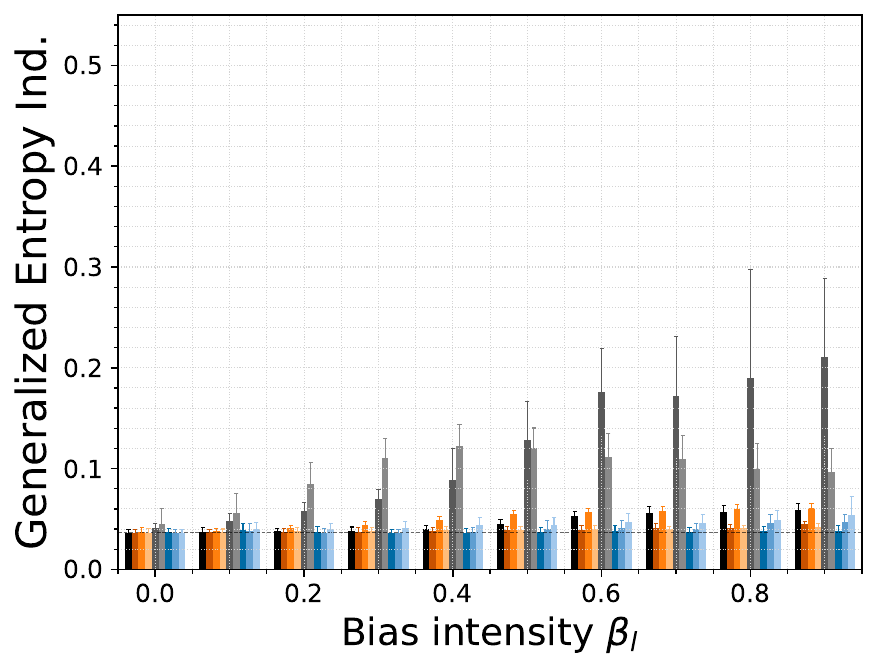}
            \end{flushright}
        \end{minipage}
        \begin{minipage}[c]{.33\textwidth}
            \begin{flushright}
            \includegraphics[height=3.8cm]{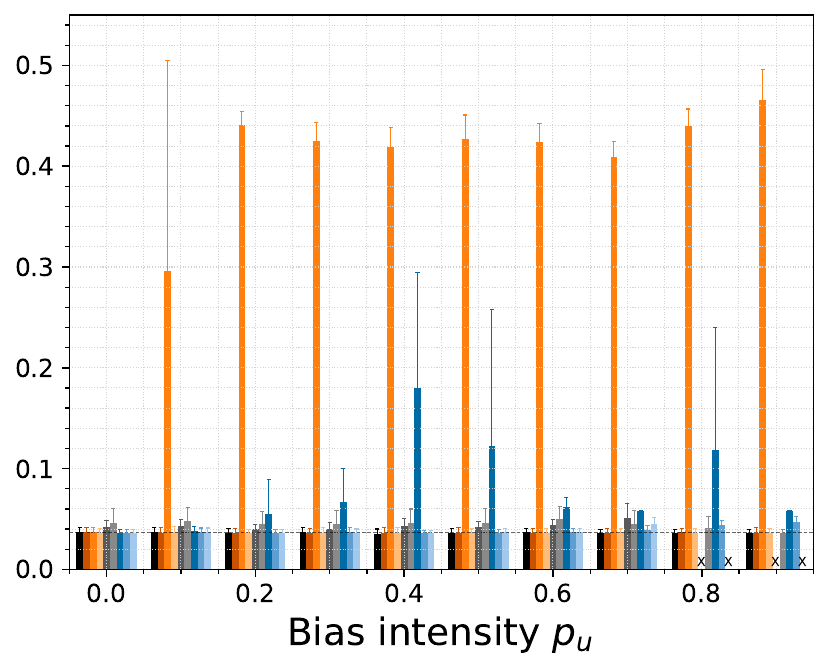}
            \end{flushright}
        \end{minipage}
        \begin{minipage}[c]{.33\textwidth}
            \begin{flushright}
            \includegraphics[height=3.8cm]{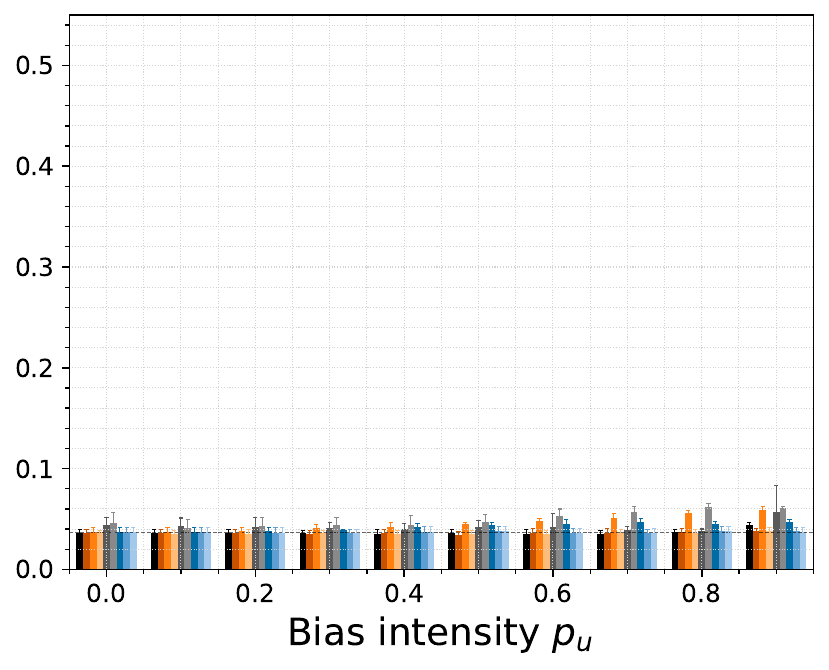}
            \end{flushright}
        \end{minipage}
    \caption{Evolution of accuracy and fairness metrics for RF models trained \textbf{OULADsocial} with increasing bias levels and evaluated on unbiased data. The fair baseline is indicated by a dashed horizontal black line. For label bias and self-selection, aggregated metrics (Accuracy and GEI) often have higher values than for OULADstem because the privileged group is a minority in OULADsocial and a majority in OULADstem.}
    \end{figure}

    \begin{figure}[H]
    \centering
        \begin{minipage}[c]{.33\textwidth}
            \begin{flushright}
            \includegraphics[height=4.3cm]{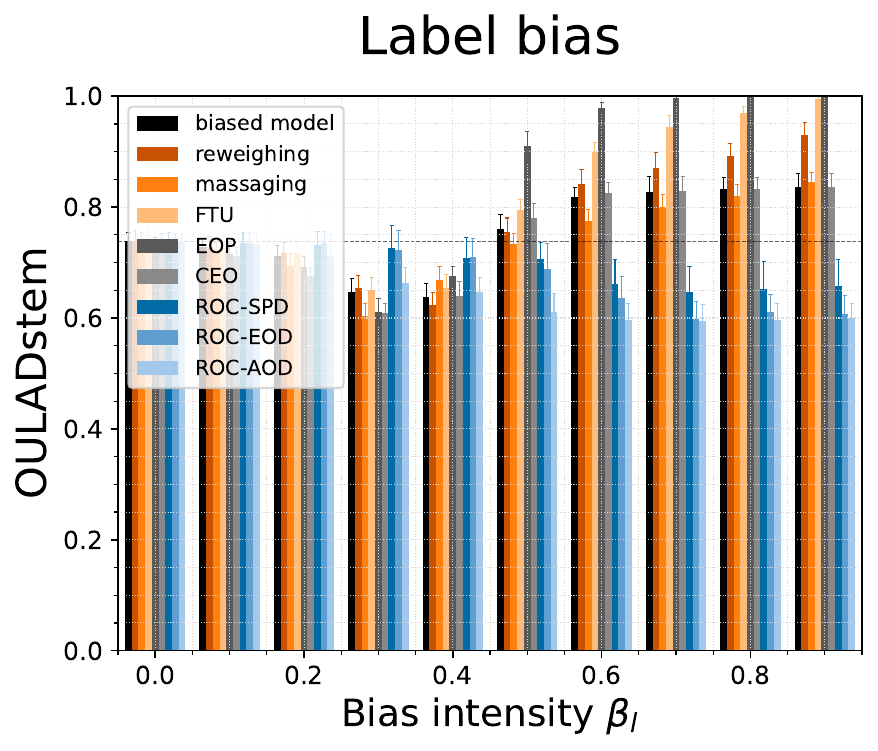}
            \end{flushright}
        \end{minipage}
        \begin{minipage}[c]{.33\textwidth}
            \begin{flushright}
            \includegraphics[height=4.3cm]{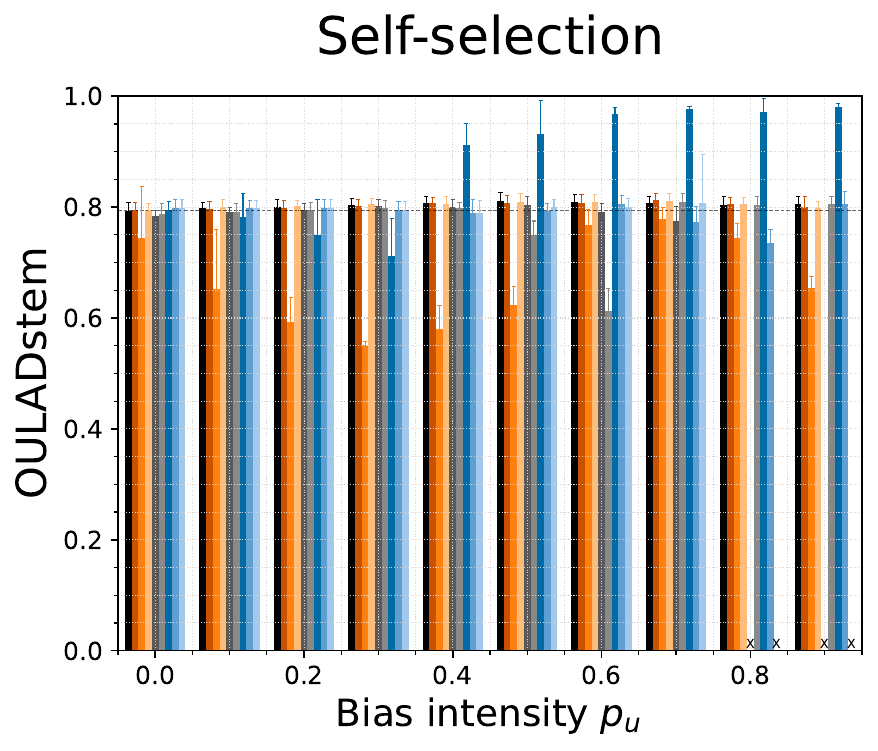}
            \end{flushright}
        \end{minipage}
        \begin{minipage}[c]{.33\textwidth}
            \begin{flushright}
            \includegraphics[height=4.3cm]{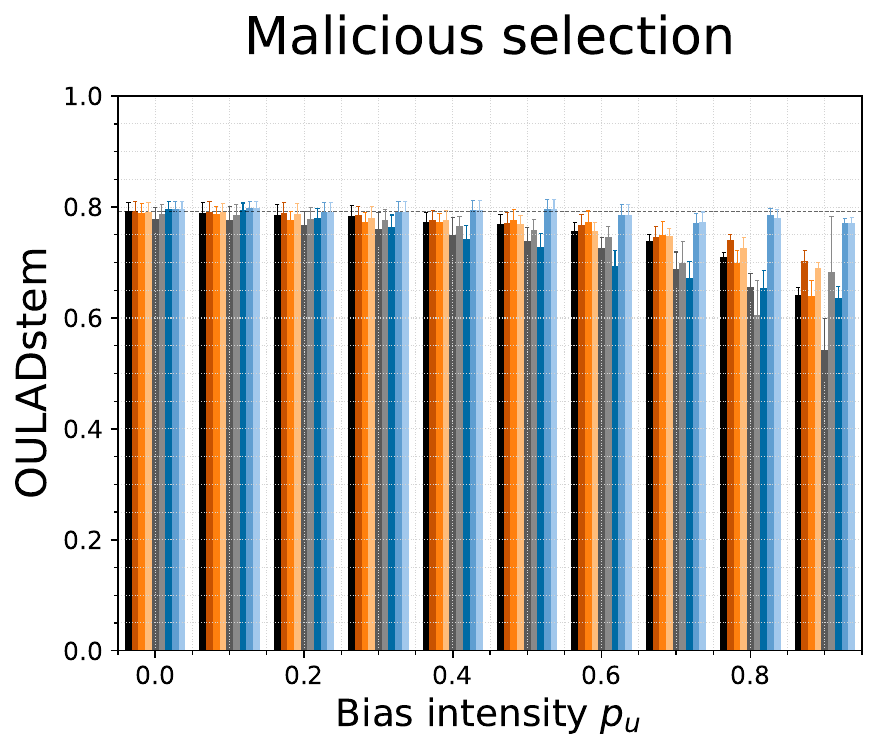}
            \end{flushright}
        \end{minipage}
        \\
        \begin{minipage}[c]{.33\textwidth}
            \begin{flushright}
            \includegraphics[height=3.8cm]{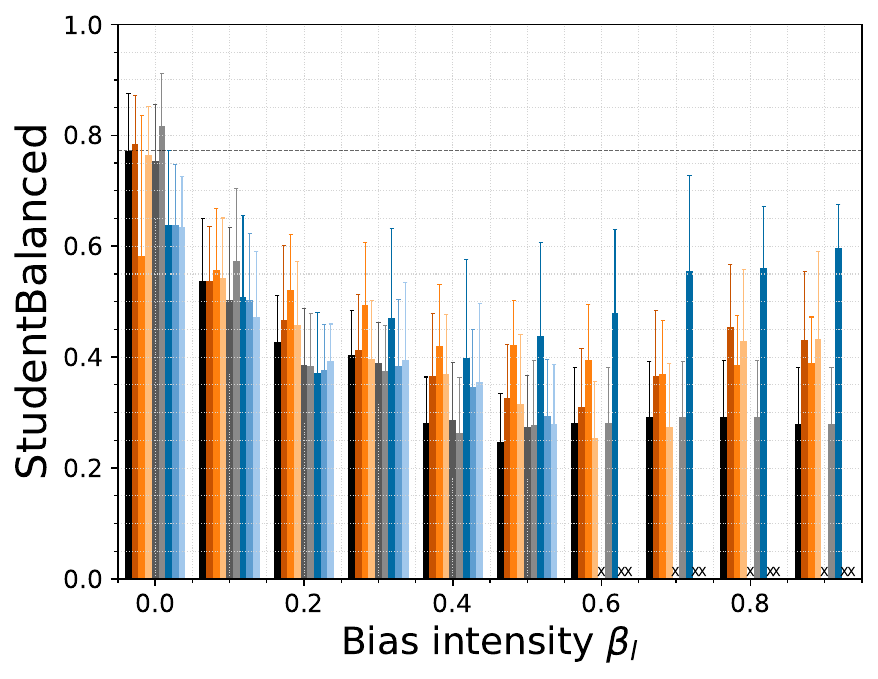}
            \end{flushright}
        \end{minipage}
        \begin{minipage}[c]{.33\textwidth}
            \begin{flushright}
            \includegraphics[height=3.8cm]{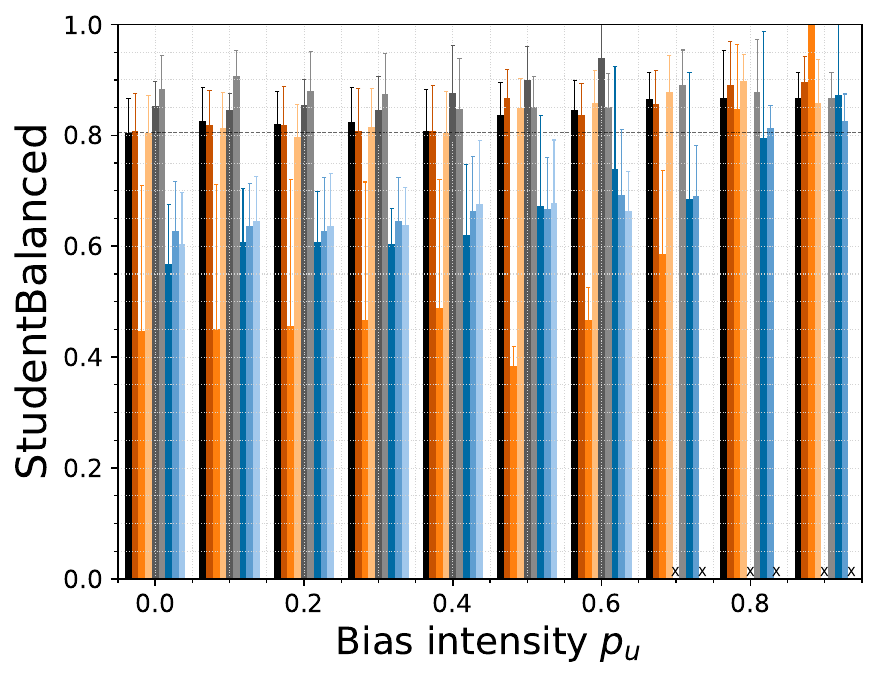}
            \end{flushright}
        \end{minipage}
        \begin{minipage}[c]{.33\textwidth}
            \begin{flushright}
            \includegraphics[height=3.8cm]{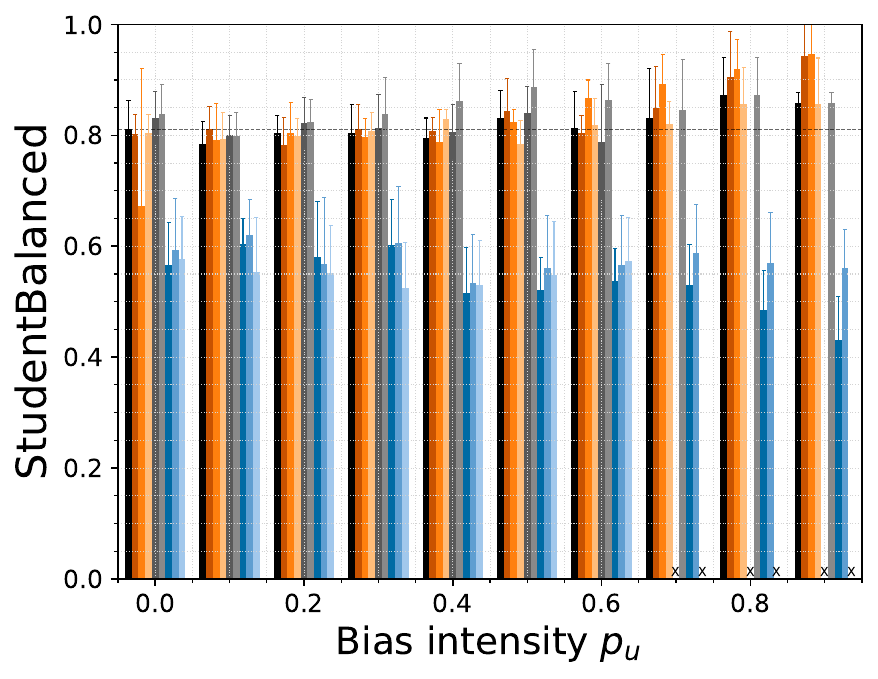}
            \end{flushright}
        \end{minipage}
        \\
        \begin{minipage}[c]{.33\textwidth}
            \begin{flushright}
            \includegraphics[height=3.8cm]{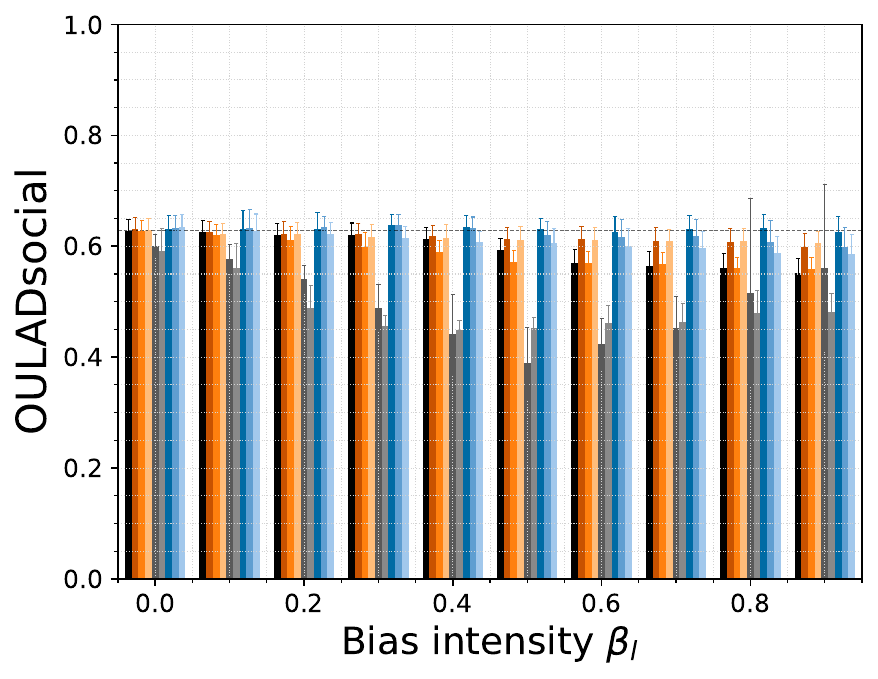}
            \end{flushright}
        \end{minipage}
        \begin{minipage}[c]{.33\textwidth}
            \begin{flushright}
            \includegraphics[height=3.8cm]{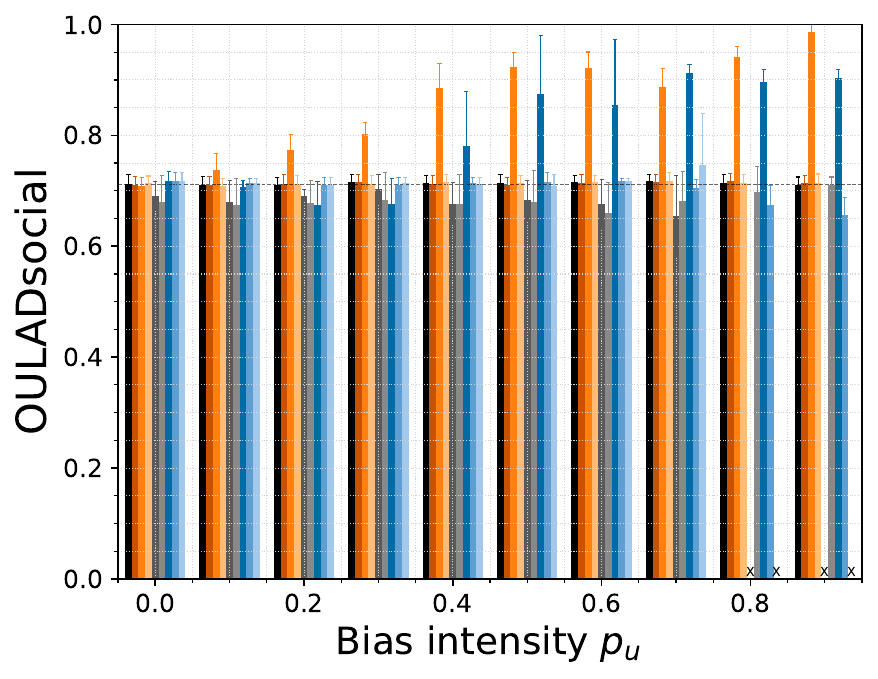}
            \end{flushright}
        \end{minipage}
        \begin{minipage}[c]{.33\textwidth}
            \begin{flushright}
            \includegraphics[height=3.8cm]{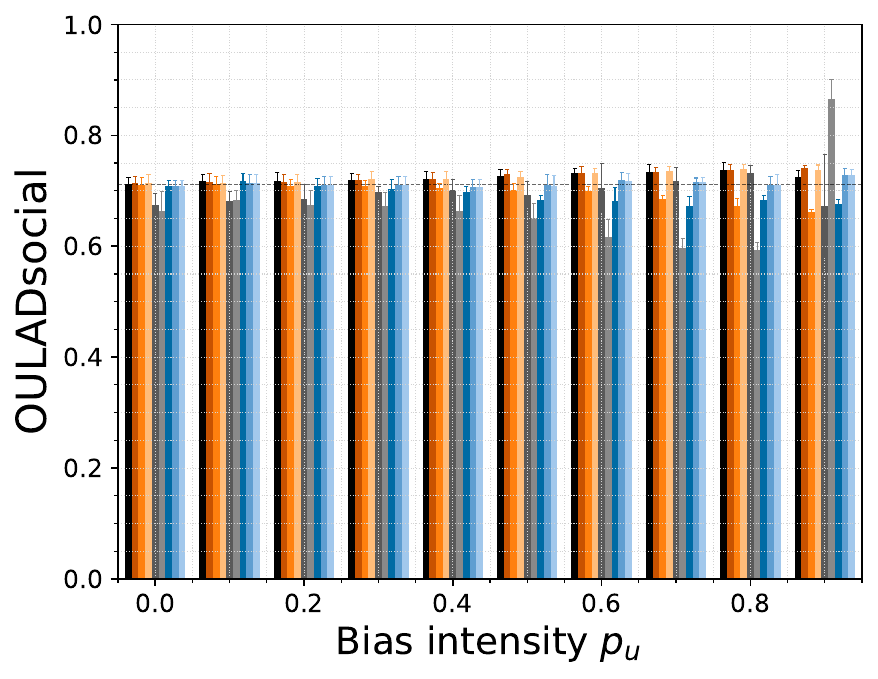}
            \end{flushright}
        \end{minipage}
    \caption{Evolution of \textbf{Balanced Conditioned Consistency} (BCC) for RF models trained on data with increasing bias levels and evaluated on unbiased data. The fair baseline is indicated by a dashed horizontal black line.}
    \end{figure}

\end{document}